\PassOptionsToPackage{table}{xcolor}
\PassOptionsToPackage{most}{tcolorbox}
\documentclass{article}
\usepackage{fancyhdr}

\usepackage{report}

\usepackage{soul}
\usepackage[utf8]{inputenc} 
\usepackage[T1]{fontenc}    
\usepackage{amsmath}
\usepackage{hyperref}       
\usepackage{cleveref}
\usepackage{url}            
\usepackage{booktabs}       
\usepackage{amsfonts}       
\usepackage{fancyhdr}       

\usepackage{nicefrac}       
\usepackage{microtype}      
\usepackage{graphicx}
\usepackage{pifont}
\usepackage{multirow}
\usepackage{CJKutf8}
\definecolor{darkmagenta}{rgb}{0.56, 0.0, 1.0}
\definecolor{softyellow}{rgb}{1.0, 0.92, 0.3} 
\definecolor{LightAquamarine}{rgb}{0.75, 1.0, 0.8} 
\definecolor{FireBrick}{RGB}{178,34,34}
\definecolor{MediumPurple}{RGB}{147,112,219}

\definecolor{uclablue}{rgb}{0.15, 0.45, 0.68}
\hypersetup{
    breaklinks,
    colorlinks=true,
    citecolor={darkmagenta},
    linkcolor={uclablue},
    urlcolor={uclablue}
}
\usepackage{wrapfig}
\usepackage{float}
\usepackage{subcaption}
\usepackage{placeins}
\usepackage{lipsum} 
\usepackage{tcolorbox}
\usepackage{amsmath}
\usepackage{amssymb}
\usepackage{utfsym}
\usepackage{fontawesome}
\usepackage{xspace}
\usepackage{enumitem}
\usepackage{multirow} 
\tcbuselibrary{most}
\usepackage{enumitem}
\usepackage{colortbl}
\usepackage{fancyhdr}

\usepackage{tcolorbox}
\usepackage{transparent}

\usepackage{xcolor}
\usepackage{xcolor} 
\usepackage{makecell}
\usepackage{tabularx}

\usepackage{tcolorbox}
\usepackage{xcolor} 
\usepackage{enumitem}       
\usepackage{listings}
\newcommand{\maybeincludegraphics}[2][]{\IfFileExists{#2}{\includegraphics[#1]{#2}}{}}
\usepackage{titlesec}
\usepackage{longtable}
\usepackage{float}
\usepackage{fontawesome}
\lstdefinelanguage{json}{
    basicstyle=\small\ttfamily,
    showstringspaces=false,
    breaklines=true,
    string=[s]{"}{"},
    stringstyle=\color{uclablue},
    morecomment=[l]{//},
    commentstyle=\color{gray}
}

\definecolor{njuPurple}{RGB}{220,205,230}     
\definecolor{njuPurpleLight}{RGB}{250,245,252}   

\newtcolorbox{abstractbox}{
    colback=njuPurpleLight,
    colframe=njuPurple,
    boxrule=1pt,
    arc=4mm,
    left=8pt,
    right=8pt,
    top=8pt,
    bottom=8pt,
    opacityback=0.95
}

\title{\textbf{OmniCap-IF: Benchmarking and Improving Instruction Following Abilities for Omni-Video Captioning}}

\author{
\textbf{Jiahao Wang$^{1*}$},
\textbf{An Ping$^{1*}$},
\textbf{Yanghai Wang$^{1*}$}, \\
\textbf{Yuanxing Zhang$^{2}$},
\textbf{Shihao Li$^{1}$},
\textbf{Hanyan Bian$^{1}$},
\textbf{Yichi Ren$^{1}$}, \\
\textbf{Yize Zhang$^{1}$},
\textbf{Han Wang$^{1}$},
\textbf{Haowen Chen$^{1}$},
\textbf{Junze Li$^{1}$}, \\
\textbf{Jiaqi Wang$^{1}$},
\textbf{Yiyang Hu$^{1}$},
\textbf{Zhuze Xu$^{1}$},
\textbf{Zijie Zhang$^{1}$},
\textbf{Jiaheng Liu$^{1, \dagger}$} \\
\vspace{4mm}
{\normalsize $^1$ NJU-LINK Team, Nanjing University} \quad
{\normalsize $^2$ Kling Team, Kuaishou Technology} \\ 
\vspace{2mm}
\texttt{jiahaowang@smail.nju.edu.cn}
\quad\quad\quad
\texttt{liujiaheng@nju.edu.cn} \\
}

\begin{document}

\maketitle
\let\oldthefootnote\thefootnote

\let\thefootnote\relax\footnotetext{*~Equal Contribution. ~~$^\dagger$~Corresponding Author.}
\let\thefootnote\oldthefootnote

\begin{abstractbox}
\begin{center}
\textbf{\Large Abstract}
\end{center}
While Omni-modal Large Language Models (OLLMs) have demonstrated impressive capabilities in jointly processing audio and visual streams, their ability to strictly adhere to complex, multi-faceted user instructions remains largely unexplored. Existing benchmarks primarily focus on holistic video understanding or text-only instruction following, failing to capture the intricate interplay between modalities and user constraints. To bridge this gap, we introduce \textbf{OmniCap-IF} \textsuperscript{\textit{\href{https://github.com/NJU-LINK/OmniCap-IF}{a}}, \textit{\href{https://huggingface.co/datasets/NJU-LINK/OmniCap-IF}{b}}}, the first comprehensive benchmark specifically designed to evaluate instruction-following capabilities in omni-modal captioning. OmniCap-IF incorporates a systematic framework that assesses captions on two dimensions: format correctness and content correctness. Our benchmark encompasses 50 distinct constraint types across pure visual, pure audio, and audio-visual modalities, while integrating Temporal Grounding to assess spatio-temporal precision. Extensive evaluations of prominent models on 1,920 high-quality samples reveal significant performance disparities. Furthermore, our analysis uncovers a critical ``format-content tradeoff,'' demonstrating that increasing formatting complexity directly degrades models' omni-modal reasoning abilities. Finally, to advance the field, we curate a 54K instruction-tuning dataset, OmniCap-IF-54K and present OmniCaptioner-IF, which achieves notable improvements in both complex instruction adherence and general omni-modal captioning performance.

\noindent\rule{0.32\linewidth}{0.4pt}

{\footnotesize
\textsuperscript{\textit{a}}\href{https://github.com/NJU-LINK/OmniCap-IF}{https://github.com/NJU-LINK/OmniCap-IF}\\
\textsuperscript{\textit{b}}\href{https://huggingface.co/datasets/NJU-LINK/OmniCap-IF}{https://huggingface.co/datasets/NJU-LINK/OmniCap-IF}
}

\end{abstractbox}

\begin{figure}[htbp]
  \centering
  \includegraphics[width=0.85\linewidth]{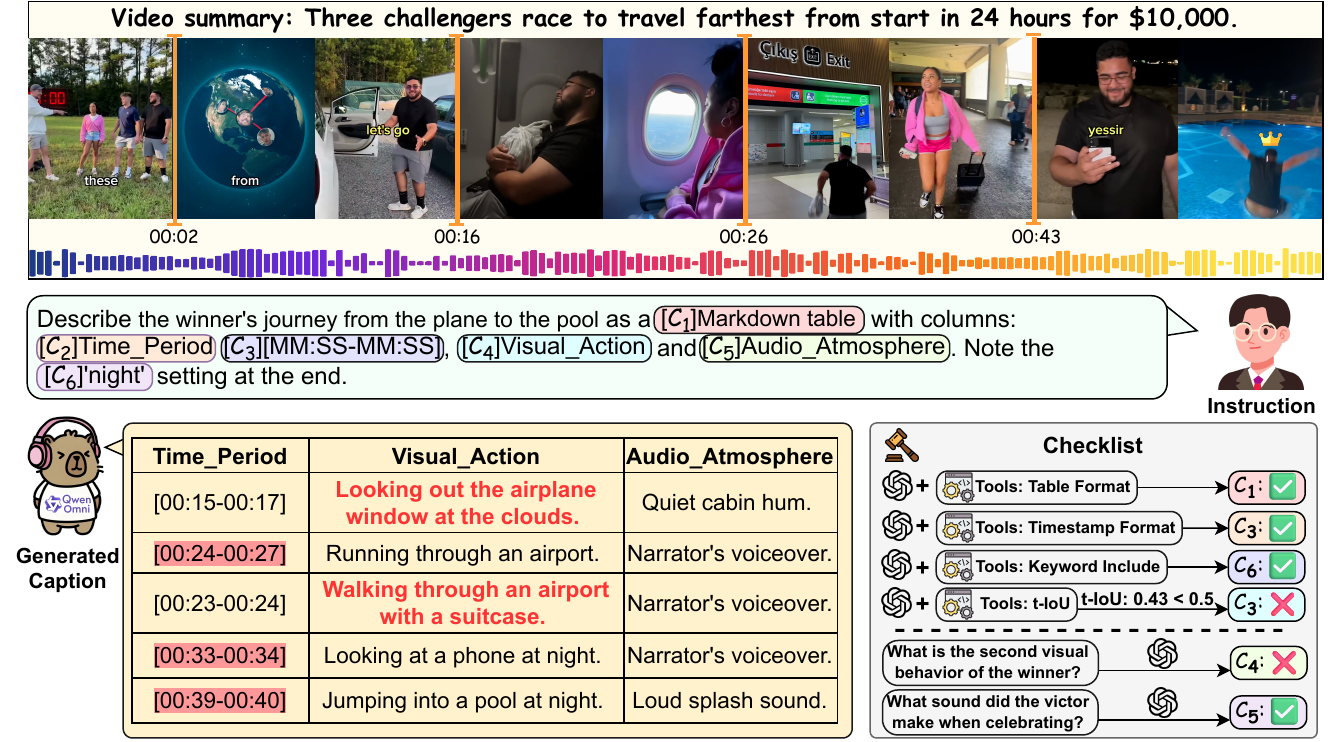}
  \caption{Overview of the OmniCap-IF evaluation framework. A typical case takes audio-visual content and a constraint-rich instruction as input. The generated response is evaluated against a specific checklist using a dual-mechanism approach: (1) Format and temporal constraints are initially extracted by the judge LLM and subsequently evaluated using predefined tools (e.g., format checkers, t-IoU) to ensure objective assessment. (2) Content constraints are verified by the judge model that answers preset questions to confirm the factual accuracy of visual and audio details.}
  \label{fig:teaser}
\end{figure}

\section{Introduction}
The evolution of Multimodal Large Language Models (MLLMs) has recently transitioned from vision-language integration to omni-modal perception, enabling joint reasoning over text, audio and visual streams natively~\citep{liu2023visual, dai2023instructblip, chu2023qwen, wang2025vr, ddk}. Despite their proficiency in general video description, high-quality, controllable outputs are crucial for a range of downstream tasks, including structured dual-track scripts for text-to-audio-video (T2AV) generation~\citep{cao2025t2av}, egocentric action descriptions for embodied task planning~\citep{chen2026egoplan}, and precise semantic fingerprints for cross-modal retrieval~\citep{JCST-2509-15922}. Models must not only understand the omni-modal content but also adhere strictly to complex, user-defined instructions. As illustrated in Figure~\ref{fig:teaser}, even leading models struggle to balance multi-modal perception with rigorous constraint satisfaction, often sacrificing instruction fidelity for descriptive verbosity~\citep{liu2023lost, guan2024hallusionbench}.


Currently, evaluating an omni-modal model's capacity to fulfill compositional constraints remains an unexplored challenge~\citep{tian2018audio}. Existing benchmarks either prioritize semantic richness and question-answering accuracy over programmatic verifiability~\citep{li2024mvbench, maaz2024video} or are confined to single-modality instruction following~\citep{zhou2023ifeval, bitton2023visit}. Consequently, they lack the joint audio-visual complexity and structural rigor required for comprehensive omni-modal evaluation.


To bridge this gap, we propose \textbf{OmniCap-IF}, the first benchmark dedicated to instruction-following in omni-modal captioning. We establish a systematic constraint framework of 50 constraint types spanning format and content dimensions—the latter decomposed into Visual, Audio, and Audio-Visual modalities. Furthermore, we incorporate Temporal Grounding \citep{krishna2017dense} to enable quantitative assessment of precise timestamp localization, better aligning the evaluation with real-world scenarios.


Moreover, through our decoupled evaluation protocol, we investigate the impact of formatting difficulty. We uncover a significant ``format-content tradeoff''—demonstrating that as structural format constraints become more rigorous, models' fundamental ability to accurately reason over audio-visual content drastically degrades. Finally, to advance controllable generation, we construct OmniCap-IF-54K, a large-scale omni-modal instruction-tuning dataset, and present OmniCaptioner-IF, demonstrating a viable path toward highly controllable omni-modal assistants.

In summary, our key contributions are:
\begin{itemize}[leftmargin=*, topsep=0pt, itemsep=0pt, parsep=0pt]
    \item \textbf{The first instruction-following benchmark for omni-modal captioning.} We introduce OmniCap-IF, featuring 1,920 complex, compositional instructions tailored for downstream applications. 
    \item \textbf{A robust evaluation protocol disentangling format and content assessment.} We design a system that separates structural verification from semantic fidelity. It comprehensively covers Visual, Audio, and Audio-Visual constraints while uniquely incorporating Temporal Grounding.
    \item \textbf{Discovery of the ``format-content tradeoff''.} We uncover and empirically prove that strict syntactic constraints (e.g., JSON) severely bottleneck models' fundamental reasoning capabilities.
    \item \textbf{A high-quality training dataset and a strong baseline for controllable generation.} We release OmniCap-IF-54K along with the OmniCaptioner-IF model. Our results demonstrate that targeted instruction tuning significantly enhances both instruction adherence and general omni-modal perception.
\end{itemize}

\section{Related Work}
\subsection{Instruction-Following Benchmarks}
Evaluating instruction adherence has evolved significantly alongside the rapid development of large language models. Early text-based benchmarks primarily focused on assessing models against verifiable programmatic constraints, multi-level structural formatting, and complex logical rules~\citep{zhou2023ifeval, jiang2024followbench, wen2024complex, zhang2025inverse}. Recent efforts have extended this evaluation paradigm to vision-language tasks~\citep{bitton2023visit, li2026ifvidcap}. Furthermore, while recent studies have observed that enforcing strict structural formatting can degrade the intrinsic reasoning capabilities of Large Language Models~\citep{tam2024let, deng2025decoupling}, this phenomenon remains largely unexplored in complex multi-modal scenarios. Despite these advancements, existing evaluations remain confined to partial modalities and fail to meet the intricate requirements of emerging downstream applications. OmniCap-IF advances this paradigm by introducing omni-modal constraints and fine-grained temporal localization, effectively bridging the gap toward comprehensive omni-modal instruction following.

\subsection{Omni-Modal Captioning Benchmarks}
The recent advent of native omni-modal large language models has significantly expanded the boundaries of joint audio-visual understanding. Consequently, recent omni-modal captioning benchmarks primarily focus on assessing the semantic accuracy and descriptive richness of generated text, rather than a model’s capacity to follow arbitrary or user-specified instructions. These benchmarks commonly adopt structured evaluation paradigms, including curated question–answer pairs~\citep{wu2025ugc,li2025omnivideobench,mvueval}, cloze-style assessments~\citep{ma2026omnicaptioner}, detailed holistic audio-visual descriptions~\citep{tang2025salmonn2} and temporally-grounded cinematographic scripts~\citep{yao2026timechat}. Moreover, traditional temporal grounding tasks typically frame fine-grained event localization as isolated, pre-defined predictive tasks~\citep{lei2020tvr}, lacking flexibility for dynamic or customized constraints. While such designs play a vital role in advancing omni-modal descriptive performance, they share a fundamental limitation: evaluation is conducted against a predefined and static set of quality criteria. In contrast, OmniCap-IF represents the first benchmark in omni-modal captioning that explicitly targets a model’s ability to understand and execute diverse, compositional instructions spanning visual, auditory, and audio-visual modalities.

\section{OmniCap-IF}
\subsection{Constraint Framework}
To systematically evaluate omni-modal controllability, we construct a taxonomy encompassing 50 constraint types categorized into two primary dimensions (Figure~\ref{fig:dist_sunburst}):

\noindent\textbf{1. Format Constraints:} Covers objective Structural (e.g., JSON arrays, Markdown tables) and Stylistic (e.g., length limits, specific delimiters) requirements.

\noindent\textbf{2. Content Constraints:} Demands fine-grained factual comprehension across three granularities: (1)~Visual (perceivable solely from the visual track, e.g., visual entities); (2)~Audio (derivable exclusively from the auditory stream, e.g., speaker timbre); (3)~Audio-Visual (requiring the simultaneous integration of both streams, such as audio-visual event alignment). 

\noindent Further granular details regarding the classification and task definitions can be found in Appendix~\ref{sec:system}.

\subsection{Data Collection and Annotation\protect\footnotemark}
\footnotetext{More details for the test set construction can be found in Appendix~\ref{sec:test_construction}.}
\subsubsection{Video Collection}
To construct a high-quality evaluation benchmark, we curate a test set of 480 videos by compiling a large-scale, copyright-free video pool sourced from YouTube, TikTok, and Ego4D~\citep{grauman2022ego4d}. The videos are rigorously filtered to ensure both audio-visual richness and audio-visual alignment. The final collection spans a wide range of domains—from comedy to technology—thereby enhancing the overall reliability and diversity of the benchmark.

\subsubsection{Annotation Pipeline}\label{annotation}
\noindent Our annotation pipeline follows a two-stage framework that integrates automated generation with human expertise, ensuring both scalability and high annotation quality.\\
\textbf{Stage 1: Automated Draft Generation.} For each video, an Instruction Generator produces paired instruction–checklist annotations. The prompts for generation can be found in Appendix~\ref{sec:test_set_prompt}.\\
\textbf{Stage 2: Human Refinement and Verification.} Professionally trained annotators carefully review and refine the automatically generated drafts, resulting in 53.1\% of samples modified and 22.7\% discarded and rewritten. Each sample is finalized only upon unanimous agreement among three annotators, with any disagreements adjudicated by a senior supervisor. Through this rigorous process, we obtain a final dataset comprising 1,920 high-quality samples.\\
\noindent Representative dataset samples, including their corresponding multi-constraint instructions and evaluation checklists, are showcased in Appendix~\ref{sec:data_sample}.



\subsection{Dataset Statistics}
\subsubsection{Overall Statistics} 
\begin{figure*}[t]
    \centering
    
    \begin{minipage}[t]{0.32\textwidth}
        \vspace{0pt} 
        
        \begin{subfigure}[b]{\textwidth}
            \centering
            \includegraphics[width=0.75\textwidth]{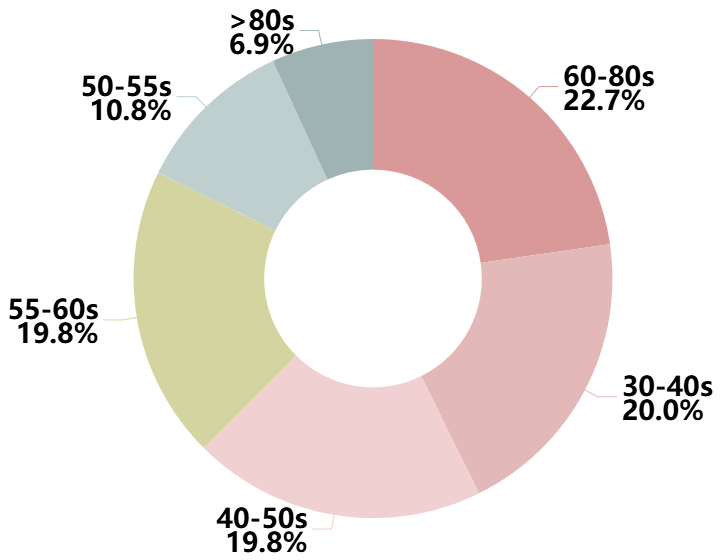}
            \caption{Video duration.}
            \label{fig:dist_duration}
        \end{subfigure}
        
        \vfill 
        \vspace{0.5cm} 
        
        \begin{subfigure}[b]{\textwidth}
            \centering
            \includegraphics[width=1.05\textwidth]{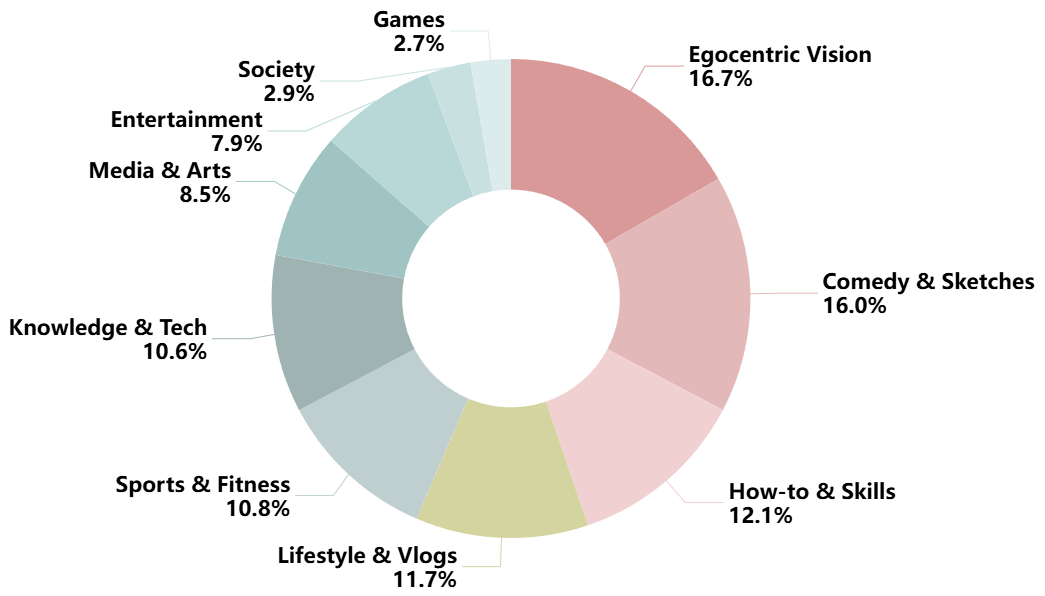}
            \caption{Video category.}
            \label{fig:dist_category}
        \end{subfigure}
        
        \vfill
        \vspace{0.5cm} 
        
        \begin{subfigure}[b]{\textwidth}
            \centering
            \includegraphics[width=\textwidth]{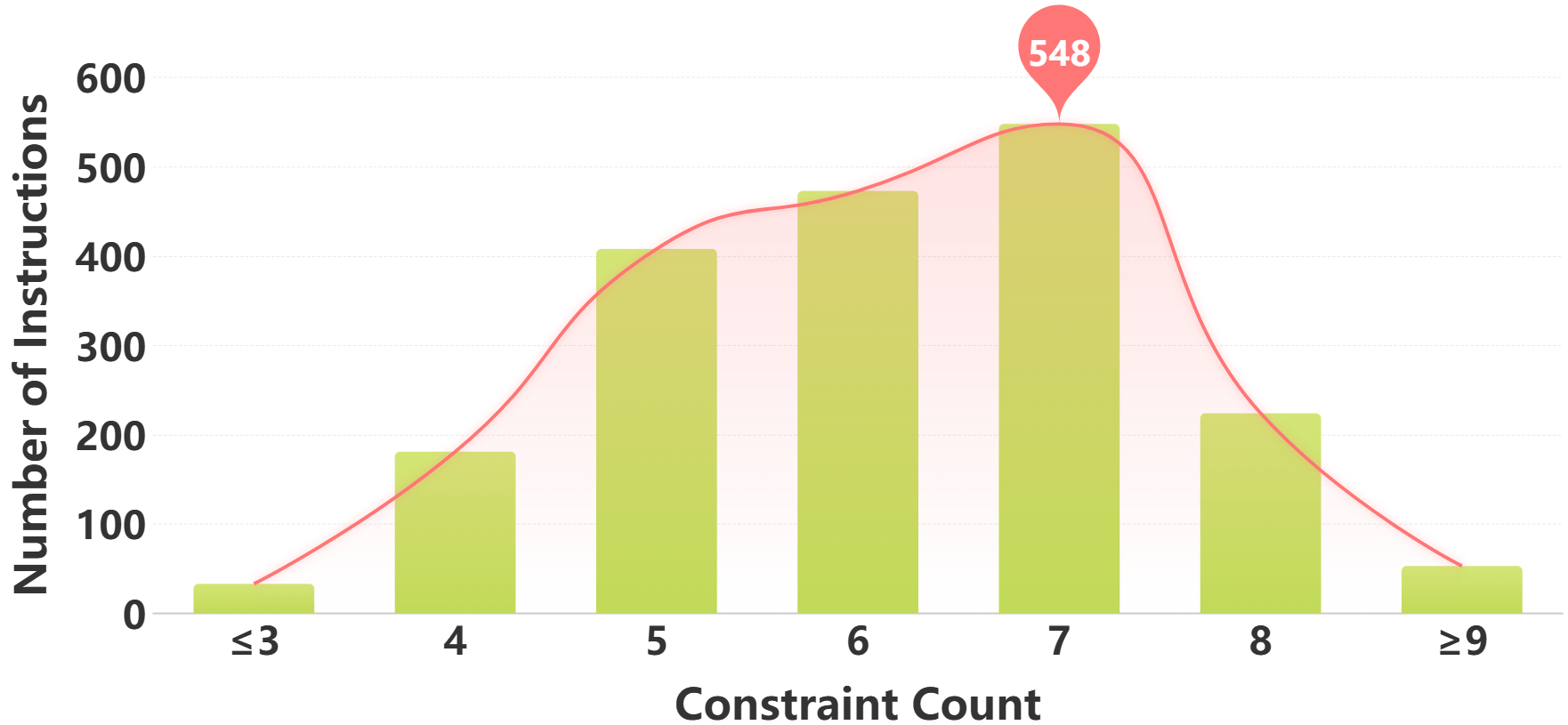}
            \caption{Constraint count.}
            \label{fig:dist_constraint}
        \end{subfigure}
        
    \end{minipage}%
    \hfill
    \begin{minipage}[t]{0.64\textwidth}
        \vspace{0pt} 
        
        \begin{subfigure}[b]{\textwidth}
            \centering
            \includegraphics[width=1.0\textwidth]{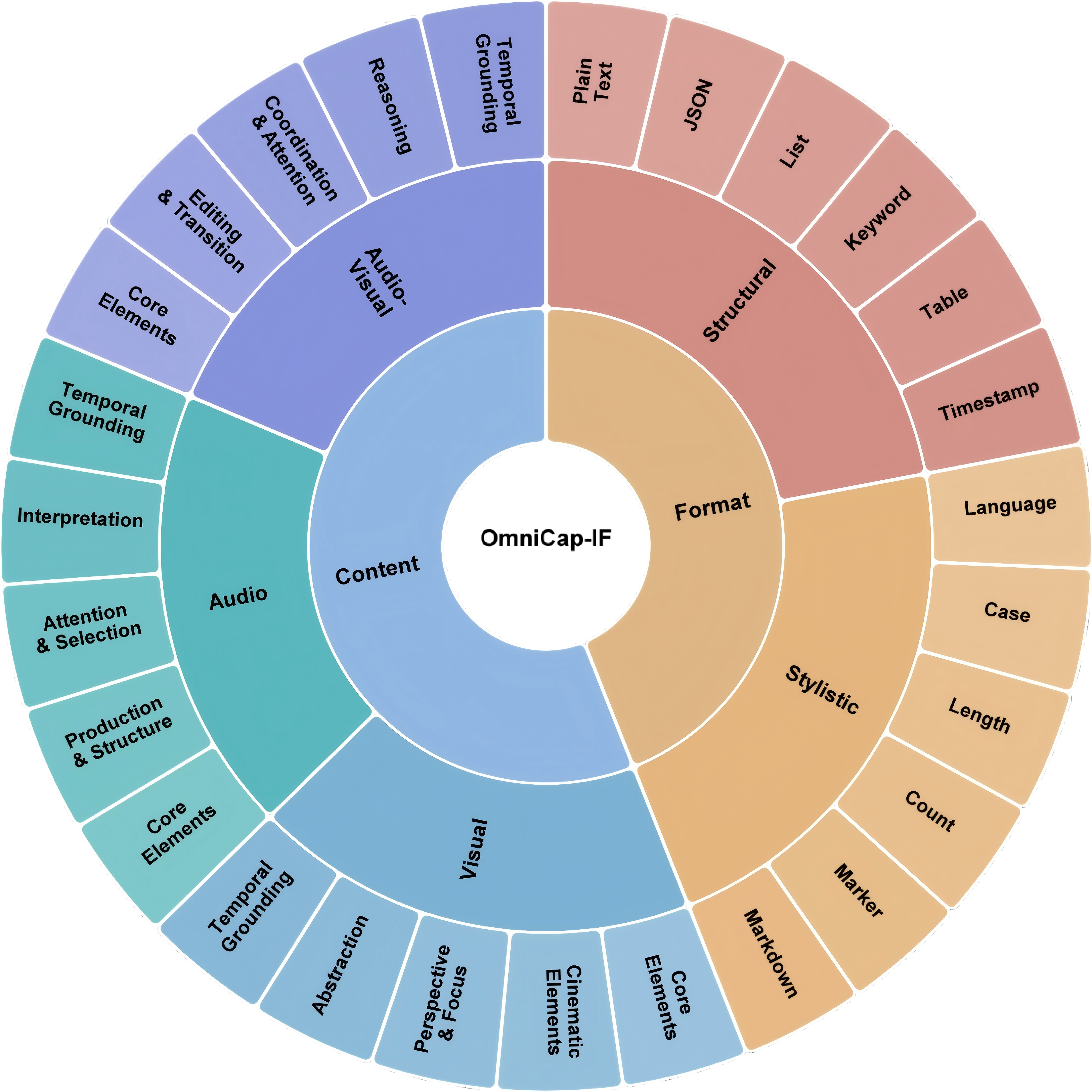}
            \caption{Overview of OmniCap-IF constraint categories.}
            \label{fig:dist_sunburst}
        \end{subfigure}
        
    \end{minipage}

    \caption{Dataset statistics for OmniCap-IF. (a-c) show distributions for video duration, category, and constraint count, respectively. (d) provides an overview of the constraint categories.}
    \label{fig:overall_statistical}
\end{figure*}

Statistical analysis underscores the comprehensive nature of OmniCap-IF, highlighting its substantial diversity in duration, content coverage, and instructional complexity (Figure~\ref{fig:overall_statistical}). The dataset exhibits a well-balanced distribution of video durations, with its average duration exceeding most existing omni-modal captioning benchmarks (Figure~\ref{fig:dist_duration}). In addition, its wide-ranging content, covering numerous categories (Figure~\ref{fig:dist_category}), supports evaluation of cross-domain generalization. The instruction set further spans a spectrum from standard prompts to highly complex cases (Figure~\ref{fig:dist_constraint}). Collectively, these properties position OmniCap-IF as a next-generation testbed for evaluating OLLMs.

\subsubsection{Comparison with Other Benchmarks.} 
When compared with other benchmarks, we adopt IFEval \citep{zhou2023ifeval}, CELLO \citep{he2024cello}, InfoBench \citep{qin2024infobench}, FollowBench \citep{jiang2024followbench}, SysBench \citep{qin2024sysbench}, \mbox{CFBench} \citep{zhang2025cfbench}, ComplexBench \citep{wen2024complex} and IF-VidCap \citep{li2026ifvidcap} as the instruction-following baselines. For omni-modal captioning, we compare against UGC-VideoCap~\citep{wu2025ugc}, Omni-Cloze~\citep{ma2026omnicaptioner}, OmniDCBench~\citep{yao2026timechat}, and video-SALMONN-2-testset~\citep{tang2025salmonn2}. As shown in Table~\ref{tab:comparison}, OmniCap-IF advances the landscape of both instruction-following and omni-modal captioning benchmarks. In contrast to prior datasets that focus solely on text-only or vision-only instruction following, it incorporates omni-modal inputs while achieving a larger scale, increased instructional complexity, and more comprehensive content coverage. From the perspective of omni-modal captioning, OmniCap-IF shifts the focus from conventional descriptive or holistic narratives toward fine-grained instruction adherence, featuring richer informational content and, in general, longer video durations than most existing benchmarks. By bridging these directions and further introducing temporal grounding mechanisms, OmniCap-IF establishes a more rigorous and versatile benchmark for evaluating controllable generation in OLLMs, facilitating progress in the diverse downstream applications detailed in Appendix~\ref{sec:app}.

\begin{table}[t]
\centering
\small  
\setlength{\tabcolsep}{5pt} 
\caption{Comparison of Instruction Following and Omni-modal Captioning Benchmarks. ``\#Size'', ``\#Types'', and ``\#Const.'' denote the total number of prompts, the number of distinct constraint types, and the average number of constraints per instruction, respectively. ``Vid. Len.'' refers to the average video duration. ``Temporal'' indicate whether the benchmark supports temporal grounding constraints. ``Mod.'' indicates the input modality (T: Text, V: Video, AV: Audio-Visual), while ``Evaluation'' specifies the methodology used for scoring.}
\label{tab:comparison}
\begin{tabular}{lccccccc}
\toprule
\textbf{Benchmark} & \textbf{\#Size} & \textbf{\#Types} & \textbf{\#Const.} & \textbf{Vid. Len.} & \textbf{Temporal} & \textbf{Mod.} & \textbf{Evaluation} \\ \midrule
\multicolumn{8}{c}{\textit{Instruction Following Benchmarks}} \\ \midrule
IFEval & 541 & 25 & 1.54 & -- & -- & T & Rule \\
CELLO & 523 & 4 & 2.18 & -- & -- & T & Rule \\
InfoBench & 500 & 5 & 5.93 & -- & -- & T & LLM \\
FollowBench & 944 & 5 & 3.00 & -- & -- & T & LLM / Rule \\
SysBench & 500 & 6 & 2.38 & -- & -- & T & LLM \\
CFBench & 1,000 & 10-25 & 4.24 & -- & -- & T & LLM \\
ComplexBench & 1,150 & 4-19 & 4.61 & -- & -- & T & LLM+Rule \\
IF-VidCap & 1,400 & 27 & 6.00 & 20.5s & -- & V & LLM+Rule \\ \midrule
\multicolumn{8}{c}{\textit{Omni-modal Captioning Benchmarks}} \\ \midrule
video-SALMONN-2-testset & 483 & -- & -- & 50.8s & -- & AV & LLM \\
UGC-VideoCap & 1,000 & -- & -- & 23.9s & -- & AV & LLM \\
Omni-Cloze & 2,340 & -- & -- & 34.2s & -- & AV & LLM \\
OmniDCBench & 1,122 & -- & -- & 59.5s & \checkmark & AV & LLM \\ \midrule
\textbf{OmniCap-IF (Ours)} & \textbf{1,920} & \textbf{50} & \textbf{6.93} & \textbf{54.6s} & \checkmark & \textbf{AV} & \textbf{LLM+Rule} \\ \bottomrule
\end{tabular}
\end{table}

\subsection{Evaluation Protocol}

\subsubsection{Evaluation Methodology}
To rigorously assess model performance, OmniCap-IF employs a bifurcated evaluation strategy that disentangles structural adherence from semantic fidelity, as comprehensively illustrated in Figure~\ref{fig:teaser}. Inspired by IF-VidCap \citep{li2026ifvidcap}, we incorporate rule-based programmatic tools into our evaluation pipeline to significantly enhance the stability and reliability of the LLM-as-a-judge \citep{zheng2023judging}. 

\noindent\textbf{Format Evaluation:} 
This component targets objective structural requirements (e.g., length, JSON schema, or ordered lists). To ensure stable and robust verification, we employ a two-step hybrid approach: an LLM first extracts the structured information from the generated output, followed by the execution of rule-based programmatic tools to deterministically verify compliance against predefined formatting rules.
\noindent \textbf{Content Evaluation:} 
This component assesses instruction following regarding content constraints, explicitly prioritizing objective factual accuracy over descriptive fluency to mitigate LLM judge biases. We evaluate this through two complementary mechanisms:
\begin{itemize}[leftmargin=*, noitemsep, topsep=2pt]
    \item \textit{Temporal Grounding Constraints:} An LLM extracts timestamps from the response, followed by rule-based tools computing temporal-IoU (t-IoU) or offsets to accurately determine temporal compliance. Comprehensive descriptions of the evaluation procedures are provided in Appendix~\ref{sec:timestamp_eval}.
    \item \textit{Multimodal Content Constraints:} For the remaining constraints across visual, audio, and audio-visual dimensions, we leverage an LLM-as-a-judge via a Question-Answering (QA) approach. The evaluation uses binary and multiple-choice questions. By providing generated captions as context, we verify the factual alignment between the content and complex instructions.
\end{itemize}

\noindent The prompts for format extraction and content evaluation are provided in Appendix~\ref{sec:judge_prompt}.

\subsubsection{Evaluation Metrics}
We employ two primary metrics to quantify performance: \textbf{Constraint Satisfaction Rate (CSR)} and \textbf{Instruction Satisfaction Rate (ISR)}.
\begin{equation}
\text{CSR} = \frac{1}{m} \sum_{i=1}^{m} \frac{1}{n_i} \sum_{j=1}^{n_i} s_i^j, \quad \quad \text{ISR} = \frac{1}{m} \sum_{i=1}^{m} \text{ISR}_i
\end{equation}
where $m$ is the total number of instructions, and $n_i$ denotes the number of constraints for the $i$-th instruction. $s_i^j \in \{0, 1\}$ indicates whether the $j$-th constraint is satisfied. $\text{ISR}_i$ is a binary indicator that equals 1 if and only if all constraints within the $i$-th instruction are simultaneously satisfied (i.e., $\sum_{j=1}^{n_i} s_i^j = n_i$), and 0 otherwise.

To provide a granular diagnostic of model capabilities, we report metrics across a hierarchical structure as categorized in Table~\ref{tab:main_results}:
\begin{itemize}[leftmargin=*, noitemsep, topsep=2pt]
    \item \textbf{Primary Evaluation Types:} We report Format CSR/ISR for structural control and Content CSR/ISR for semantic fidelity.
    \item \textbf{Modality-Specific Content Analysis:} The Content CSR is further decomposed into three distinct dimensions—Visual, Audio, and Audio-Visual (AV)—to precisely pinpoint the modality-specific instruction-following capabilities of various models.
\end{itemize}





\subsection{OmniCap-IF-54K}
\begin{figure}[htbp]
    \centering
    \includegraphics[width=0.8\textwidth]{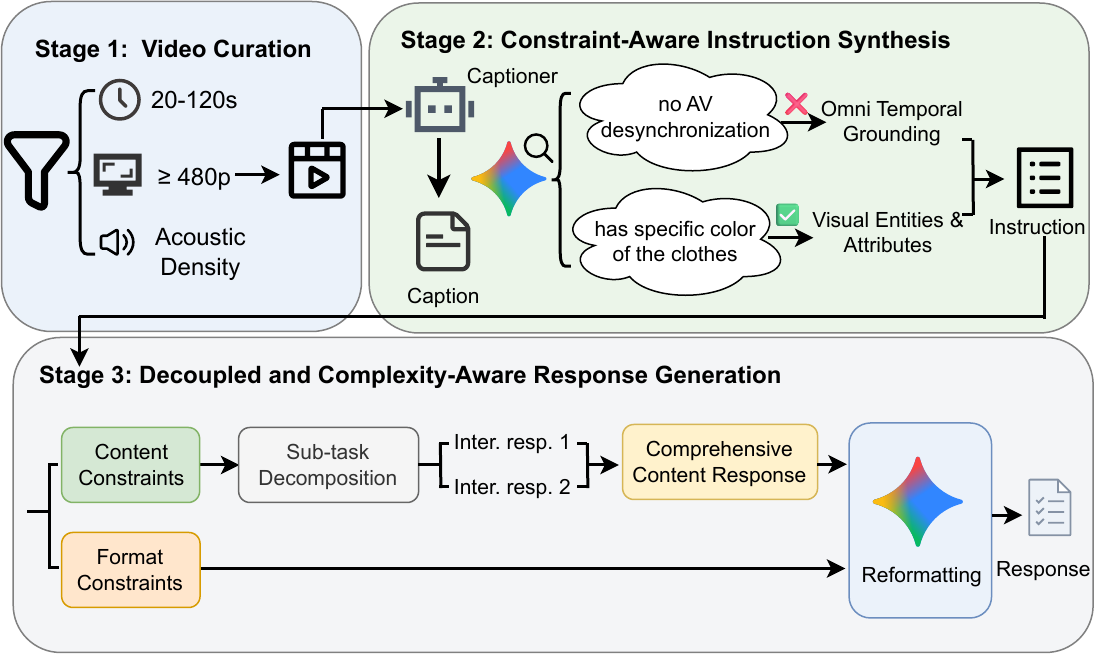} 
    \caption{The training set generation pipeline. ``Inter. resp.'' stands for Intermediate Response.}
    \label{fig:trainset}
\end{figure}

To endow models with generalizable and robust instruction-following capabilities, we introduce a large-scale, high-quality fine-tuning dataset. To prevent data leakage, the generation pipeline is strictly decoupled from our evaluation benchmark. As illustrated in Figure~\ref{fig:trainset}, the process consists of three stages, ultimately yielding OmniCap-IF-54K, which comprises 54K meticulously curated video-instruction-response triplets.

\noindent \textbf{Stage 1: High-Quality Omni-Modal Video Curation.} 
We source raw videos from LLaVA-Video-178K~\citep{zhang2024llavavideo} and TikTok-10M~\citep{tiktok_10m_2025}. To ensure multimodal richness, we apply strict heuristic filters: (1) durations between 20 to 120 seconds, (2) visual resolutions of at least 480p, and (3) high acoustic density, filtered using PANNs~\citep{kong2020panns} to guarantee the presence of diverse ambient sounds and speech. This results in 14K high-quality video samples.

\noindent \textbf{Stage 2: Constraint-Aware Instruction Synthesis.} 
We first generate fine-grained textual captions for all videos using ASID-Captioner-7B~\citep{li2026asid} to serve as dense multimodal proxies. Gemini-3-Flash~\citep{Google2026Gemini3} then synthesizes instructions by sampling from our constraint system. Crucially, we implement a negative constraint filter: any constraint whose prerequisite elements are absent from the proxy caption (e.g., blacklisting the ``omni temporal grounding'' constraint if the caption lacks any description of audio-visual desynchronization) is excluded to prevent hallucinations. Valid constraints are then combined to form complex, multi-constraint instructions.

\noindent \textbf{Stage 3: Decoupled and Complexity-Aware Response Generation.} 
As demonstrated in Figure~\ref{fig:complexity_overall}, a model's ability to satisfy constraints degrades significantly as the number of constraints increases. Consequently, generating a response for a heavily constrained instruction in a single pass often yields flawed ground truth. To circumvent this, we adopt a decomposed generation strategy. First, we separate the instruction into content constraints and format constraints. The content constraints are further partitioned into smaller, manageable sub-tasks containing only 2--3 constraints each. Based on the video caption, Gemini-3-Flash generates high-fidelity intermediate responses for these sub-tasks, which are then aggregated into a comprehensive, multi-constraint content response. Furthermore, as illustrated in Figure~\ref{fig:format_tradeoff}, enforcing rigid format constraints simultaneously can severely compromise the factual correctness of the generated content. Thus, we apply format constraints exclusively in the final stage: the model is instructed to reformat the aggregated content response to produce the final ground truth, ensuring both semantic richness and structural compliance. In addition, we conduct a study on 500 randomly sampled triplets by comparing our decomposed strategy with direct generation. Using the same checklist-based evaluation, we find that in 96.3\% of cases, the decomposed-and-aggregated approach yields superior results to direct generation.

\noindent The prompts used in the process and training details are provided in Appendix~\cref{sec:train_set_prompt,sec:train}.

\section{Experiments}
\subsection{Main Results}
We evaluate 14 leading omni-modal models, including Gemini-3.1-Pro~\citep{Google2026Gemini3}, Gemini-3-Flash, MiMo-V2.5~\citep{Xiaomi2026MimoV25}, MiMo-V2-Omni~\citep{Xiaomi2026MimoV2Omni}, Qwen3-Omni~\citep{xu2025qwen3omnitechnicalreport}, Qwen2.5-Omni~\citep{xu2025qwen25omnitechnicalreport}, ARC-Hunyuan-Video~\citep{ge2025arc}, HumanOmniV2~\citep{yang2025humanomniv2}, MiniCPM-o~\citep{yao2024minicpm}, video-SALMONN2~\citep{tang2025salmonn2} and ASID-Captioner~\citep{li2026asid}. The system prompts and evaluation settings used for the models are detailed in Appendix~\ref{sec:system_prompt} and Appendix~\ref{sec:eval}.

The main results in Table~\ref{tab:main_results} yield several key observations: (1)~In the same model family, performance consistently improves as model size increases. (2)~Generally, models perform better on Audio and Visual constraints independently than on Audio-Visual constraints, highlighting the difficulty of joint audio-visual integration. (3)~Models demonstrate stronger capability in format control than in adhering to content-related requirements, likely because content understanding demands more complex multi-modal reasoning, whereas format constraints are predominantly text-based. (4)~The human baseline exhibits a distinct performance pattern compared to advanced models. Benefiting from deliberate verification and self-reflective reasoning, human annotators achieve better results in format control, significantly outperforming all evaluated models.

Additionally, we develop the \textbf{OmniCaptioner-IF} series by fine-tuning Qwen2.5-Omni on OmniCap-IF-54K. OmniCaptioner-IF outperforms the base model across all metrics. Notably, it exhibits strong structural controllability, performing on par with the proprietary Gemini-3.1-Pro in format metrics, highlighting the effectiveness of our instruction-tuning for enforcing rigid constraints. This gain likely stems from format control relying on low-level textual signals that are easier to learn from limited supervision, whereas content constraints require more complex multimodal reasoning.


\begin{table}[t]
\renewcommand\arraystretch{1.1} 
\setlength{\fboxsep}{1.5pt} 
\newcommand{\best}[1]{\colorbox{cyan!25}{\strut #1}}
\caption{Main Evaluation Results on the OmniCap-IF Benchmark. The content CSR is further decomposed into Visual, Audio, and Audio-Visual modalities.}
\label{tab:main_results}

\setlength{\tabcolsep}{4pt} 

\resizebox{\textwidth}{!}{
\begin{tabular}{l c cc c cc c ccccc}
\toprule
\multirow{2}{*}{\textbf{Model}} & 
\phantom{spacer} & \multicolumn{2}{c}{\textbf{Overall}} & 
\phantom{spacer} & \multicolumn{2}{c}{\textbf{Format}} & 
\phantom{spacer} & \multicolumn{5}{c}{\textbf{Content}} \\

\cmidrule(lr){3-4} \cmidrule(lr){6-7} \cmidrule(lr){9-13} 

 & & \textbf{CSR} & \textbf{ISR} & & \textbf{CSR} & \textbf{ISR} & & \textbf{CSR} & \textbf{ISR} & \makecell[c]{\textbf{Visual} \\ \textbf{CSR}} & \makecell[c]{\textbf{Audio} \\ \textbf{CSR}} & \makecell[c]{\textbf{AV} \\ \textbf{CSR}} \\
\midrule

Human  & & 83.29 & 35.31 & & 94.83 & 84.19 & & 78.23 & 40.19 & 78.38 & 80.05 & 72.43 \\

\midrule
\multicolumn{13}{c}{\textit{Closed-Source Large Multimodal Models}} \\
\midrule
Gemini-3.1-Pro  & & 80.65 & 25.82 & & 90.45 & 78.65 & & 75.02 & 32.45 & 74.15 & 77.45 & 73.40 \\
Gemini-3-Flash  & & 79.50 & 23.55 & & 88.57 & 74.15 & & 74.29 & 31.17 & 73.60 & 76.63 & 72.35 \\
MiMo-V2.5    & & 76.22 & 20.50 & & 86.40 & 71.81 & & 70.37 & 26.75 & 69.82 & 74.73 & 67.68 \\
MiMo-V2-Omni  & & 74.40 & 17.21 & & 80.60 & 62.04 & & 70.84 & 26.43 & 70.14 & 73.51 & 68.95 \\

\midrule
\multicolumn{13}{c}{\textit{Open-Source Large Multimodal Models}} \\
\midrule
Qwen3-Omni-30B-A3B-Thinking & & 71.91 & 14.27 & & 84.29 & 67.34 & & 64.79 & 19.90 & 65.63 & 69.08 & 61.58 \\
MiniCPM-o-4.5-9B        & & 64.69 & 9.27 & & 78.60 & 56.04 & & 56.70 & 13.07 & 59.24 & 62.64 & 51.86 \\
Qwen3-Omni-30B-A3B-Instruct & & 62.65 & 7.24 & & 77.37 & 54.64 & & 54.19 & 10.83 & 58.13 & 59.92 & 49.31 \\
Qwen2.5-Omni-7B         & & 49.19 & 2.34  & & 62.97 & 34.17 & & 41.27 & 4.53 & 47.68 & 47.51 & 34.88 \\
MiniCPM-o-2.6-8B        & & 47.38 & 1.88  & & 62.31 & 31.46 & & 38.81 & 3.75  & 46.78 & 44.45 & 32.28 \\
Qwen2.5-Omni-3B         & & 40.13 & 0.78  & & 52.49 & 22.97 & & 33.02 & 2.14  & 41.55 & 38.16 & 26.62 \\
HumanOmniV2-7B          & & 32.95 & 0.60  & & 32.32 & 11.04 & & 33.31 & 3.19  & 42.34 & 36.38 & 28.30 \\
video-SALMONN-2-7B      & & 32.80 & 0.42  & & 41.09 & 13.80 & & 28.03 & 1.25  & 34.27 & 33.74 & 22.10 \\
ARC-Hunyuan-Video-7B    & & 29.74 & 0.31  & & 20.27 & 5.75  & & 34.71 & 4.17  & 44.51 & 37.24 & 26.62 \\
ASID-Captioner-7B       & & 24.52 & 0.47  & & 17.50 & 4.43  & & 28.56 & 2.76  & 39.49 & 32.71 & 23.64 \\

\midrule
\multicolumn{13}{c}{\textit{Ours}} \\
\midrule
\textbf{OmniCaptioner-IF-7B (ours)} & & \textbf{70.73} & \textbf{11.46} & & \textbf{90.39} & \textbf{77.92} & & \textbf{59.43} & \textbf{13.59} & \textbf{58.71} & \textbf{64.71} & \textbf{55.62} \\
\textbf{OmniCaptioner-IF-3B (ours)} & & \textbf{66.67} & \textbf{7.86} & & \textbf{87.73} & \textbf{73.12} & & \textbf{54.57} & \textbf{9.79} & \textbf{55.91} & \textbf{60.39} & \textbf{50.06} \\
\bottomrule
\end{tabular} 
}
\end{table}

\subsection{Results on Existing Benchmarks}
We evaluate OmniCaptioner-IF on several external benchmarks—IF-VidCap (vision-only instruction following), Omni-Cloze (cloze-style fine-grained omni perception), and UGC-VideoCap (QA-based omni video captioning)—to comprehensively assess its generalizable omni-modal perception capabilities. The results of Omni-VideoQA benchmarks can be found in the Appendix~\ref{sec:omni_videoqa}.

\noindent \textbf{Results on IF-VidCap.} 
IF-VidCap strictly focuses on visual-only instruction adherence. We evaluate our model by providing only the video track. As shown in Table~\ref{tab:if_vidcap}, OmniCaptioner-IF-3B surpasses the vision-expert model Qwen2.5-VL-Instruct-3B across all metrics. This demonstrates that our omni-modal instruction tuning not only preserves but enhances pure visual grounding capabilities.

\begin{table}[h]
\caption{Results on the IF-VidCap Benchmark.}
\label{tab:if_vidcap}
\centering
\begin{tabular}{lcccccc}
\toprule
\multirow{2}{*}{\textbf{Model}} & \multirow{2}{*}{\textbf{CSR}} & \multirow{2}{*}{\textbf{ISR}} & \multicolumn{2}{c}{\textbf{Format}} & \multicolumn{2}{c}{\textbf{Content}} \\
\cmidrule(lr){4-5} \cmidrule(lr){6-7}
 &  &  & \textbf{CSR} & \textbf{ISR} & \textbf{CSR} & \textbf{ISR} \\
\midrule
Gemini-2.5-Pro~\citep{Google2025Gemini25Pro} & 74.53 & 27.83 & 87.81 & 74.35 & 59.00 & 35.22 \\
Qwen2.5-VL-Instruct-7B~\citep{Qwen2.5-VL} & 58.12 & 10.92 & 73.81 & 52.51 & 39.65 & 18.75 \\
Qwen2.5-VL-Instruct-3B~\citep{Qwen2.5-VL} & 51.74 & 6.54 & 66.50 & 43.46 & 34.47 & 13.15 \\
\midrule
Qwen2.5-Omni-7B (w/o Audio) & 56.49 & 8.17 & 74.41 & 54.12 & 36.76 & 14.04 \\
Qwen2.5-Omni-3B (w/o Audio) & 49.66 & 5.73 & 65.77 & 43.23 & 31.95 & 11.10 \\
\textbf{OmniCaptioner-IF-7B (w/o Audio)} & \textbf{61.20} & \textbf{12.21} & \textbf{79.92} & \textbf{61.33} & \textbf{40.63} & \textbf{16.57} \\
\textbf{OmniCaptioner-IF-3B (w/o Audio)} & \textbf{57.56} & \textbf{8.70} & \textbf{76.30} & \textbf{57.58} & \textbf{36.99} & \textbf{13.70} \\
\bottomrule
\end{tabular}
\end{table}

\noindent \textbf{Results on Omni-modal Captioning Benchmarks.}
We further validate our model on comprehensive audio-visual benchmarks. As illustrated in Table~\ref{tab:omnicloze_results}, OmniCaptioner-IF-7B demonstrates a remarkable performance leap on Omni-Cloze, effectively doubling the total accuracy compared to the original baseline.

\begin{table}[h]
\caption{Results on the Omni-Cloze Benchmark.}
\label{tab:omnicloze_results}
\centering
\begin{tabular}{lcccc}
\toprule
\textbf{Model} & \textbf{Visual\%$\uparrow$} & \textbf{Audio\%$\uparrow$} & \textbf{AV\%$\uparrow$} & \textbf{Total\%$\uparrow$} \\
\midrule
Gemini-2.5-Flash~\citep{comanici2025gemini25} & 31.50 & 18.40 & 39.10 & 27.90 \\
video-SALMONN-13B~\citep{sun2024salmonn} & 2.60 & 1.70 & 4.00 & 2.50 \\
VideoLLaMA-2-7B~\citep{cheng2024videollama2} & 5.70 & 2.60 & 7.30 & 4.80 \\
\midrule
Qwen2.5-Omni-7B & 10.40 & 12.90 & 18.90 & 12.90 \\
\textbf{OmniCaptioner-IF-7B (Ours)} & \textbf{23.86} & \textbf{24.23} & \textbf{32.30} & \textbf{25.17} \\
\textbf{OmniCaptioner-IF-3B (Ours)} & \textbf{21.27} & \textbf{21.81} & \textbf{28.94} & \textbf{22.53} \\
\bottomrule
\end{tabular}
\end{table}
\noindent On the UGC-VideoCap benchmark (Table~\ref{tab:ugc_results}), OmniCaptioner-IF-7B achieves performance comparable to Gemini-2.5-Pro. This highlights the efficacy of fine-grained constraint adherence as a proxy for enhancing general omni-modal understanding. 

\begin{table}[h]
\caption{Results on the UGC-VideoCap Benchmark.}
\label{tab:ugc_results}
\centering
\begin{tabular}{lcccc}
\toprule
\textbf{Model} & \textbf{Audio$\uparrow$} & \textbf{Visual$\uparrow$} & \textbf{Detail$\uparrow$} & \textbf{Avg.$\uparrow$} \\
\midrule
Gemini-2.5-Pro & 69.50 & 74.70 & 73.70 & 72.60 \\
Qwen3-Omni-30B-A3B-Instruct & 67.50 & 74.80 & 72.30 & 71.50 \\
HumanOmniV2-7B & 45.60 & 66.30 & 59.50 & 57.10 \\
video-SALMONN-2-7B & 61.80 & 71.40 & 68.50 & 67.20 \\
\midrule
Qwen2.5-Omni-7B & 46.90 & 66.10 & 60.00 & 57.70 \\
Qwen2.5-Omni-3B & 48.20 & 55.60 & 52.60 & 52.18 \\
\textbf{OmniCaptioner-IF-7B (Ours)} & \textbf{69.79} & \textbf{75.94} & \textbf{73.19} & \textbf{72.97} \\
\textbf{OmniCaptioner-IF-3B (Ours)} & \textbf{67.71} & \textbf{73.91} & \textbf{70.43} & \textbf{70.68} \\
\bottomrule
\end{tabular}
\end{table}
\subsection{Further Analysis}

\begin{figure*}[t] 
    \centering
    
    \begin{subfigure}[b]{0.48\textwidth}
        \centering
        \includegraphics[width=\textwidth]{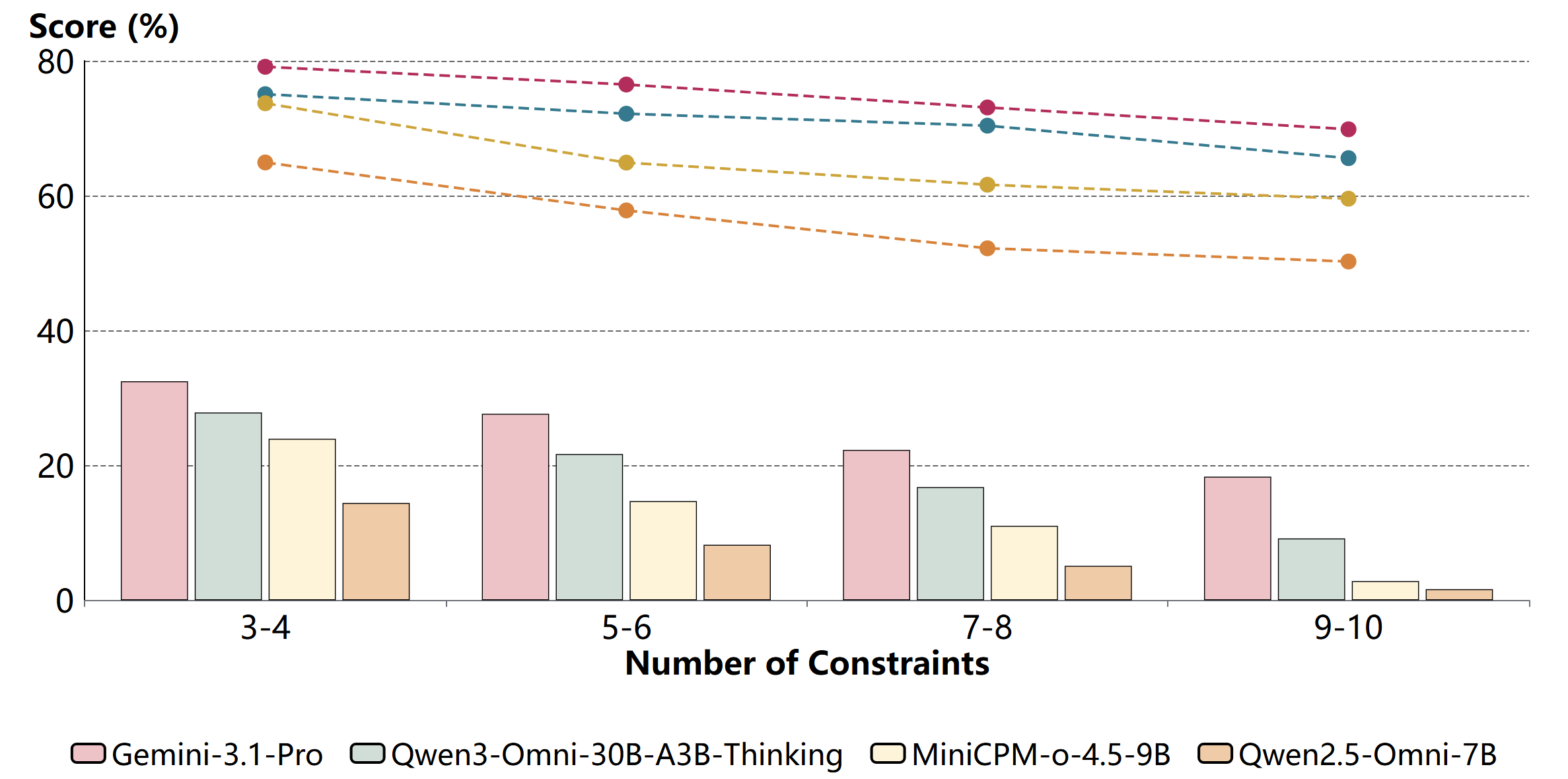}
        \caption{Constraint count.}
        \label{fig:complexity_a}
    \end{subfigure}
    \hfill 
    \begin{subfigure}[b]{0.48\textwidth}
        \centering
        \includegraphics[width=\textwidth]{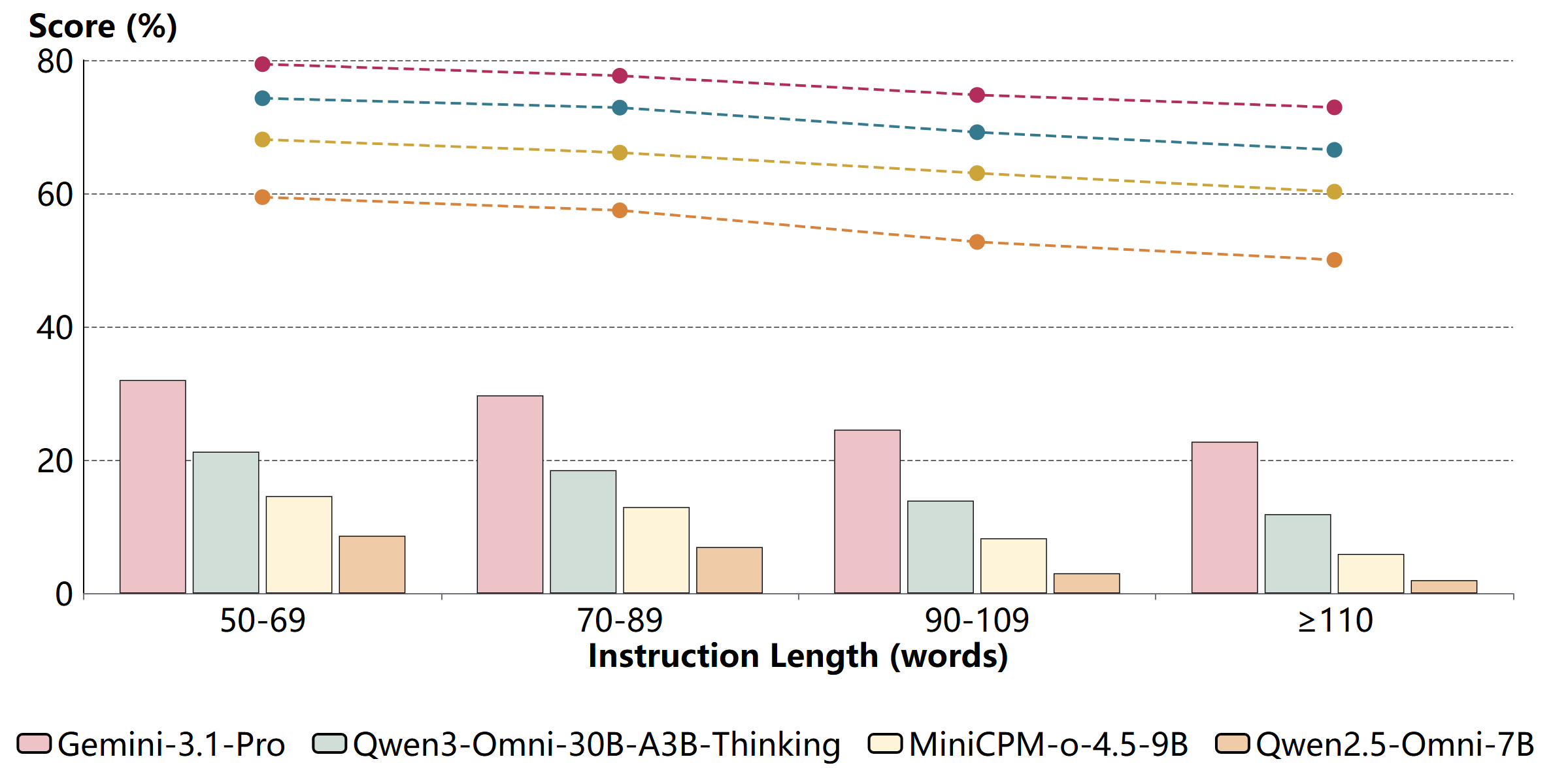}
        \caption{Instruction length.}
        \label{fig:complexity_b}
    \end{subfigure}
    
    \caption{The impact of constraint count, instruction length on model performance.}
    \label{fig:complexity_overall}
\end{figure*}

\noindent \textbf{Impact of Instruction Complexity.}
We examine four representative models to explore the relationship between instruction complexity---jointly determined by prompt length and constraint count---and two metrics: Constraint Satisfaction Rate (CSR) and Instruction Satisfaction Rate (ISR). Evaluations are carried out on an expert-filtered subset comprising 1,000 high-quality instances from the benchmark. Within these samples, constraints are interdependent (e.g., branching, chaining) instead of being strictly isolated. This specific selection guarantees that the total number of constraints reliably reflects the true complexity of the task. Figures~\ref{fig:complexity_a} and \ref{fig:complexity_b} explicitly illustrate that a model's proficiency in satisfying constraints and following directives deteriorates with escalating complexity, substantiating that more nuanced and difficult commands severely strain models' instruction-following abilities.

\begin{figure}[htbp]
    \centering
    \includegraphics[width=0.6\textwidth]{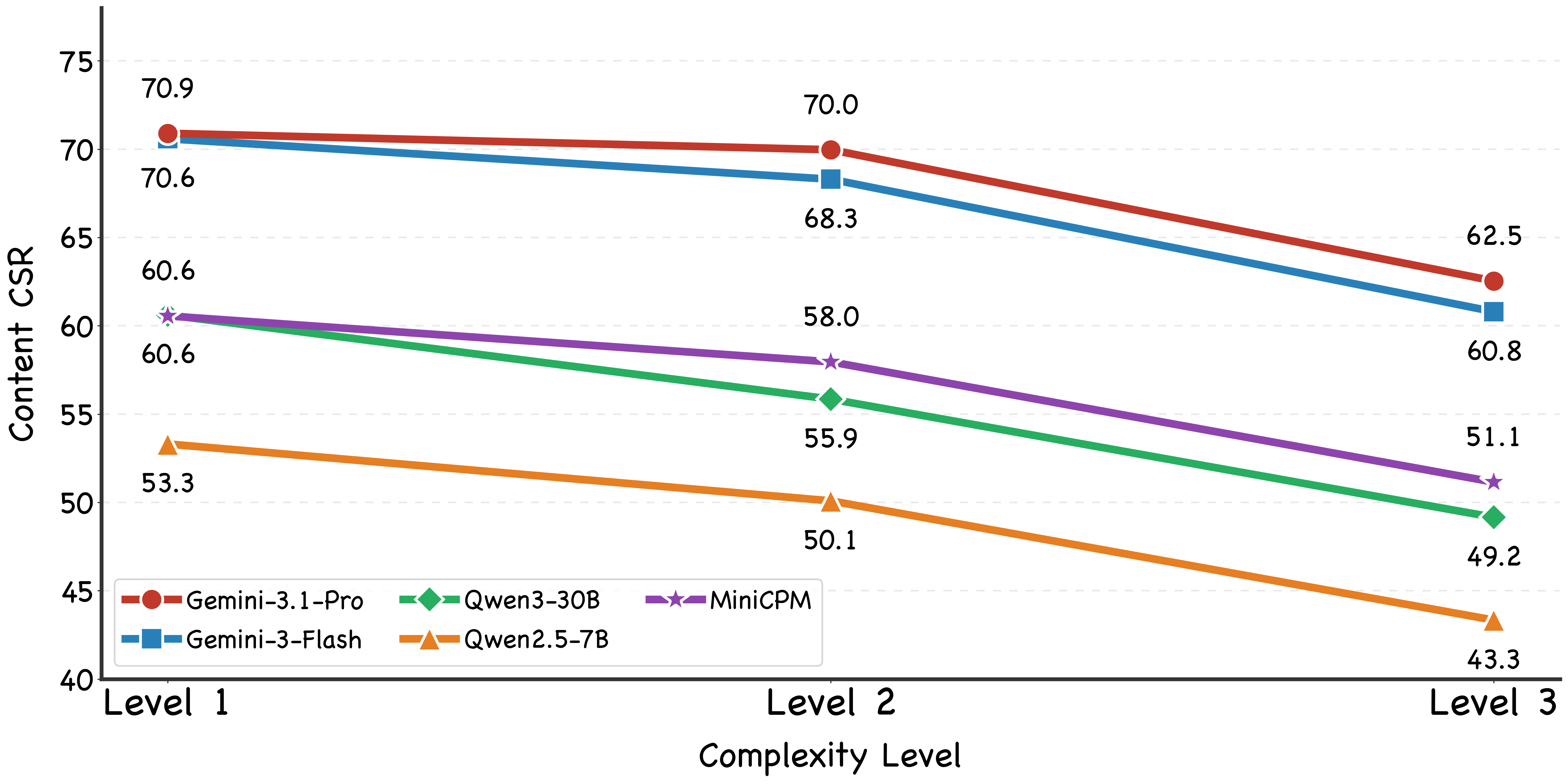} 
    \caption{The format-content tradeoff.}
    \label{fig:format_tradeoff}
\end{figure}

\noindent \textbf{Format-content Tradeoff.}\label{tradeoff}
To examine the impact of formatting complexity on a model's ability to retain semantic depth, we designed a controlled experiment evaluated on 1,200 curated samples across five representative models. Specifically, we held the content constraints strictly constant while varying the format constraints across three levels:
\begin{itemize}[leftmargin=*, topsep=0pt, itemsep=0pt, parsep=0pt]
    \item \noindent \textbf{Level 1 (Loose):} Natural language, basic paragraphs/bullets (e.g., plain text, length).
    \item \noindent \textbf{Level 2 (Styled):} Human-readable visual structuring requiring layout awareness (e.g., Markdown table, ordered lists).
    \item \noindent \textbf{Level 3 (Syntactic):} Machine-readable, strict grammatical rules (e.g., JSON arrays, forced keywords).
\end{itemize}
As illustrated in Figure~\ref{fig:format_tradeoff}, as the formatting level increases from the lowest level to the highest level, the content CSR drops continuously and noticeably. This indicates that forcing models to allocate attention to rigid syntactic generation (e.g., JSON nesting) directly cannibalizes their capacity for complex cross-modal reasoning.



\begin{figure}[htbp] 
\centering
\includegraphics[width=0.6\textwidth]{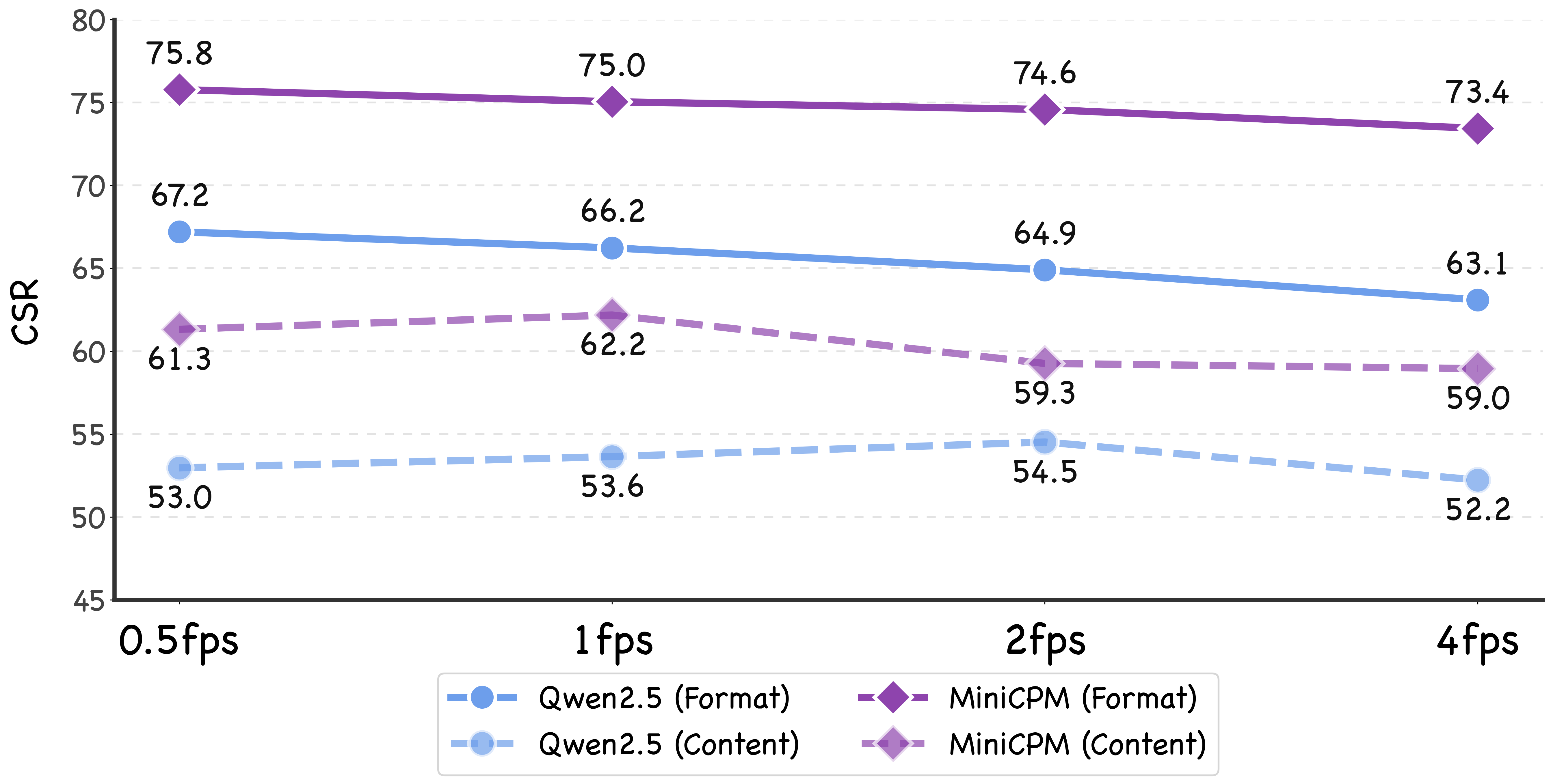} 
\caption{Impact of FPS on model performance.}
\label{fig:fps_results}
\end{figure}

\noindent \textbf{Impact of Video Parameters.} 
We examine Qwen2.5-Omni-7B and MiniCPM-o-4.5-9B under varying frame sampling rates (FPS). As shown in Figure~\ref{fig:fps_results}, increasing FPS causes Format CSR to drop steadily, as more visual tokens overwhelms the context window and reduces the models’ ability to maintain strict structural adherence. Content CSR first increases and then decreases. The initial gain arises from richer visual evidence supporting fine-grained event perception, while excessive frame density adds redundant noise and context pressure, deviating from the models’ optimal training distributions and impairing abilities such as precise temporal grounding. The exact turning point varies across models, reflecting differences in their preferred visual token density.


\begin{figure}[htbp] 
\centering
\includegraphics[width=0.6\textwidth]{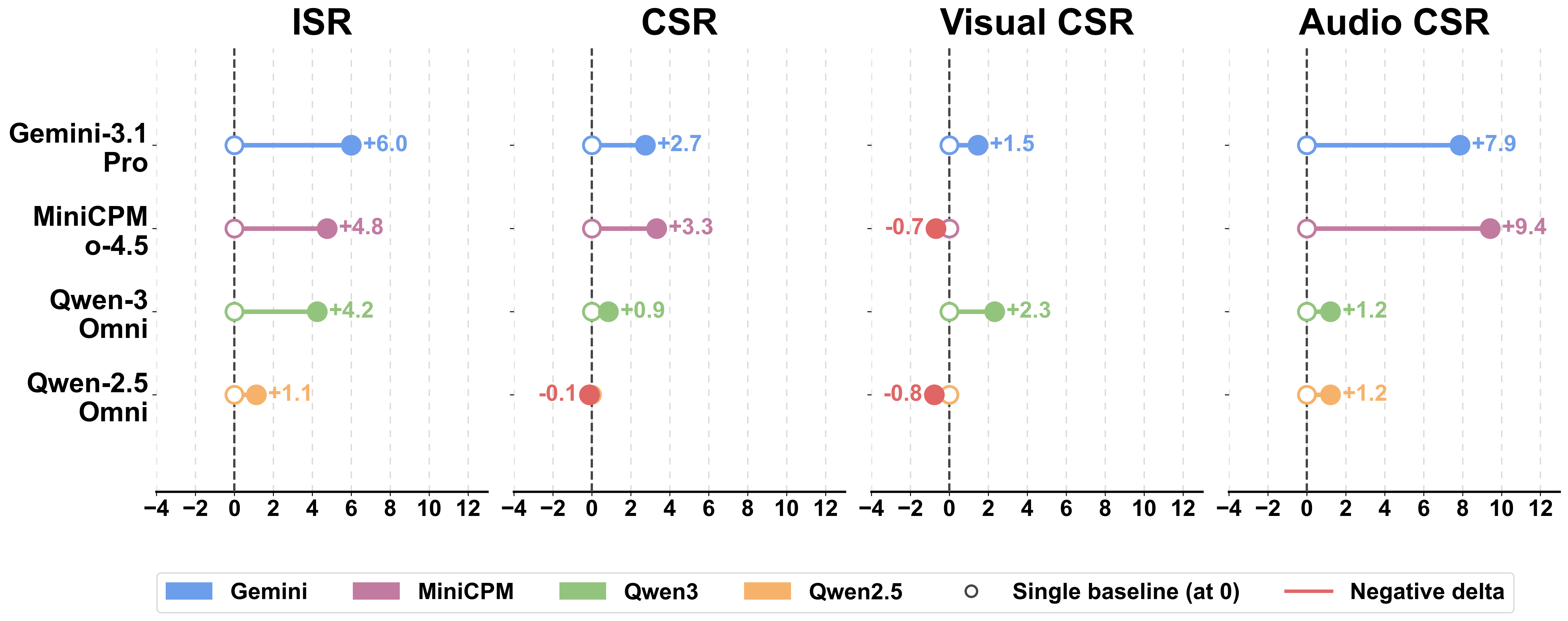} 
\caption{Analysis of Cross-Modal Synergy.}
\label{fig:modality_synergy}
\end{figure}

\noindent \textbf{Analysis of Cross-Modal Synergy.} 
To assess whether current OLLMs achieve true audio-visual synergy, we perform a modality decoupling experiment. We derive uni-modal instructions by retaining only constraints of the original prompts that can be resolved solely through visual or auditory evidence (while preserving all format constraints). These instructions are then paired with their corresponding single-modal inputs and compared against the full omni-modal setting (Figure~\ref{fig:modality_synergy}). Gemini-3.1-Pro and MiniCPM-o-4.5 exhibit strong cross-modal gains: adding visual context significantly boosts their Audio CSR, showing effective use of visual cues to ground acoustic events. In contrast, the Qwen series shows minimal synergy. Qwen3-Omni and Qwen2.5-Omni achieve only slight improvements, with Qwen2.5-Omni even declining in Overall CSR. Degraded Visual CSR in MiniCPM-o-4.5 and Qwen2.5-Omni further highlights cross-modal interference, suggesting that while these models handle uni-modal inputs well, they largely process audio and visual streams independently rather than in a deeply fused manner.

\noindent \textbf{Agreement Evaluation.}
To validate our evaluation framework, we compare automated assessments with human judgment using the professional annotations described in Section~\ref{annotation}. Agreement is measured across three assessor models: GPT-5-mini \citep{singh2025gpt5}, Gemini-3-Flash \citep{Google2026Gemini3}, and Qwen3.5-27B \citep{qwen3.5blog}. As shown in Table~\ref{tab:agreement}, GPT-5-mini achieves the highest concordance with human evaluations across all metrics. The strong agreement across these diverse models highlights the robustness and general applicability of our methodology.


\begin{table}[h]
\centering
\caption{Agreement between automated evaluation and human evaluation across different models.}
\label{tab:agreement}
\begin{tabular}{lccc}
\toprule
\textbf{Model} & \textbf{Overall Agreement} & \textbf{Format} & \textbf{Content} \\
\midrule
GPT-5-mini & 94.70 & 96.12 & 94.29 \\
Gemini-3-Flash & 93.16 & 94.17 & 92.86 \\
Qwen3.5-27B & 92.49 & 94.66 & 91.86 \\
\bottomrule
\end{tabular}
\end{table}

\noindent \textbf{Constraint Type Analysis.}
Our analysis of the CSR across representative models (Figure \ref{fig:constraint_type_csr}) reveals a pervasive performance bottleneck: while current OLLMs handle basic textual formats well, they struggle significantly with rigid structural formatting and fine-grained audio-visual constraints.
Regarding format control, OLLMs face challenges with complex structures like JSON and strict patterns such as Timestamps, reflecting limitations in token-level output regulation. In content constraints, models show difficulties with directives related to Editing Transition, Temporal Grounding, and Anchor. The lower performance on Editing Transition suggests limited internalization of professional cinematic techniques, while the gaps in Temporal Grounding and Anchor indicate that visual and auditory streams are often processed as isolated channels. Notably, specialized video captioning models do not outperform general-purpose models (e.g., ASID-Captioner versus Qwen2.5-Omni-7B) under our evaluation, because our benchmark emphasizes precise adherence to instruction-specified attributes, actions, or events rather than unconstrained, detailed video descriptions.\

The OmniCaptioner-IF series addresses these limitations with comprehensive improvements, outperforming baselines across the entire constraint spectrum. Notably, the models transform previously weak adherence to rigid formats like JSON and Timestamp into robust performance, while also showing stronger handling of fine-grained audio-visual constraints. This demonstrates that OmniCaptioner-IF excels both in strict output regulation and in deep cross-modal understanding. More details can be found in Appendix~\ref{sec:error}.



\begin{figure}[htbp] 
\centering
\includegraphics[width=0.9\textwidth]{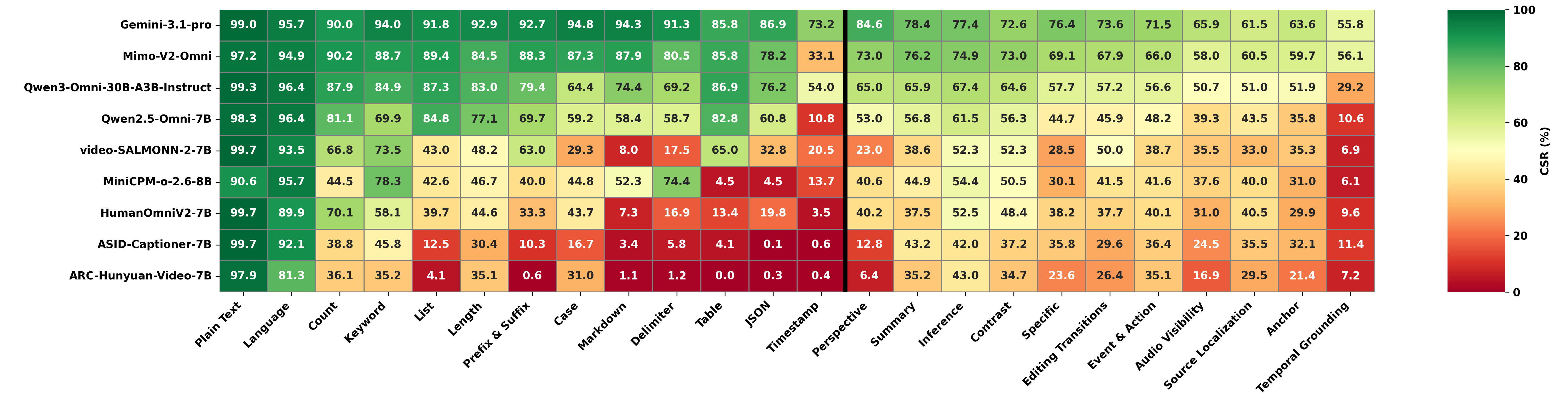} 
\caption{CSR performance of different models on different formats and audio-visual constraint types.}
\label{fig:constraint_type_csr}
\end{figure}

\noindent \textbf{Error Analysis.}
Our analysis of model responses reveals key error categories. For format constraints, common violations are (1) malformed JSON (e.g., missing keys or bracket mismatches) and (2) incorrect timestamp formatting (e.g., not following the ``MM:SS'' template). For content constraints, frequent issues include (1) misidentifying or omitting Editing Transitions, (2) inaccurate Temporal Grounding of events, and (3) failing to establish cross-modal Anchors. We also find that when only audio is provided, the models’ audio temporal grounding capability is significantly weaker than when both audio and visual modalities are available. More examples are provided in Appendix~\ref{sec:error}.

\section{Conclusion}
In this work, we introduce OmniCap-IF, a pioneering benchmark explicitly designed to evaluate instruction-following capabilities in omni-modal video captioning. By systematically defining 50 distinct constraints across format, visual, audio, and cross-modal dimensions, and deploying a rigorous dual evaluation protocol, OmniCap-IF provides a comprehensive diagnostic testbed. Our extensive evaluations yield profound insights into the limitations of current OLLMs,
and observe a distinct lack of deep cross-modal synergy in open-source models compared to their proprietary counterparts. 
Moreover, we also curate OmniCap-IF-54K, a 54K instruction-tuning dataset, and develop OmniCaptioner-IF. Our model not only masters complex structural constraints but also demonstrates remarkable ability in omni-modal captioning.

\section*{Limitations}
\label{sec:limit}
While OmniCap-IF and OmniCaptioner-IF significantly advance the evaluation and generation of instruction-following omni-modal captions, our work has certain limitations. 

First, our evaluation relies partially on LLM-as-a-judge for content constraints. Although we mitigate potential biases by prioritizing factual QA over fluency and utilizing rigorous rule-based programmatic tools for format/temporal verification, the inherent hallucinations of judge models cannot be entirely eliminated. 
Second, as revealed by our "format-content tradeoff" analysis, current OLLMs still struggle to maintain deep cross-modal reasoning when burdened with overly strict syntactic constraints (e.g., deeply nested JSONs). While instruction tuning alleviates this issue, bridging the gap between rigid output formatting and complex multi-step reasoning remains a formidable challenge for future research.
Finally, the current benchmark primarily focuses on videos ranging from 30 to 90 seconds. Evaluating models on ultra-long videos (e.g., hour-long movies or podcasts) with dense, multi-constraint instructions represents a crucial next step for the community.

\bibliographystyle{unsrtnat}
\bibliography{references} 

\appendix
\newtcolorbox{promptbox}[1]{
    enhanced,
    breakable,    
    arc=5pt,                         
    boxrule=0.6pt,                   
    colframe=gray!60!black,          
    colbacktitle=gray!60!black,      
    colback=gray!12!white,           
    coltitle=white,                  
    fonttitle=\bfseries\small,       
    before skip=12pt,                
    after skip=6pt,                  
    toptitle=2pt,                    
    bottomtitle=2pt,                 
    left=12pt,                       
    right=12pt,                      
    top=6pt,                        
    bottom=6pt,                     
    title=#1
}

\lstset{
    basicstyle=\small\ttfamily,
    breaklines=true,
    columns=fullflexible
}

\definecolor{mainheader}{RGB}{160, 160, 160}     
\definecolor{border}{RGB}{125, 180, 125}    
\definecolor{header}{RGB}{198, 230, 198}    
\definecolor{body}{RGB}{232, 245, 232}      
\definecolor{categorybg}{RGB}{240, 240, 240}    
\newtcolorbox{categorybox}[1]{
    colback=gray!10,
    colframe=white,
    sharp corners,
    boxrule=0pt,
    fontupper=\bfseries\small,
    center upper,
    height=1.5em,
    valign=center,
    upperbox=visible,
    top=0pt, bottom=0pt,
    title=#1,
    colbacktitle=gray!10,
    coltitle=black
}
\newtcolorbox{rulebox}[1]{
    enhanced,
    arc=2pt,                         
    boxrule=0.6pt,                   
    colframe=border,             
    colbacktitle=header,         
    colback=body,                
    coltitle=black,                  
    fonttitle=\bfseries\small,       
    left=8pt, right=8pt,             
    top=4pt, bottom=4pt,
    toptitle=3pt, bottomtitle=3pt,   
    titlerule=0.4pt,                 
    titlerule style=border,      
    title={#1}                       
}

\newtcolorbox{openendedcontainer}[1]{
    enhanced,
    colback=header,
    colframe=border,
    arc=5pt,
    boxrule=1pt,
    title={#1},
    coltitle=white,
    left=1pt, right=1pt, top=2pt, bottom=2pt,
    middle=5pt
}
\newtcolorbox{innerquestion}{
    colback=body,
    colframe=border!60,
    arc=3pt,
    boxrule=0.8pt,
    left=2pt, right=2pt, top=2pt, bottom=2pt,
    before skip=2pt, after skip=2pt, 
    fontupper=\small
}

\definecolor{errorred}{RGB}{120, 30, 30}       
\definecolor{captionbg}{RGB}{248, 249, 250}   
\definecolor{captionborder}{RGB}{220, 225, 230}
\definecolor{analysisbg}{RGB}{255, 252, 240}  
\definecolor{analysisborder}{RGB}{230, 150, 50} 
\definecolor{trainedbg}{RGB}{246, 252, 246}     
\definecolor{trainedborder}{RGB}{85, 150, 105}  

\newtcolorbox{erroranalysiscontainer}[1]{
    enhanced,
    breakable,
    colback=white,
    colframe=errorred,
    arc=5pt,
    boxrule=1.2pt,
    title={\large\bfseries #1}, 
    colbacktitle=errorred,
    coltitle=white
}

\newtcolorbox{error_analysis_captionbox}[1]{
    enhanced,
    colback=captionbg,
    colframe=captionborder,
    arc=3pt,
    boxrule=0.5pt,
    title={\small\bfseries #1},
    colbacktitle=captionborder!80,
    coltitle=black,
    left=2pt, right=2pt, top=2pt, bottom=2pt,
    before skip=4pt, after skip=4pt
}

\newtcolorbox{error_analysis_analysisbox}[2]{
    enhanced,
    colback=#2,
    colframe=analysisborder,
    arc=3pt,
    boxrule=0.5pt,
    title={\small\bfseries #1},
    colbacktitle=analysisborder!60,
    coltitle=black,
    left=2pt, right=2pt, top=2pt, bottom=2pt,
    before skip=4pt, after skip=4pt
}

\newtcolorbox{error_analysis_trainedbox}[1]{
    enhanced,
    colback=trainedbg,                 
    colframe=trainedborder,     
    arc=3pt,
    boxrule=0.5pt,
    title={\small\bfseries #1}, 
    colbacktitle=trainedborder, 
    coltitle=white,             
    left=2pt, right=2pt, top=2pt, bottom=2pt,
    before skip=4pt, after skip=4pt
}

\section{Real-World Applications of Omni-Modal Instruction-Following}
\label{sec:app}
The rapid progress of omni-modal video understanding models has driven the growing adoption of instruction-following video captioning, a task that requires generating textual descriptions aligned with specific, predefined constraints. In contrast to holistic video summarization, this more targeted paradigm plays a crucial role in a wide range of downstream applications. We further outline six representative real-world use cases:
\begin{itemize}[leftmargin=*, topsep=0pt, itemsep=0pt, parsep=0pt]
    \item\textbf{Text-to-Audio-Video (T2AV) Generation:} Generative models (e.g., Sora, Veo) require dual-track scripts that provide highly detailed, imaginative, and sensory-rich descriptions of both visual scenes and synchronized audio tracks. Generic event-level captions cannot provide sufficient granularity. In this case, captions must explicitly maintain a dual-narrative structure (visual track and audio track) with precise cinematic and acoustic attributes.

    \item\textbf{Embodied Task Planning:} Autonomous agents (e.g., home robots, autonomous vehicles) must simultaneously process visual environments and off-screen auditory alerts (e.g., a crying baby, an approaching siren) to make rapid situational decisions. In this case, the caption must act as a first-person, action-oriented summary that explicitly identifies anomalies across both modalities.

    \item\textbf{Cross-Modal Video Retrieval:} Multimodal search engines require unique semantic fingerprints to resolve the ambiguity inherent in single-modality queries. By pinpointing moments where specific visual actions intersect with distinct audio events (e.g., "crying while chopping onions"), models can filter out irrelevant noise. In this case, the caption must extract highly exclusive cross-modal features and strictly exclude negative keywords.
    
    \item\textbf{Automated Understanding and Surveillance:} Security and automated meeting analysis systems rely on extracting structured information, emphasizing the spatiotemporal alignment and causal reasoning between audio and visual streams (e.g., matching a speaker's face with their voice). In this case, the caption must strictly adhere to data schemas (e.g., JSON) to ensure interoperability in industrial pipelines.
    
    \item\textbf{Accessibility and Modality Compensation:} To assist visually or hearing-impaired users, models must provide fluent modality translation, such as Audio Descriptions (translating visual actions into TTS-friendly narratives) or Closed Captions (transcribing speech and environmental sounds). In this case, the caption must selectively filter and prioritize information from one modality to compensate for the absence of another.
    
    \item\textbf{Video Editing and Script Reverse-Engineering:} Editors require chronological, scene-by-scene structural breakdowns with precise timestamp alignments. Advanced editing techniques like J-cuts or L-cuts demand strict attention to audio-visual desynchronization. In this case, both structural formatting (e.g., Markdown tables) and fine-grained temporal comprehension are necessary for the caption.
\end{itemize}

\section{Constraint System}
\label{sec:system}
As mentioned in the main paper, our taxonomy encompasses 50 constraint types divided into Format Constraints (Structural, Stylistic) and Content Constraints (Visual, Audio, Audio-Visual). Tables \ref{tab:format_constraints}, ~\ref{tab:vis_aud_constraints} and \ref{tab:omni_constraints} provide the precise definitions and corresponding examples for each constraint category.

\begin{table}[h]
  \caption{Detailed Definitions of Format Constraints (Structural and Stylistic).}
  \label{tab:format_constraints}
  \footnotesize
  \renewcommand{\arraystretch}{1.3}
  \begin{tabularx}{\textwidth}{l >{\raggedright\arraybackslash}p{3cm} X X}
    \toprule
    \textbf{Category} & \textbf{Constraint Name} & \textbf{Definition} & \textbf{Example Prompt} \\
    \midrule
    \multirow{18}{*}{\textbf{Structural}} 
    & Plain Text & Natural language text without any special structures or markers. & ``Please describe this video in a paragraph.'' \\ \cmidrule{2-4}
    & JSON Object & A collection of key-value pairs complying with JSON specifications. & ``Output the core entities and their attributes in a JSON object format.'' \\ \cmidrule{2-4}
    & JSON Array & A list that complies with JSON specifications. & ``List all the actions performed by characters in the form of a JSON array.'' \\ \cmidrule{2-4}
    & Unordered List & Use symbols such as -, * to organize information into a list. & ``List all transportation vehicles using an unordered list starting with '-'.'' \\ \cmidrule{2-4}
    & Ordered List & Use ordered symbols (1., A., etc.) to organize information. & ``Describe the three key behaviors using an ordered list starting with '1.'.'' \\ \cmidrule{2-4}
    & Table & Use a table in Markdown syntax to organize information. & ``Use a Markdown table to record the items, setting name, color, and size columns.'' \\ \cmidrule{2-4}
    & Keyword & Precisely include or absolutely exclude designated literal strings. & ``Your answer must precisely include the keyword 'bicycle'.'' \\ \cmidrule{2-4}
    & Timestamp Format & Include timestamp strings conforming to specifications (e.g., [MM:SS]). & ``Mark the time period before each action in the format of [Min:Sec - Min:Sec].'' \\ \midrule
    
    \multirow{14}{*}{\textbf{Stylistic}} 
    & Markdown Syntax & Use designated Markdown syntax (headings, bold, highlight, italics). & ``Summarize the content, bold the names, and set scenes as level-two headings.'' \\ \cmidrule{2-4}
    & Prefix Suffix & Add specified strings to the beginning and end of the output text. & ``The beginning must be 'Video Summary:', and the ending must be '--End--'.'' \\ \cmidrule{2-4}
    & Delimiter & Use specific symbols (e.g., , | ; ---) to separate information fragments. & ``List the characters and actions, using '|' to separate each group.'' \\ \cmidrule{2-4}
    & Length & Limit the length of the output in units of words, sentences, or paragraphs. & ``Please summarize the video content in 50 to 60 words.'' \\ \cmidrule{2-4}
    & Count & Place quantity limits on the description of elements (e.g., objects). & ``Please describe three character features in the video.'' \\ \cmidrule{2-4}
    & Case & Specify the uppercase or lowercase format for English output. & ``Please describe the video in uppercase.'' \\ \cmidrule{2-4}
    & Language & Specify the language of the output (entirely or partially). & ``Describe the text in English and translate them into Chinese.'' \\ \bottomrule
  \end{tabularx}
\end{table}

\clearpage

\renewcommand{\arraystretch}{1.3} 

{\footnotesize
\begin{longtable}{l p{3cm} p{6cm} p{6cm}}
    
    \caption{Detailed Definitions of Visual and Audio Content Constraints.} \label{tab:vis_aud_constraints} \\
    
    \toprule
    \textbf{Modality} & \textbf{Constraint Name} & \textbf{Definition} & \textbf{Example Prompt} \\
    \midrule

    \multirow{26}{*}{\shortstack{\textbf{Visual}}} 
    & \multicolumn{3}{l}{\cellcolor{gray!15}\textit{Core Elements}} \\
    & Visual Entities Attributes & Identify key entities (persons, objects, scenes) and their static/dynamic attributes. & ``Describe the appearance of that red car based solely on the visuals.'' \\ \cmidrule{2-4}
    & Visual Events Actions & Describe key events, single/interactive actions, and state changes occurring in the video. & ``Describe in detail the complete physical action process of the boy feeding the puppy.'' \\ \cmidrule{2-4}
    
    & \multicolumn{3}{l}{\cellcolor{gray!15}\textit{Cinematic Elements}} \\
    & Visual Cinematic Elements & Describe camera movements, shot sizes, and editing skills (e.g., panning, close-up). & ``Describe the shot language of this clip, including the main camera movements.'' \\ \cmidrule{2-4}
    
    & \multicolumn{3}{l}{\cellcolor{gray!15}\textit{Perspective \& Focus}} \\
    & Visual Perspective & Specify the narrative perspective for generating the description. & ``As the cat in the video, describe your day in the first person.'' \\ \cmidrule{2-4}
    & Visual Focus & Focus only on particular aspects of the video, entities, or regions. & ``Only describe all the activities of the girl wearing the yellow dress.'' \\ \cmidrule{2-4}
    & Visual Include & Constrain the model to necessarily mention specific facts or entities. & ``Describe the video, and you must mention the transportation used by the protagonist.'' \\ \cmidrule{2-4}
    & Visual Exclude & Constrain the model to deliberately ignore specific facts or entities. & ``Describe the visuals, but do not mention any conditions regarding the weather.'' \\ \cmidrule{2-4}
    & Visual Comparative & Compare the similarities and differences between entities or time points. & ``Compare the changes of the items on the table at the beginning and the end.'' \\ \cmidrule{2-4}
    
    & \multicolumn{3}{l}{\cellcolor{gray!15}\textit{Abstraction}} \\
    & Visual Specific & Describe the visual frame content in detail and objectively. & ``Provide a detailed description of the appearance of all the characters in the video...'' \\ \cmidrule{2-4}
    & Visual Summary & Perform a high-level generalization and summary of the video content. & ``Summarize the main events of this video in one sentence.'' \\ \cmidrule{2-4}
    & Visual Inference & Infer intentions, emotions, or causal relationships based strictly on visual cues. & ``Based on the expression of the character, infer his current mood.'' \\ \cmidrule{2-4}
    
    & \multicolumn{3}{l}{\cellcolor{gray!15}\textit{Temporal Grounding}} \\
    & Visual Temporal Grounding & Accurately point out the precise time periods when specific visual events occur. & ``Write down the time points when the girl in the red clothes appears and disappears.'' \\ 
    \midrule
    
    \multirow{24}{*}{\shortstack{\textbf{Audio}}} 
    & \multicolumn{3}{l}{\cellcolor{gray!15}\textit{Core Elements}} \\
    & Audio Entities Attributes & Identify sound entities and attributes (timbre, pitch, volume, musical style). & ``Describe the timbre and pitch characteristics of that crisp bird song in the audio.'' \\ \cmidrule{2-4}
    & Audio Events Actions & Describe key sound events and specific sounding actions/processes. & ``Describe in detail the whole process of the wind sound changing from gentle to rapid.'' \\ \cmidrule{2-4}
    
    & \multicolumn{3}{l}{\cellcolor{gray!15}\textit{Production \& Structure}} \\
    & Audio Production Structure & Describe sound processing, transitions, and composition layers. & ``Describe the audio design, including sound transition methods and layers.'' \\ \cmidrule{2-4}
    
    & \multicolumn{3}{l}{\cellcolor{gray!15}\textit{Attention \& Selection}} \\
    & Audio Perspective & Specify the narrative perspective for generating the audio description. & ``As the singer in the audio, describe your vocal feelings in the first person.'' \\ \cmidrule{2-4}
    & Audio Focus & Focus on specific sound entities/layers or audio details. & ``Only describe the changes in the man's tone in the audio.'' \\ \cmidrule{2-4}
    & Audio Include & Explicitly require the inclusion of specific audio content. & ``Describe the audio, and include the timbre changes of the background music.'' \\ \cmidrule{2-4}
    & Audio Exclude & Constrain the model to deliberately ignore specific acoustic facts or entities. & ``Describe the sound clip, but do not mention any sounds made by humans.'' \\ \cmidrule{2-4}
    & Audio Comparative & Compare the similarities and differences of sounds at different time points. & ``Compare the changes in volume of the background music at the beginning and end.'' \\ \cmidrule{2-4}
    
    & \multicolumn{3}{l}{\cellcolor{gray!15}\textit{Interpretation}} \\
    & Audio Specific & Objectively and elaborately describe the sound content and change process. & ``Accurately transcribe the dialogue content...'' \\ \cmidrule{2-4}
    & Audio Summary & Perform a high-level generalization of the pure audio content. & ``Summarize the core auditory events in one sentence.'' \\ \cmidrule{2-4}
    & Audio Inference & Infer the speaker's emotion or off-screen state based on intonation. & ``Infer the speaker's hidden emotion based on their voice.'' \\ \cmidrule{2-4}
    
    & \multicolumn{3}{l}{\cellcolor{gray!15}\textit{Temporal Grounding}} \\
    & Audio Temporal Grounding & Accurately point out the precise time periods of sound events. & ``Write down the times when the siren starts ringing and completely stops.'' \\ 
    \bottomrule
\end{longtable}
} 

\begin{table}[H]
  \caption{Detailed Definitions of Audio-Visual Content Constraints.}
  \label{tab:omni_constraints}
  \footnotesize
  \renewcommand{\arraystretch}{1.3}
  \begin{tabularx}{\textwidth}{l >{\raggedright\arraybackslash}p{3.8cm} X X}
    \toprule
    \textbf{Modality} & \textbf{Constraint Name} & \textbf{Definition} & \textbf{Example Prompt} \\
    \midrule
    \multirow{22}{*}{\shortstack{\textbf{Audio-Visual}}} 
    & \multicolumn{3}{l}{\cellcolor{gray!15}\textit{Core Elements}} \\
    & Omni Events Actions & Describe the causal relationships and cross-modal interactions between visuals and sounds. & ``Describe the process of the arguing, including physical actions and tone changes.'' \\ \cmidrule{2-4}
    & Omni Audio Visibility & Judge whether the heard sound entity exists in the current visual frame (On/Off-screen). & ``List original dialogues from inside the frame and voiceovers from outside separately.'' \\ \cmidrule{2-4}
    & Omni Source Localization & Locate the entity emitting the sound and describe its visual attributes or states. & ``Point out what object emits the 'beep' sound and describe its color and location.'' \\ \cmidrule{2-4}
    & \multicolumn{3}{l}{\cellcolor{gray!15}\textit{Editing \& Transitions}} \\
    & Omni Editing Transitions & Describe the temporal correlation between visual cuts and sound cuts (J-cut, L-cut). & ``Describe how the background music rhythm matches the beat of the fast editing.'' \\ \cmidrule{2-4}
    & \multicolumn{3}{l}{\cellcolor{gray!15}\textit{Coordination \& Attention}} \\
    & Omni Perspective & Specify an immersive narrative perspective combining what is seen and heard. & ``As a skier with a GoPro, describe the snowscape (visual) and howling wind (audio).'' \\ \cmidrule{2-4}
    & Omni Anchor & Use one modality as an anchor to extract relevant information from the other. & ``When hearing the explosion, focus on describing the expressions of all characters.'' \\ \cmidrule{2-4}
    & Omni Contrast & Compare contradictions between the semantics of the visual frame and audio stream. & ``Compare the funeral scene in the visuals with the upbeat music style playing.'' \\ \cmidrule{2-4}
    & \multicolumn{3}{l}{\cellcolor{gray!15}\textit{Reasoning}} \\
    & Omni Specific & Objectively retelling the seen and heard content intertwined along the timeline. & ``Record every lightning flash and the volume changes of the thunder simultaneously.'' \\ \cmidrule{2-4}
    & Omni Summary & Comprehensively extract audio-visual events and summarize the overall core narrative. & ``Combining the chasing behaviors and shouting, summarize the core conflict.'' \\ \cmidrule{2-4}
    & Omni Inference & Infer deep intentions or materials by combining visual and auditory cues simultaneously. & ``Based on his micro-expressions and trembling voice, infer his true psychological state.'' \\ \cmidrule{2-4}
    & \multicolumn{3}{l}{\cellcolor{gray!15}\textit{Temporal Grounding}} \\
    & Omni Temporal Grounding & Locate key time points where synchronization or misalignment occurs between modalities. & ``Find the specific period where there is a desynchronization between lips and voice.'' \\ \bottomrule
  \end{tabularx}
\end{table}

\section{Temporal Grounding Evaluation Scheme}
\label{sec:timestamp_eval}

For temporal grounding constraints (including Visual, Audio, and Audio-Visual modalities), our programmatic evaluation engine employs two distinct verification schemes based on the nature of the instruction: Time Intervals and Precise Time Points.

\subsection{Time Intervals: Temporal Intersection over Union (t-IoU)}
When a prompt requires identifying the duration of an event (e.g., ``the duration of the siren [00:10 - 00:18]''), we utilize the Temporal Intersection over Union (t-IoU) metric to measure the overlap between the predicted interval and the ground truth. The calculation is defined as:
\begin{equation}
    t\text{-IoU} = \frac{\text{Intersection}(\text{Predicted}, \text{Ground Truth})}{\text{Union}(\text{Predicted}, \text{Ground Truth})} = \frac{|\mathcal{I}_{pred} \cap \mathcal{I}_{gt}|}{|\mathcal{I}_{pred} \cup \mathcal{I}_{gt}|}
\end{equation}
where $\mathcal{I}_{pred}$ and $\mathcal{I}_{gt}$ represent the predicted and ground truth intervals, respectively. We adopt a threshold of \textbf{$t\text{-IoU} \ge 0.5$} as the success criterion for fulfilling the temporal grounding constraint.

\subsection{Precise Time Points: Dynamic Tolerance Margin}
For instructions requesting the identification of a specific trigger point (e.g., ``the exact moment the glass breaks [00:15]''), we apply a dynamic tolerance margin ($\Delta t$) to account for the inherent characteristics of video sampling and human annotation. The tolerance is calculated based on the total video duration:
\begin{equation}
    \Delta t = \max(1.0\,\text{s}, \text{Total Video Length} \times 5\%)
\end{equation}
A prediction is considered successful if the absolute error between the predicted time ($T_{pred}$) and the ground truth time ($T_{gt}$) satisfies:
\begin{equation}
    |T_{pred} - T_{gt}| \le \Delta t
\end{equation}

The rationale for this dual-adaptive design includes:
\begin{itemize}[leftmargin=*, topsep=2pt]
    \item \textbf{Floor Mechanism:} The 1.0s minimum tolerance provides a safety net that accounts for human reaction time and the subjective lag in manual annotation. It ensures models are not penalized for sub-second offsets that are perceptually negligible.
    \item \textbf{Dynamic Scaling:} By scaling the tolerance to 5\% of the total video length (e.g., 3.0s tolerance for a 60s video), we acknowledge the practical limitations of Large Multimodal Models (LMMs), which typically operate at a sampling rate of 1 FPS. This dynamic expansion allows the model to be rewarded for successful semantic localization without being unfairly measured against unrealistic millisecond-level precision in long-form content.
\end{itemize}

\section{Dataset Samples}
\label{sec:data_sample}
\begin{tcolorbox}[
    width=\linewidth, 
    enhanced,
    breakable,
    colback=white, 
    colframe=gray!80, 
    title={Data Sample - 1}, 
    colbacktitle=mainheader,
    fonttitle=\bfseries,
    arc=5pt
]
    \setlength{\tabcolsep}{0pt} 
    \renewcommand{\arraystretch}{0}
    \begin{tabularx}{\textwidth}{XXXX}
        \includegraphics[width=\linewidth]{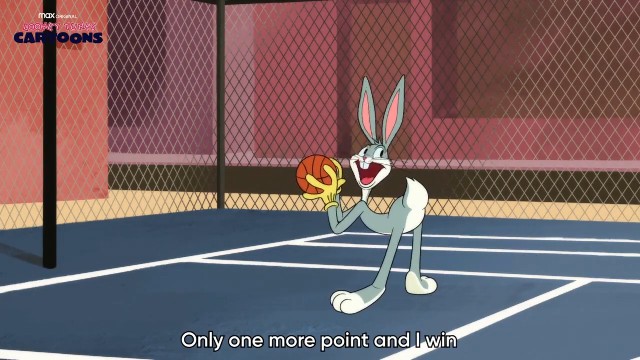} & 
        \includegraphics[width=\linewidth]{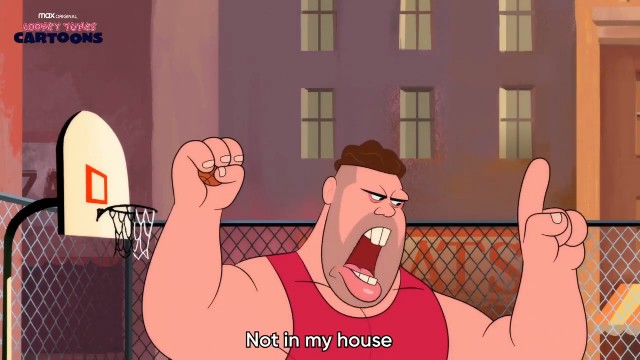} & 
        \includegraphics[width=\linewidth]{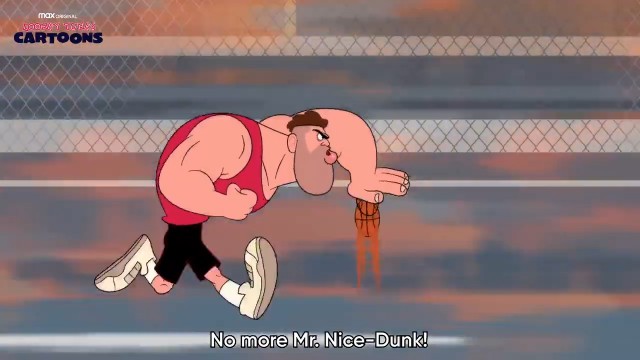} & 
        \includegraphics[width=\linewidth]{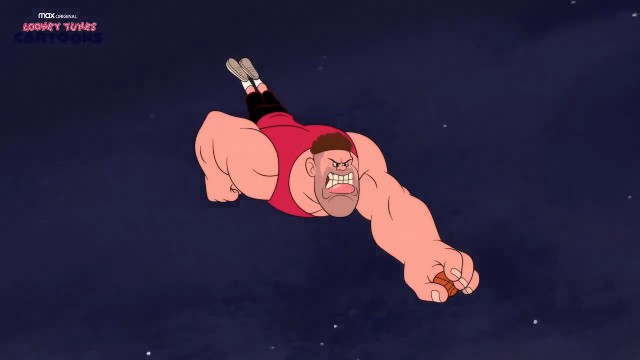} \\
        \includegraphics[width=\linewidth]{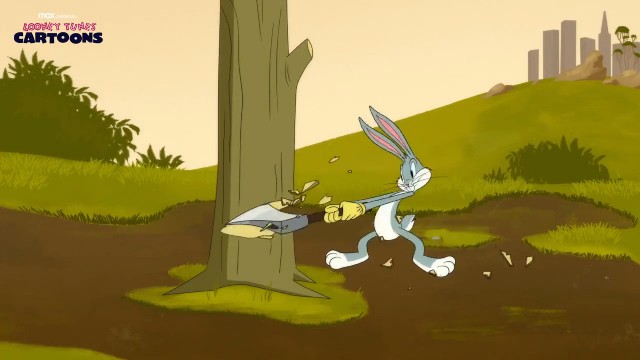} & 
        \includegraphics[width=\linewidth]{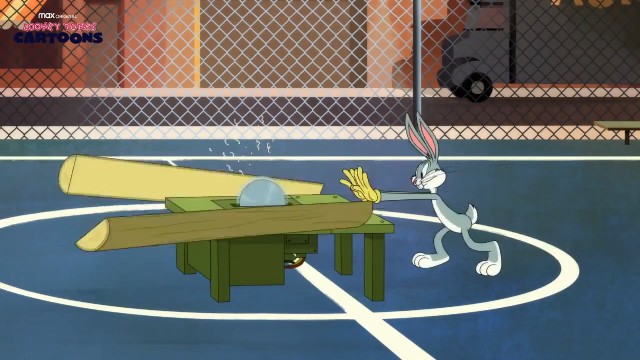} & 
        \includegraphics[width=\linewidth]{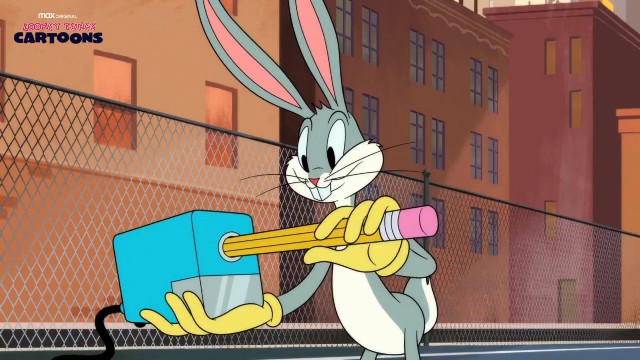} & 
        \includegraphics[width=\linewidth]{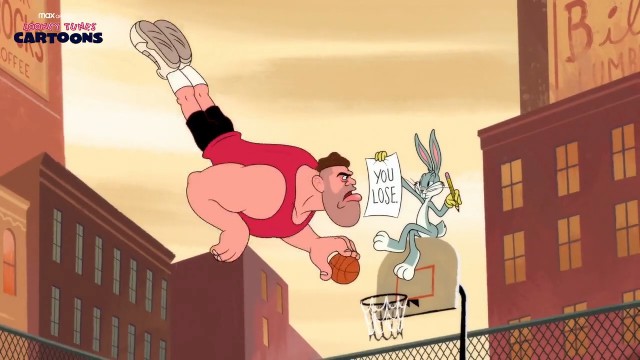}
    \end{tabularx}

    \maybeincludegraphics[width=\linewidth]{yingui/001.svg}
    
    \vspace{2pt}
    
    \textbf{Prompt:}\\
    \textit{Analyze the defensive play scene around 00:08. Use a Markdown table with columns `Timestamp', `Visual Action', `Audio Content', and `Character State'. Record the exact moment the character blocks the ball, his spoken line "Not in my house", and his subsequent landing. If the landing produces a squishing sound, you must describe the process of his landing.}
    
    \begin{categorybox}{}
        FORMAT CHECKS
    \end{categorybox}

    \begin{rulebox}{Format-001: Use a Markdown table with columns `Timestamp', `Visual Action', `Audio Content', and `Character State'.}
        \small \textbf{Constraint:} table \qquad 
        \textbf{Parameters:} "col\_name": ["Timestamp", "Visual Action", "Audio Content","Character State"]
    \end{rulebox}

    \begin{categorybox}{}
        CONTENT CHECKS
    \end{categorybox}

    \begin{openendedcontainer}{Content-001: Use a Markdown table with columns `Visual Action', `Audio Content', and `Character State'.}
        
        \textbf{Constraint:} \texttt{omni\_events\_actions}
        \vspace{2pt}
        \begin{innerquestion}
            \small \textbf{Does the response try to analyze the defensive play scene as required?} \\[6pt]
            \begin{tabularx}{\linewidth}{@{}p{0.47\linewidth} @{\hspace{0.05\linewidth}} X @{}}
                A. Yes & B. No
            \end{tabularx} \\[6pt]
            \textit{Correct Answer: \textbf{Yes}}
        \end{innerquestion}

        \begin{innerquestion}
            \small \textbf{How does the character block the ball in the scene around 00:08?} \\[6pt] 
            \begin{tabularx}{\linewidth}{@{}p{0.47\linewidth} @{\hspace{0.05\linewidth}} X @{}}
                A. He jumps high and swats the ball away & B. He stands on the ground and grabs the ball
            \end{tabularx} \\[4pt] 
            \begin{tabularx}{\linewidth}{@{}p{0.47\linewidth} @{\hspace{0.05\linewidth}} X @{}}
                C. He kicks the ball out of the air & D. None of the above
            \end{tabularx} \\[6pt] 
            \textit{Correct Answer: \textbf{A}}
        \end{innerquestion}

        \begin{innerquestion}
            \small \textbf{What is the character's state immediately after blocking the ball?} \\[6pt] 
            \begin{tabularx}{\linewidth}{@{}p{0.47\linewidth} @{\hspace{0.05\linewidth}} X @{}}
                A. He falls to the ground & B. He runs to the other side of the court
            \end{tabularx} \\[4pt] 
            \begin{tabularx}{\linewidth}{@{}p{0.47\linewidth} @{\hspace{0.05\linewidth}} X @{}}
                C. He holds the ball while standing still & D. None of the above
            \end{tabularx} \\[6pt] 
            \textit{Correct Answer: \textbf{C}}
        \end{innerquestion}

        \begin{innerquestion}
            \small \textbf{What sound does the muscular man make during the landing?} \\[6pt] 
            \begin{tabularx}{\linewidth}{@{}p{0.47\linewidth} @{\hspace{0.05\linewidth}} X @{}}
                A. Laughter & B. Grunt
            \end{tabularx} \\[4pt] 
            \begin{tabularx}{\linewidth}{@{}p{0.47\linewidth} @{\hspace{0.05\linewidth}} X @{}}
                C. Shout & D. None of the above
            \end{tabularx} \\[6pt] 
            \textit{Correct Answer: \textbf{D}}
        \end{innerquestion}

    \end{openendedcontainer}
    
    \begin{openendedcontainer}{Content-002: Record the exact moment the character blocks the ball}
        
        \textbf{Constraint:} \texttt{visual\_temporal\_grounding}
        \vspace{2pt}
        \begin{innerquestion}
            \small \textbf{According to the video, at what exact moment does the character block (grab) the ball?} \\[6pt]
            \textit{Correct Answer: \textbf{00:11}}
        \end{innerquestion}

    \end{openendedcontainer}

    \begin{openendedcontainer}{Content-003: Record the exact moment of his spoken line `Not in my house'}
        
        \textbf{Constraint:} \texttt{audio\_temporal\_grounding}
        \vspace{2pt}
        \begin{innerquestion}
            \small \textbf{According to the video, when does the character start saying the line `Not in my house'?} \\[6pt]
            \textit{Correct Answer: \textbf{00:13}}
        \end{innerquestion}

    \end{openendedcontainer}

    \begin{openendedcontainer}{Content-004: If the landing produces a squishing sound, you must describe the process of his landing.}
        
        \textbf{Constraint:} \texttt{omni\_specific}
        \vspace{2pt}
        \begin{innerquestion}
            \small \textbf{Does the response describe a landing process and explain a squishing sound?} \\[6pt]
            \begin{tabularx}{\linewidth}{@{}p{0.47\linewidth} @{\hspace{0.05\linewidth}} X @{}}
                A. Yes & B. No
            \end{tabularx} \\[6pt]
            \textit{Correct Answer: \textbf{Yes}}
        \end{innerquestion}

        \begin{innerquestion}
            \small \textbf{Regarding the 'subsequent landing' of the character who blocked the ball (the Crusher), which statement is factually correct for this specific scene?} \\[6pt] 
            \begin{tabularx}{\linewidth}{@{}p{0.47\linewidth} @{\hspace{0.05\linewidth}} X @{}}
                A. He lands heavily, producing a loud squishing sound & B. He lands softly on his toes
            \end{tabularx} \\[4pt] 
            \begin{tabularx}{\linewidth}{@{}p{0.47\linewidth} @{\hspace{0.05\linewidth}} X @{}}
                C. He does not land because he did not jump to block the ball & D. He lands simultaneously with Bugs Bunny
            \end{tabularx} \\[6pt] 
            \textit{Correct Answer: \textbf{C}}
        \end{innerquestion}

        \begin{innerquestion}
            \small \textbf{Who actually hits the ground (lands) in the sequence?} \\[6pt] 
            \begin{tabularx}{\linewidth}{@{}p{0.47\linewidth} @{\hspace{0.05\linewidth}} X @{}}
                A. The Crusher (the blocker) & B. The referee
            \end{tabularx} \\[4pt] 
            \begin{tabularx}{\linewidth}{@{}p{0.47\linewidth} @{\hspace{0.05\linewidth}} X @{}}
                C. Bugs Bunny & D. None of the above
            \end{tabularx} \\[6pt] 
            \textit{Correct Answer: \textbf{D}}
        \end{innerquestion}

    \end{openendedcontainer}

\end{tcolorbox}

\begin{tcolorbox}[
    width=\linewidth, 
    enhanced,
    breakable,
    colback=white, 
    colframe=gray!80, 
    title={Data Sample - 2}, 
    colbacktitle=mainheader,
    fonttitle=\bfseries,
    arc=5pt
]
    \setlength{\tabcolsep}{0pt} 
    \renewcommand{\arraystretch}{0}
    \begin{tabularx}{\textwidth}{XXXX}
        \includegraphics[width=\linewidth]{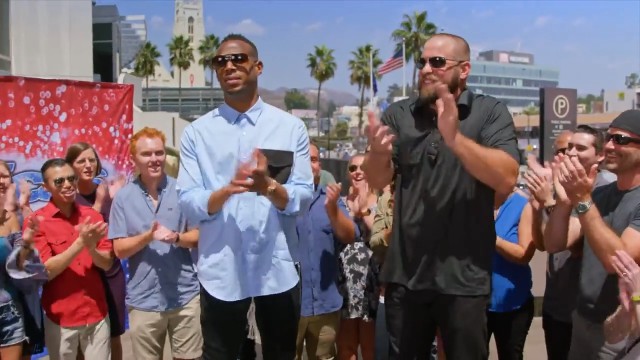} & 
        \includegraphics[width=\linewidth]{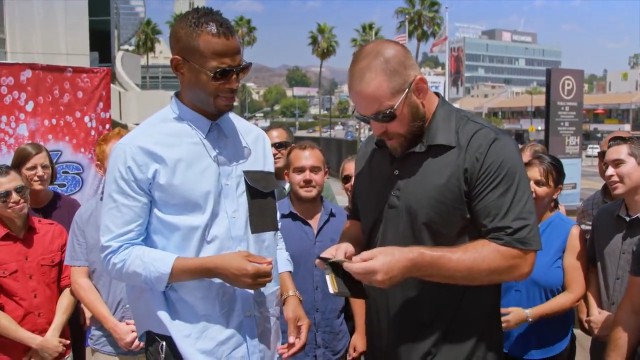} & 
        \includegraphics[width=\linewidth]{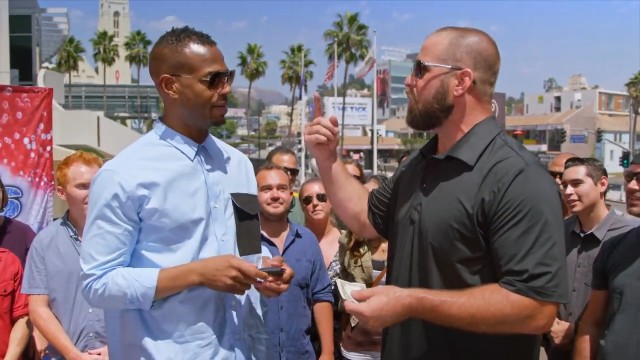} & 
        \includegraphics[width=\linewidth]{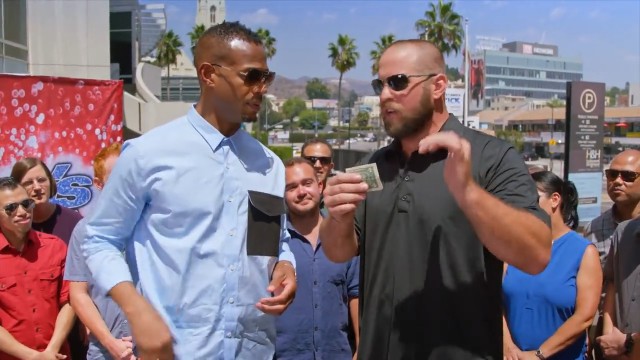} \\
        \includegraphics[width=\linewidth]{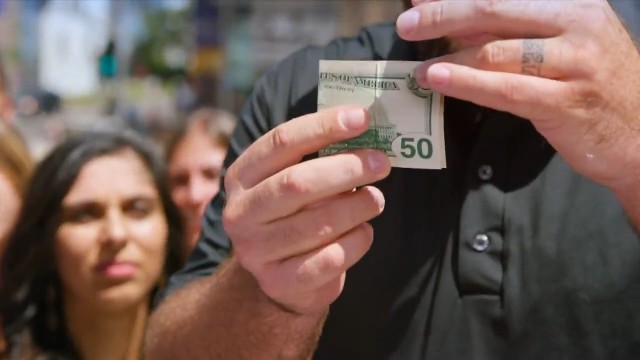} & 
        \includegraphics[width=\linewidth]{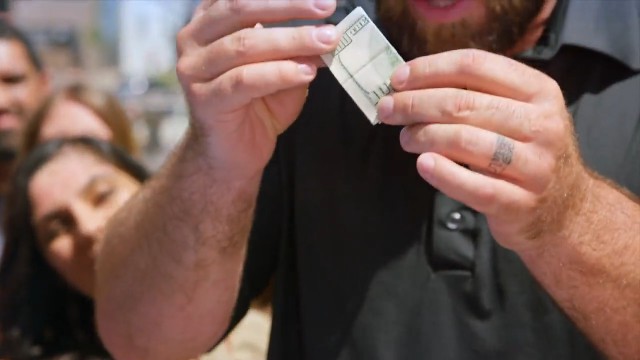} & 
        \includegraphics[width=\linewidth]{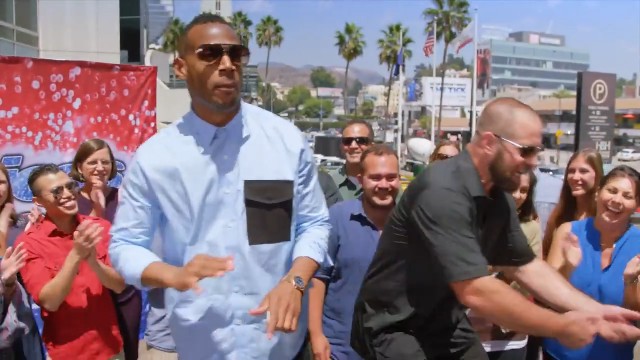} & 
        \includegraphics[width=\linewidth]{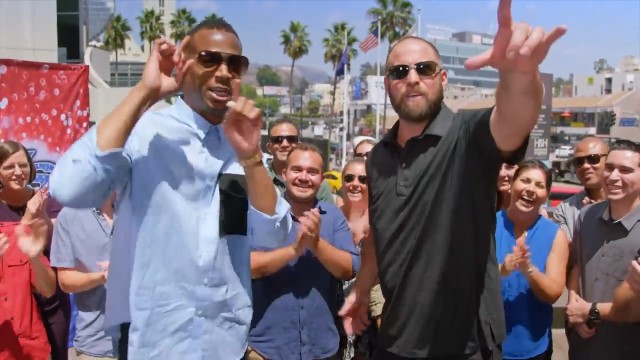}
    \end{tabularx}

    \maybeincludegraphics[width=\linewidth]{yingui/340.svg}
    
    \vspace{2pt}
    
    \textbf{Prompt:}\\
    \textit{Please analyze the initial interaction between the magician and the participant named Marlon. Use a JSON object to extract exactly two pieces of information: `magician\_request'(recognizing the specific words the magician says) and `participant\_compliance'(recording the participants' actions in detail). }
    
    \begin{categorybox}{}
        FORMAT CHECKS
    \end{categorybox}

    \begin{rulebox}{Format-001: Use a JSON object to extract exactly two pieces of information: `magician\_request' and `participant\_compliance'.}
        \small \textbf{Constraint:} json\_object \qquad 
        \textbf{Parameters:} "schema": \{"type": "object", "properties": \{"magician\_request": \{
      "type": "string"\}, "participant\_compliance": \{"type": "string"\} \}, "required": ["magician\_request", "participant\_compliance"] \}
    \end{rulebox}

    \begin{rulebox}{Format-002: extract exactly two pieces of information}
        \small \textbf{Constraint:} count \qquad 
        \textbf{Parameters:} "min\_count": 2, "max\_count": 2
    \end{rulebox}

    \begin{categorybox}{}
        CONTENT CHECKS
    \end{categorybox}

    \begin{openendedcontainer}{Content-001: Recognize the specific words the magician says for `magician\_request'}
        
        \textbf{Constraint:} \texttt{audio\_specific}
        \vspace{2pt}
        \begin{innerquestion}
            \small \textbf{Does the value for `magician\_request' contain quoted speech or a transcription of what the magician said?} \\[6pt]
            \begin{tabularx}{\linewidth}{@{}p{0.47\linewidth} @{\hspace{0.05\linewidth}} X @{}}
                A. Yes & B. No
            \end{tabularx} \\[6pt]
            \textit{Correct Answer: \textbf{Yes}}
        \end{innerquestion}

        \begin{innerquestion}
            \small \textbf{Which of the following quotes accurately reflects the magician's initial request to the participant?} \\[6pt] 
            \begin{tabularx}{\linewidth}{@{}p{0.47\linewidth} @{\hspace{0.05\linewidth}} X @{}}
                A. "Can you hand me a one-dollar bill?" & B. "I need you to give me a fifty."
            \end{tabularx} \\[4pt] 
            \begin{tabularx}{\linewidth}{@{}p{0.47\linewidth} @{\hspace{0.05\linewidth}} X @{}}
                C. "I want to do a trick with some money. You got your wallet?" & D. None of the above
            \end{tabularx} \\[6pt] 
            \textit{Correct Answer: \textbf{C}}
        \end{innerquestion}

    \end{openendedcontainer}
    
    \begin{openendedcontainer}{Content-002: Record the participants' actions in detail for `participant\_compliance'}
        
        \textbf{Constraint:} \texttt{visual\_events\_actions}
        \vspace{2pt}
        \begin{innerquestion}
            \small \textbf{Does the value for `participant\_compliance' describe the physical movements or actions of the participant?} \\[6pt]
            \begin{tabularx}{\linewidth}{@{}p{0.47\linewidth} @{\hspace{0.05\linewidth}} X @{}}
                A. Yes & B. No
            \end{tabularx} \\[6pt]
            \textit{Correct Answer: \textbf{Yes}}
        \end{innerquestion}

        \begin{innerquestion}
            \small \textbf{What specific actions does the participant take in response to the magician's request?} \\[6pt] 
            \begin{tabularx}{\linewidth}{@{}p{0.47\linewidth} @{\hspace{0.05\linewidth}} X @{}}
                A. He retrieves his wallet and hands it to the magician & B. He takes off his sunglasses and hands them to the magician
            \end{tabularx} \\[4pt] 
            \begin{tabularx}{\linewidth}{@{}p{0.47\linewidth} @{\hspace{0.05\linewidth}} X @{}}
                C. He cuts a dollar bill in half with scissors & D. None of the above
            \end{tabularx} \\[6pt] 
            \textit{Correct Answer: \textbf{A}}
        \end{innerquestion}
        
    \end{openendedcontainer}

\end{tcolorbox}

\begin{tcolorbox}[
    width=\linewidth, 
    enhanced,
    breakable,
    colback=white, 
    colframe=gray!80, 
    title={Data Sample - 3}, 
    colbacktitle=mainheader,
    fonttitle=\bfseries,
    arc=5pt
]
    \setlength{\tabcolsep}{0pt} 
    \renewcommand{\arraystretch}{0}
    \begin{tabularx}{\textwidth}{XXXX}
        \includegraphics[width=\linewidth]{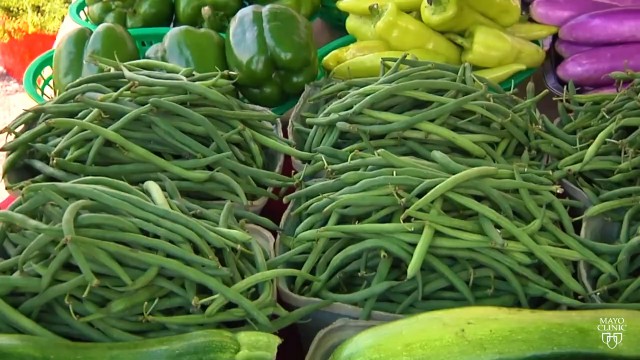} & 
        \includegraphics[width=\linewidth]{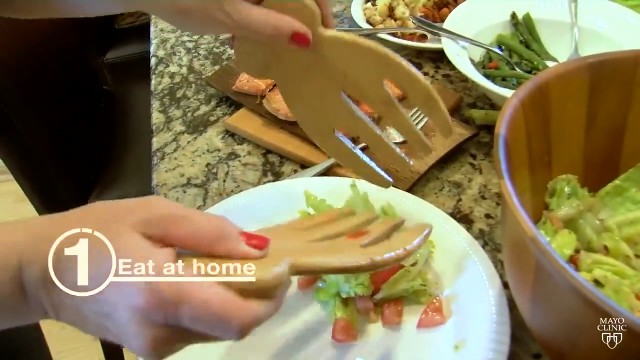} & 
        \includegraphics[width=\linewidth]{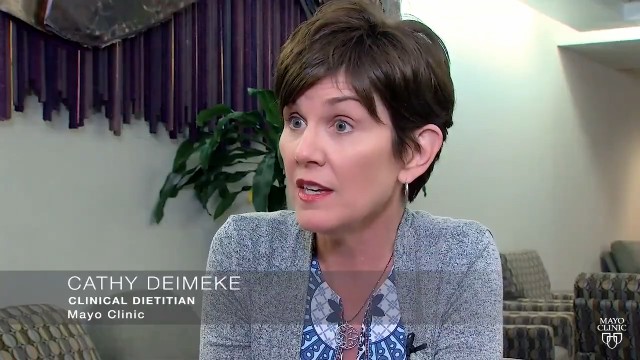} & 
        \includegraphics[width=\linewidth]{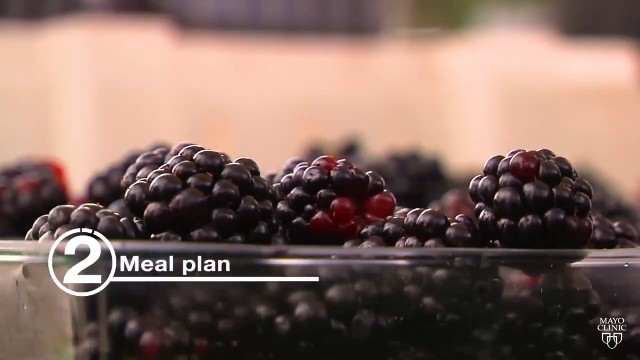} \\
        \includegraphics[width=\linewidth]{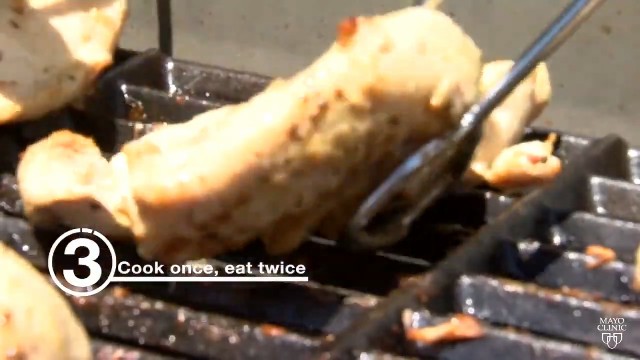} & 
        \includegraphics[width=\linewidth]{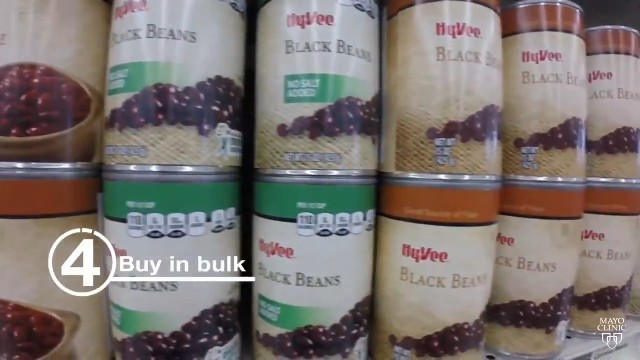} & 
        \includegraphics[width=\linewidth]{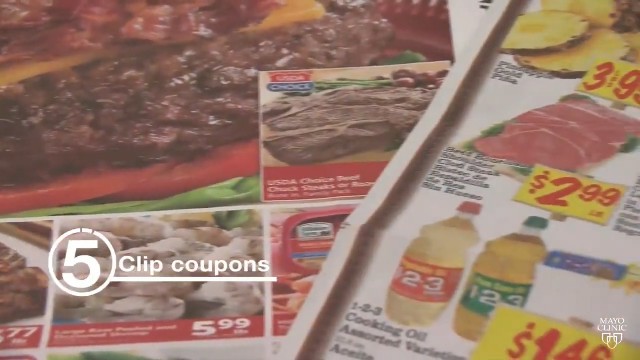} & 
        \includegraphics[width=\linewidth]{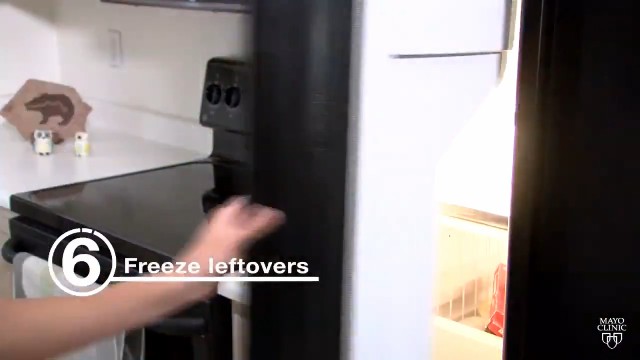}
    \end{tabularx}

    \maybeincludegraphics[width=\linewidth]{yingui/341.svg}
    
    \vspace{2pt}
    
    \textbf{Prompt:}\\
    \textit{Search for the `Cook once, eat twice' segment. Output in ALL CAPS using `|' as a delimiter. First, provide the timestamp interval [MM:SS-MM:SS] where the text overlay matches the spoken phrase; second, describe the grill marks on the chicken; third, infer if the speaker is the dietitian or the narrator based on the voice texture.}
    
    \begin{categorybox}{}
        FORMAT CHECKS
    \end{categorybox}

    \begin{rulebox}{Format-001: Output in ALL CAPS.}
        \small \textbf{Constraint:} case \qquad 
        \textbf{Parameters:} "schema": "upper"
    \end{rulebox}

    \begin{rulebox}{Format-002: Use `|' as a delimiter.}
        \small \textbf{Constraint:} delimiter \qquad 
        \textbf{Parameters:} "symbol": "|"
    \end{rulebox}

    \begin{rulebox}{Format-003: Provide the timestamp interval [MM:SS-MM:SS].}
        \small \textbf{Constraint:} timestamp\_format \qquad 
        \textbf{Parameters:} "format\_type": "period"
    \end{rulebox}

    \begin{categorybox}{}
        CONTENT CHECKS
    \end{categorybox}

    \begin{openendedcontainer}{Content-001: Provide the timestamp interval where the text overlay matches the spoken phrase}

        \textbf{Constraint:} \texttt{omni\_temporal\_grounding}
        \vspace{2pt}
        \begin{innerquestion}
            \small \textbf{According to the video, what is the timestamp interval where the text overlay 'Cook once, eat twice' appears and matches the spoken phrase?} \\[6pt]
            \textit{Correct Answer: \textbf{00:18 - 00:20}}
        \end{innerquestion}

    \end{openendedcontainer}
    
    \begin{openendedcontainer}{Content-002: Describe the grill marks on the chicken}
        
        \textbf{Constraint:} \texttt{visual\_specific}
        \vspace{2pt}
        \begin{innerquestion}
            \small \textbf{Does the response attempt to describe the grill marks on the chicken?} \\[6pt]
            \begin{tabularx}{\linewidth}{@{}p{0.47\linewidth} @{\hspace{0.05\linewidth}} X @{}}
                A. Yes & B. No
            \end{tabularx} \\[6pt]
            \textit{Correct Answer: \textbf{Yes}}
        \end{innerquestion}

        \begin{innerquestion}
            \small \textbf{Which of the following patterns correctly describes the grill marks on the chicken in the identified segment?} \\[6pt] 
            \begin{tabularx}{\linewidth}{@{}p{0.47\linewidth} @{\hspace{0.05\linewidth}} X @{}}
                A. Grid-like pattern & B. A circular branded pattern
            \end{tabularx} \\[4pt] 
            \begin{tabularx}{\linewidth}{@{}p{0.47\linewidth} @{\hspace{0.05\linewidth}} X @{}}
                C. A random, spotless browning & D. None of the above
            \end{tabularx} \\[6pt] 
            \textit{Correct Answer: \textbf{A}}
        \end{innerquestion}
        
    \end{openendedcontainer}

    \begin{openendedcontainer}{Content-003: Infer if the speaker is the dietitian or the narrator based on the voice texture}
        
        \textbf{Constraint:} \texttt{audio\_inference}
        \vspace{2pt}
        \begin{innerquestion}
            \small \textbf{Does the response infer whether the speaker is the dietitian or the narrator?} \\[6pt]
            \begin{tabularx}{\linewidth}{@{}p{0.47\linewidth} @{\hspace{0.05\linewidth}} X @{}}
                A. Yes & B. No
            \end{tabularx} \\[6pt]
            \textit{Correct Answer: \textbf{Yes}}
        \end{innerquestion}

        \begin{innerquestion}
            \small \textbf{Based on the voice texture, who is the speaker of the phrase 'Cook once, eat twice'?} \\[6pt] 
            \begin{tabularx}{\linewidth}{@{}p{0.47\linewidth} @{\hspace{0.05\linewidth}} X @{}}
                A. The Dietitian (Cathy Deimeke) & B. The Narrator
            \end{tabularx} \\[4pt] 
            \begin{tabularx}{\linewidth}{@{}p{0.47\linewidth} @{\hspace{0.05\linewidth}} X @{}}
                C. A male chef & D. Cannot be determined
            \end{tabularx} \\[6pt] 
            \textit{Correct Answer: \textbf{B}}
        \end{innerquestion}
        
    \end{openendedcontainer}
    
\end{tcolorbox}

\section{Error Analysis}
\label{sec:error}

As a supplement to the partial results analyzed in the main text, Figure \ref{fig:constraint_type_csr} presents the complete Constraint Success Rate (CSR) heatmap. This comprehensive visualization includes all evaluated models across the full spectrum of formatting and audio-visual constraint categories, serving as a complete reference for the overall performance landscape.

\begin{figure}[H]
\centering
\includegraphics[width=\linewidth]{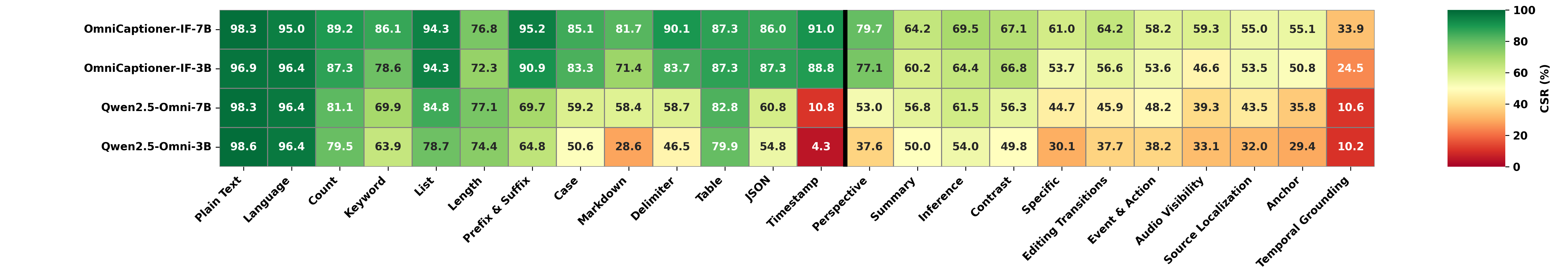} 
\caption{Comparison between OmniCaptioner-IF series and baselines.}
\label{fig:constraint_type_csr_ours}
\end{figure}

To thoroughly investigate the efficacy of our training approach, Figure \ref{fig:constraint_type_csr_ours} provides a direct and detailed comparison between the proposed \textbf{OmniCaptioner-IF} series and the \textbf{Qwen2.5-Omni} baselines. The most striking improvement is observed in the format constraints. While the baseline models exhibit severe deficiencies when handling strict structural formatting instructions, our OmniCaptioner-IF models effectively rectify these shortcomings, achieving consistently high Constraint Success Rates (CSR) across all formatting tasks.

Specifically, the baseline models struggle significantly with the Timestamp constraint, where Qwen2.5-Omni-7B and 3B fail almost entirely with CSRs of merely 10.8\% and 4.3\%. After our instruction-following tuning, OmniCaptioner-IF-7B and 3B achieve massive leaps to 91.0\% and 88.8\%, respectively. Similar transformative improvements are evident in other rigid formats such as Markdown (surging from 28.6\% to 71.4\% for the 3B model), Delimiter (from 46.5\% to 83.7\% for the 3B model), and JSON (from 60.8\% to 86.0\% for the 7B model). Furthermore, beyond format adherence, OmniCaptioner-IF also demonstrates substantial enhancements in complex audio-visual constraints, notably Temporal Grounding (increasing from 10.6\% to 33.9\% for the 7B model) and Perspective. This proves that our training method drastically bolsters comprehensive instruction-following capabilities, enabling the model to strictly adhere to both rigid formatting rules and fine-grained multimodal content requirements.

\begin{figure}[H]
\centering
\includegraphics[width=\linewidth]{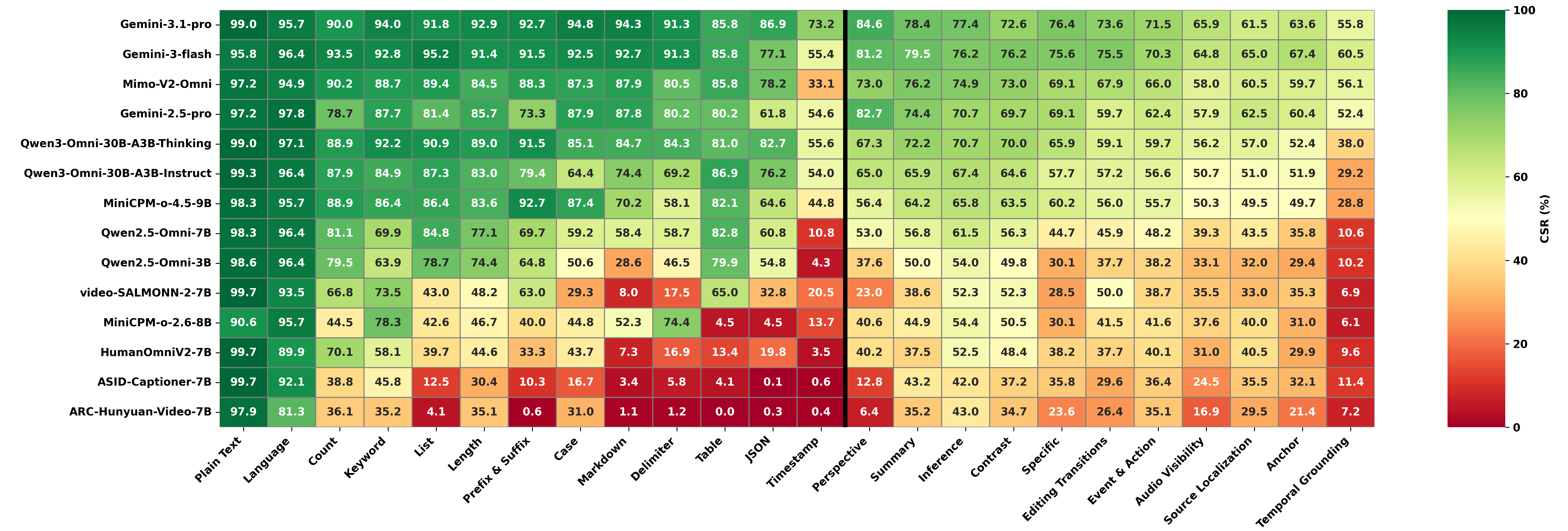} 
\caption{CSR performance of all models on different formats and audio-visual constraint types.}
\label{fig:constraint_type_csr}
\end{figure}

\begin{erroranalysiscontainer}{Example 1}

    \setlength{\tabcolsep}{0pt} 
    \renewcommand{\arraystretch}{0}
    \begin{tabularx}{\linewidth}{@{}XXXX@{}}
        \includegraphics[width=\linewidth]{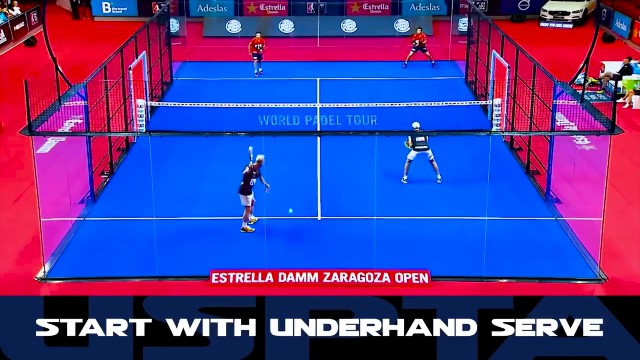} & \includegraphics[width=\linewidth]{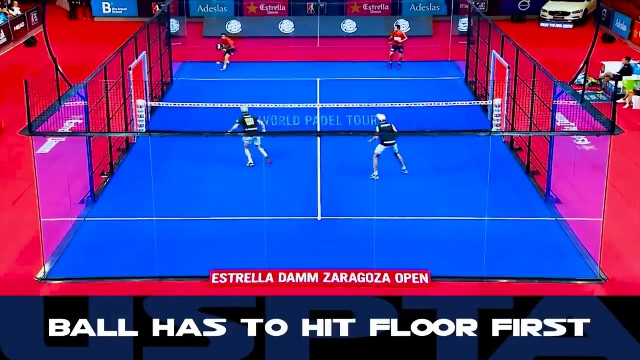} & \includegraphics[width=\linewidth]{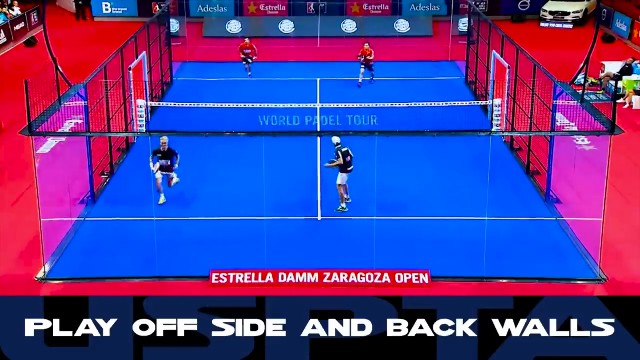} & \includegraphics[width=\linewidth]{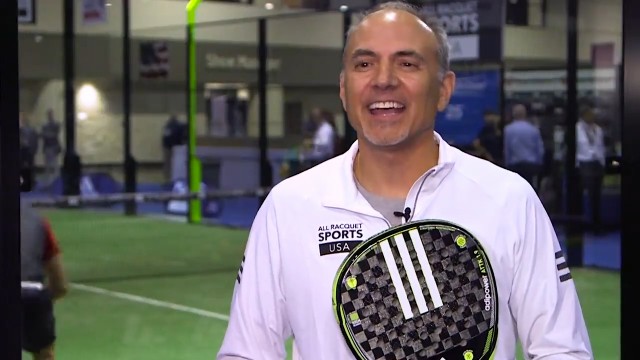} \\
    \end{tabularx}

    \maybeincludegraphics[width=\linewidth]{yingui/184.svg}
    \vspace{2pt}

    \small \textbf{Prompt:} \\
    \textit{Use a JSON object to analyze the `Rules of Padel' segment [00:13-00:25]. The object must contain exactly three keys: `rule\_text\_overlay' (transcribe the text on screen), `spoken\_explanation' (transcribe the commentary), and `consistency\_check' (describe if the audio elaborates or just reads the text).}

    \vspace{2pt}
    \small \textbf{Constraints:} json\_object, count, visual\_entities\_attributes, audio\_specific, omni\_contrast.

    \begin{error_analysis_captionbox}{Caption by Qwen2.5-Omni-7B}
        \{
          "rule\_text\_overlay": \{
            "start\_time": "00:13",
            "end\_time": "00:25",
            "content": "Ball has to hit floor first, play off side and back walls, thinking person's sport, easy to learn hard to master, play and have fun right away."
          \},
          "spoken\_explanation": "The speaker explains the rules of padel, emphasizing that the ball must first hit the floor before being played off the walls, highlighting that it is a thinking person's sport that is easy to learn but hard to master, and encouraging players to have fun immediately.",
          "consistency\_check": "The audio elaborates on the text by providing additional context and explanation about the rules and the nature of the game."
        \}
    \end{error_analysis_captionbox}

    \begin{error_analysis_analysisbox}{Error Analysis}{analysisbg}
        \begin{itemize}[leftmargin=*, topsep=0pt, itemsep=0pt, parsep=0pt]
            \item \textbf{Instruction Violation (Summarization vs. Transcription):} The prompt explicitly required the model to transcribe the commentary. The baseline model instead provided a third-person summary ("The speaker explains..."), failing to follow the specific audio\_specific constraint.
            \item \textbf{Content Error (Visual Hallucination):} For the "rule\_text\_overlay", the baseline hallucinated phrases that were not visually present on screen (e.g., "thinking person's sport", "easy to learn hard to master").
            \item \textbf{Format Violation (Schema Over-complication):} Instead of assigning a simple string value to "rule\_text\_overlay" as implied by the instruction, the baseline erroneously generated a nested JSON object containing "start\_time", "end\_time", and "content", deviating from the expected schema.
        \end{itemize}
    \end{error_analysis_analysisbox}

    \begin{error_analysis_trainedbox}{Caption by OmniCaptioner-IF-7B(ours)}
        \{
          "rule\_text\_overlay": "START WITH UNDERHAND SERVE, BALL HAS TO HIT FLOOR FIRST, PLAY OFF SIDE AND BACK WALLS",
          "spoken\_explanation": "You start with an underhand serve. After you serve, the ball has to bounce onto the turf, and you can play off the walls and play off the side.",
          "consistency\_check": "The audio elaborates on the rules by providing specific instructions on the serve and the bounce requirement, whereas the text overlays are static visual labels."
        \}
    \end{error_analysis_trainedbox}
    
\end{erroranalysiscontainer}

\begin{erroranalysiscontainer}{Example 2}

    \setlength{\tabcolsep}{0pt} 
    \renewcommand{\arraystretch}{0}
    \begin{tabularx}{\linewidth}{@{}XXXX@{}}
        \includegraphics[width=\linewidth]{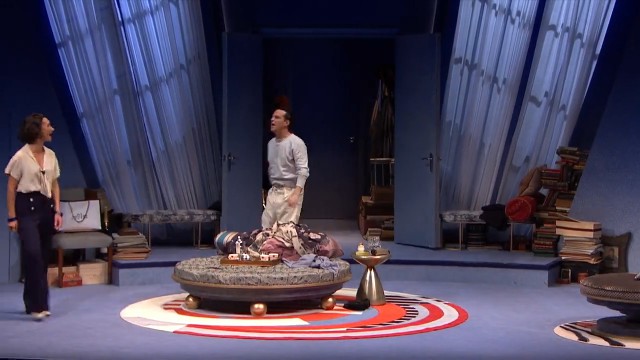} & \includegraphics[width=\linewidth]{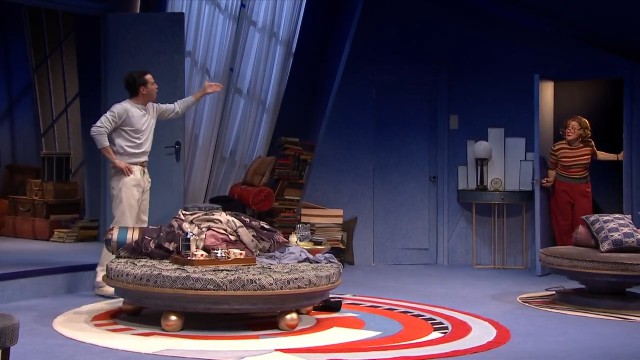} & \includegraphics[width=\linewidth]{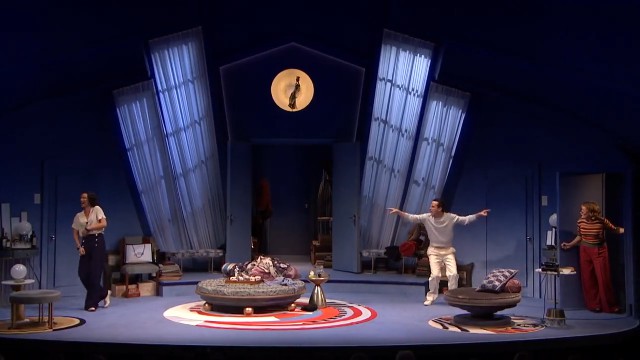} & \includegraphics[width=\linewidth]{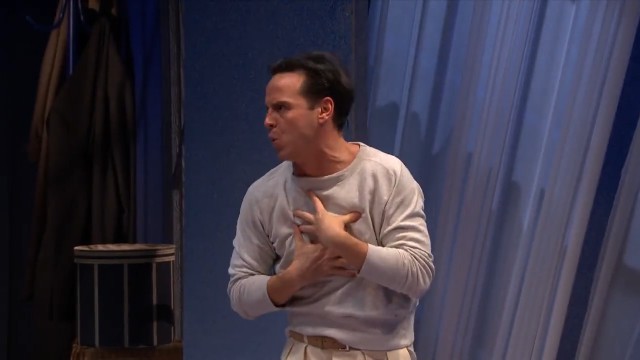} \\
    \end{tabularx}

    \maybeincludegraphics[width=\linewidth]{yingui/342.svg}
    
    \vspace{2pt}

    \small \textbf{Prompt:} \\
    \textit{Adopt the first-person perspective of the male character. First, describe your frustration with the woman in the white blouse; then, describe the interruption by the woman in red, specifically mentioning her top; finally, describe your return to the dramatic monologue. Use an ordered list starting with `I.', and for each point, include a specific timestamp [MM:SS] of the action.}

    \vspace{2pt}
    \small \textbf{Constraints:} ordered\_list, timestamp\_format, omni\_perspective, visual\_events\_actions, visual\_entities\_attributes, audio\_events\_actions.

    \begin{error_analysis_captionbox}{Caption by Qwen2.5-Omni-7B}
        1. I am frustrated with the woman in the white blouse. She is constantly interrupting me, and I feel like she doesn't understand my situation. This is evident from the timestamp [00:00.00] to [00:05.00].

        2. The woman in red, wearing a striped top, interrupts me again. She seems to be trying to get my attention, but I am not interested in her. This happens at the timestamp [00:05.00] to [00:10.00].

        3. I return to my dramatic monologue, expressing my belief in myself and my determination to succeed. This is shown at the timestamp [00:10.00] to [00:20.00].
    \end{error_analysis_captionbox}

    \begin{error_analysis_analysisbox}{Error Analysis}{analysisbg}
        \begin{itemize}[leftmargin=*, topsep=0pt, itemsep=0pt, parsep=0pt]
            \item \textbf{Format Violation (List Marker):} The prompt explicitly required an ordered list starting with \texttt{'I.'} (Roman numerals). The baseline model used Arabic numerals (\texttt{1., 2., 3.}), failing to adhere to the strict ordered\_list constraint.
            \item \textbf{Format Violation (Timestamp Syntax):} The instruction specified a precise timestamp format of \texttt{[MM:SS]} for each point. The baseline erroneously generated time ranges containing milliseconds (e.g., \texttt{[00:00.00] to [00:05.00]}), violating the timestamp\_format constraint.
            \item \textbf{Content Error (Action Misattribution):} The baseline misidentified the interrupting character and their intent. It incorrectly hallucinated that the woman in the white blouse was "constantly interrupting" the speaker, and falsely assumed the woman in red interrupted "again" to "get my attention". This shows a failure to accurately ground the actual event, where only the woman in red caused the interruption by bursting in.
        \end{itemize}
    \end{error_analysis_analysisbox}

    \begin{error_analysis_trainedbox}{Caption by OmniCaptioner-IF-7B(ours)}
        I. [00:00] I am walking through the room when I hear the woman in the white blouse shouting at me to go away, which is incredibly frustrating because I've only just come back from a long journey.
        II. [00:05] Suddenly, the woman in the red pants and striped top bursts through the door, shouting "Monica!" and interrupting my thoughts, causing me to stop and turn around in disbelief.
        III. [00:14] I quickly regain my composure and resume my dramatic monologue, gesturing wildly with my hands to emphasize my point about being the breadwinner and how my self-belief is being undermined.
    \end{error_analysis_trainedbox}
    
\end{erroranalysiscontainer}

\begin{erroranalysiscontainer}{Example 3}

    \setlength{\tabcolsep}{0pt} 
    \renewcommand{\arraystretch}{0}
    \begin{tabularx}{\linewidth}{@{}XXXX@{}}
        \includegraphics[width=\linewidth]{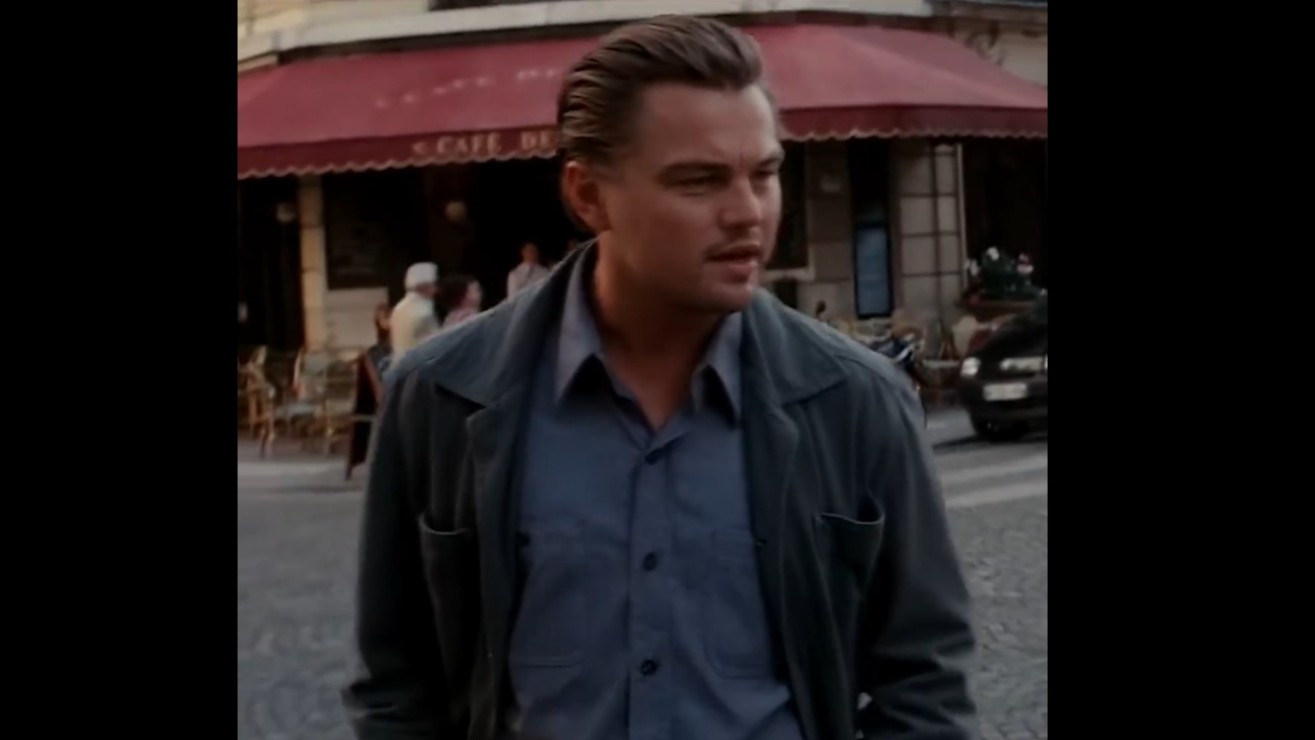} & \includegraphics[width=\linewidth]{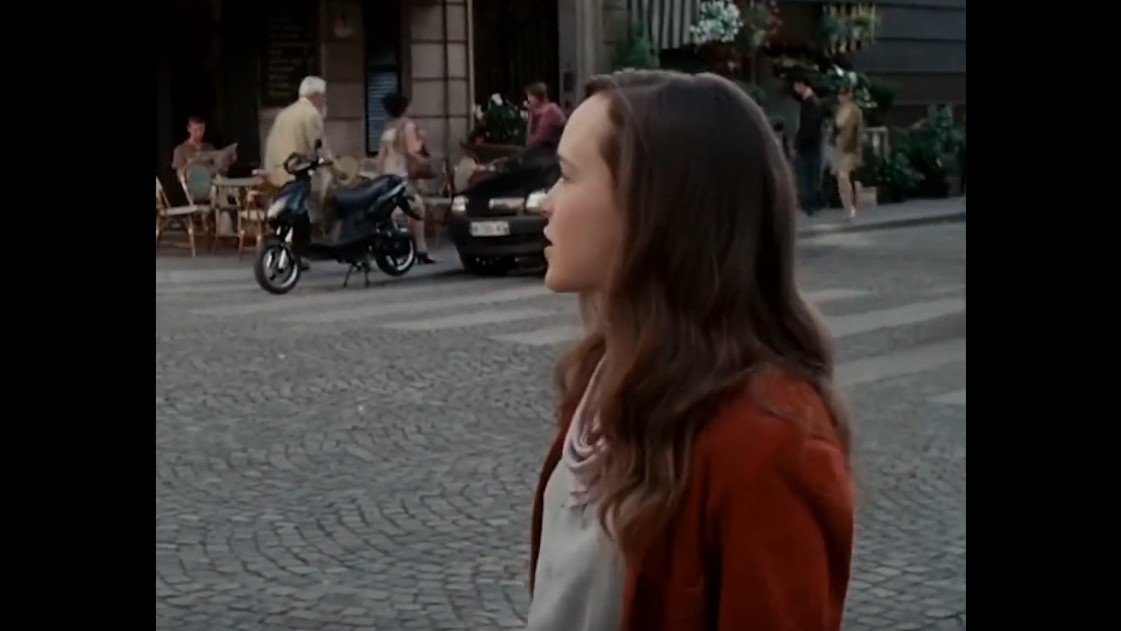} & \includegraphics[width=\linewidth]{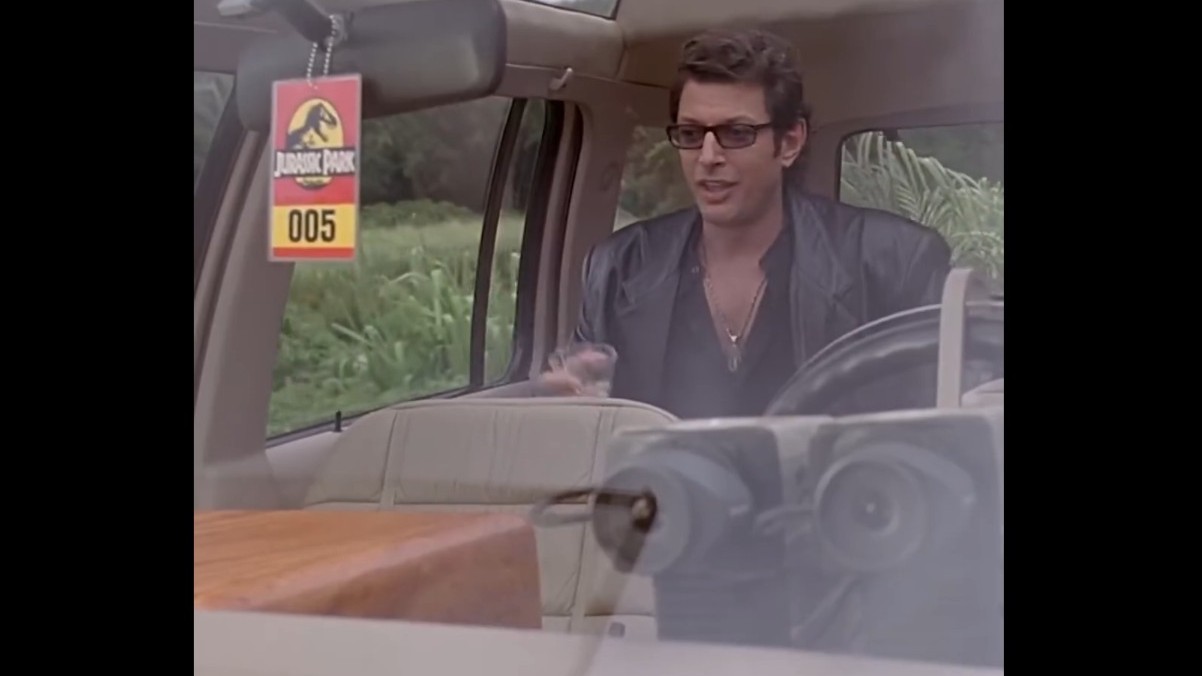} & \includegraphics[width=\linewidth]{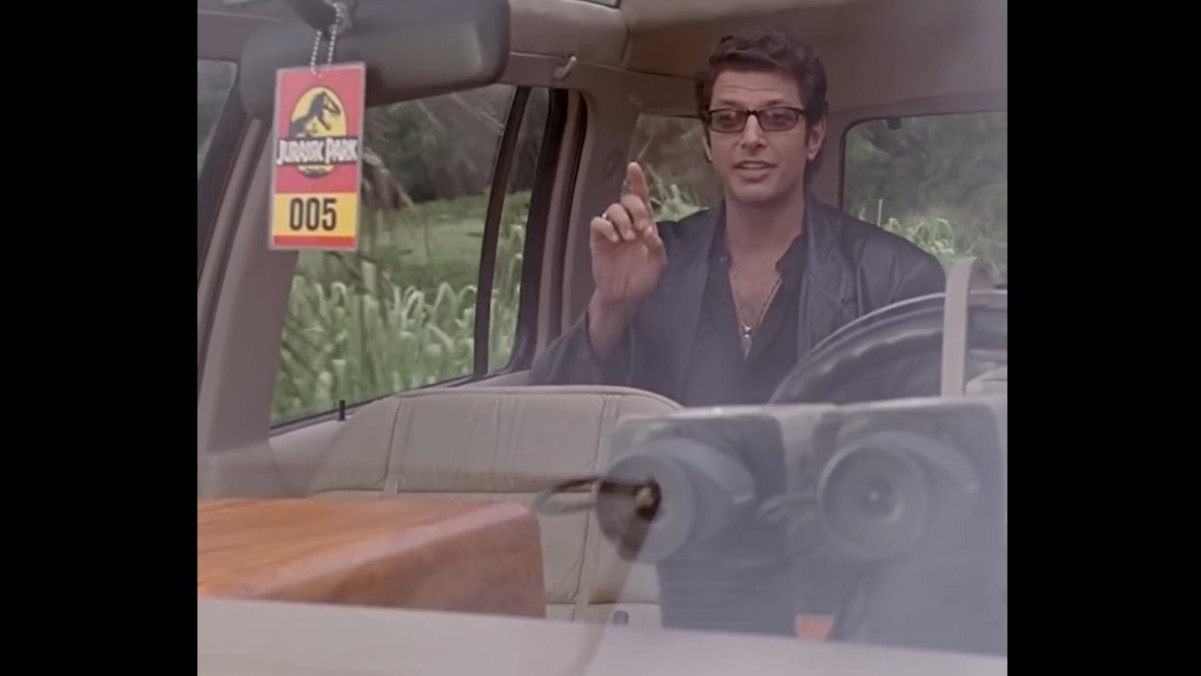} \\
    \end{tabularx}

    \maybeincludegraphics[width=\linewidth]{yingui/189.svg}
    
    \vspace{2pt}

    \small \textbf{Prompt:} \\
    \textit{Generate a production script for the jeep scene in a JSON array. Each element must contain: `time\_range' (using [MM:SS-MM:SS] format), `visual\_shot' (describe the character's sunglasses and hand gestures), `audio\_line' (transcribe the stuttering speech exactly), and `editing\_note' (describe any cuts). Locate the source of the speech and confirm if his lip movements synchronize with the audio.}

    \vspace{2pt}
    \small \textbf{Constraints:} json\_array, timestamp\_format, visual\_entities\_attributes, audio\_specific, omni\_editing\_transitions, omni\_source\_localization, omni\_temporal\_grounding.

    \begin{error_analysis_captionbox}{Caption by Qwen2.5-Omni-7B}
        [
          \{
            "time\_range": "00:36:44-00:37:00",
            "visual\_shot": "A man wearing sunglasses and making hand gestures",
            "audio\_line": "I drink your milkshake",
            "editing\_note": "No cuts"
          \}
        ]
    \end{error_analysis_captionbox}

    \begin{error_analysis_analysisbox}{Error Analysis}{analysisbg}
        \begin{itemize}[leftmargin=*, topsep=0pt, itemsep=0pt, parsep=0pt]
            \item \textbf{Format \& Content Error (Temporal):} The baseline failed in both syntax and grounding for the time\_range. It violated the timestamp\_format by outputting \texttt{"00:36:44-00:37:00"} (omitting brackets and adding hours), and completely hallucinated the timestamps (the actual event occurs at \texttt{[00:20-00:28]}), failing the omni\_temporal\_grounding constraint.
            \item \textbf{Content Error (Audio Hallucination):} Instead of transcribing the character's stuttering speech, the baseline completely hallucinated an unrelated famous movie quote ("I drink your milkshake"). This highlights a severe failure in audio\_specific grounding, relying on parametric memory instead of actual audio input.
            \item \textbf{Content Error (Missed Editing Transition):} For the editing\_note, the baseline incorrectly stated "No cuts," failing to detect the actual cut. This indicates a deficiency in omni\_editing\_transitions understanding.
            \item \textbf{Instruction Violation (Incomplete Task):} The prompt included a secondary instruction outside the JSON schema to locate the speech source and confirm lip synchronization. The baseline completely ignored this multi-task requirement, failing the omni\_source\_localization constraint.
        \end{itemize}
    \end{error_analysis_analysisbox}

    \begin{error_analysis_trainedbox}{Caption by OmniCaptioner-IF-7B(ours)}
        [
          \{
            "time\_range": "[00:20-00:28]",
            "visual\_shot": "The character wears dark sunglasses and gestures with his right hand while sitting in the driver's seat of a jeep.",
            "audio\_line": "See here I'm now by myself, talking to myself. That's, that's chaos theory.",
            "editing\_note": "The scene uses a medium shot with a slight cut to a close-up as the character begins his explanation."
          \}
        ]
        The lip movements of the character are perfectly synchronized with the audio, confirming the source is the character in the jeep.
    \end{error_analysis_trainedbox}
    
\end{erroranalysiscontainer}

\begin{erroranalysiscontainer}{Example 4}

    \setlength{\tabcolsep}{0pt} 
    \renewcommand{\arraystretch}{0}
    \begin{tabularx}{\linewidth}{@{}XXXX@{}}
        \includegraphics[width=\linewidth]{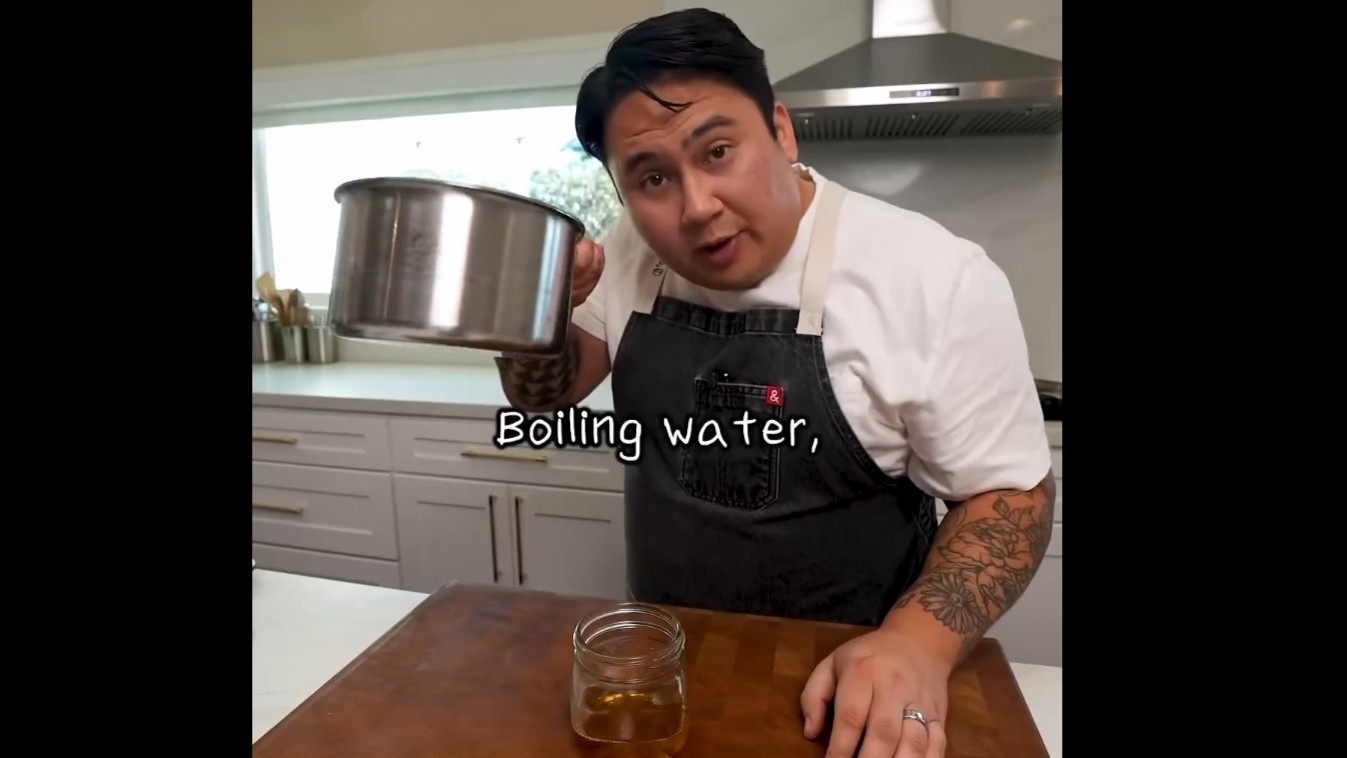} & \includegraphics[width=\linewidth]{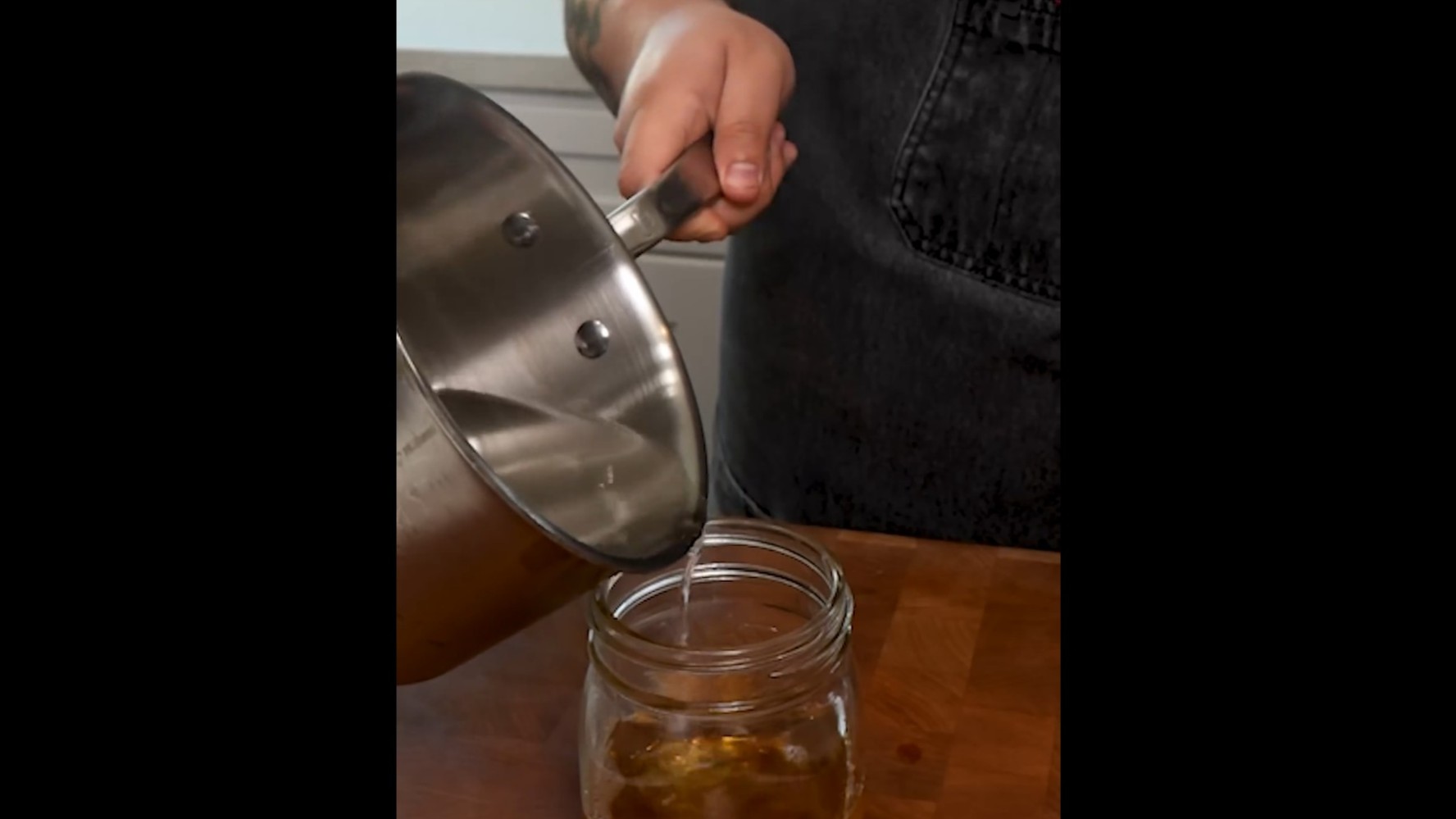} & \includegraphics[width=\linewidth]{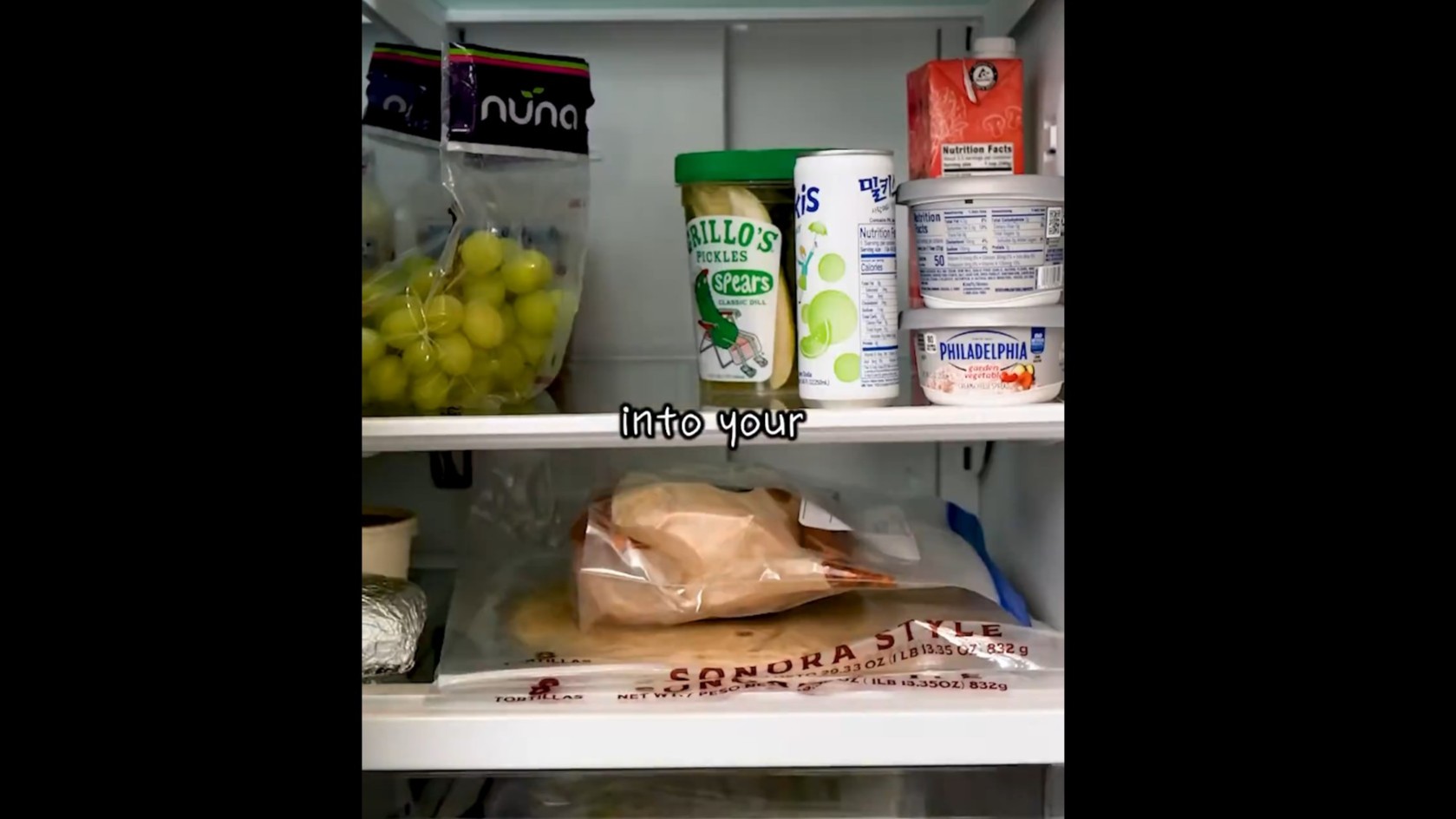} & \includegraphics[width=\linewidth]{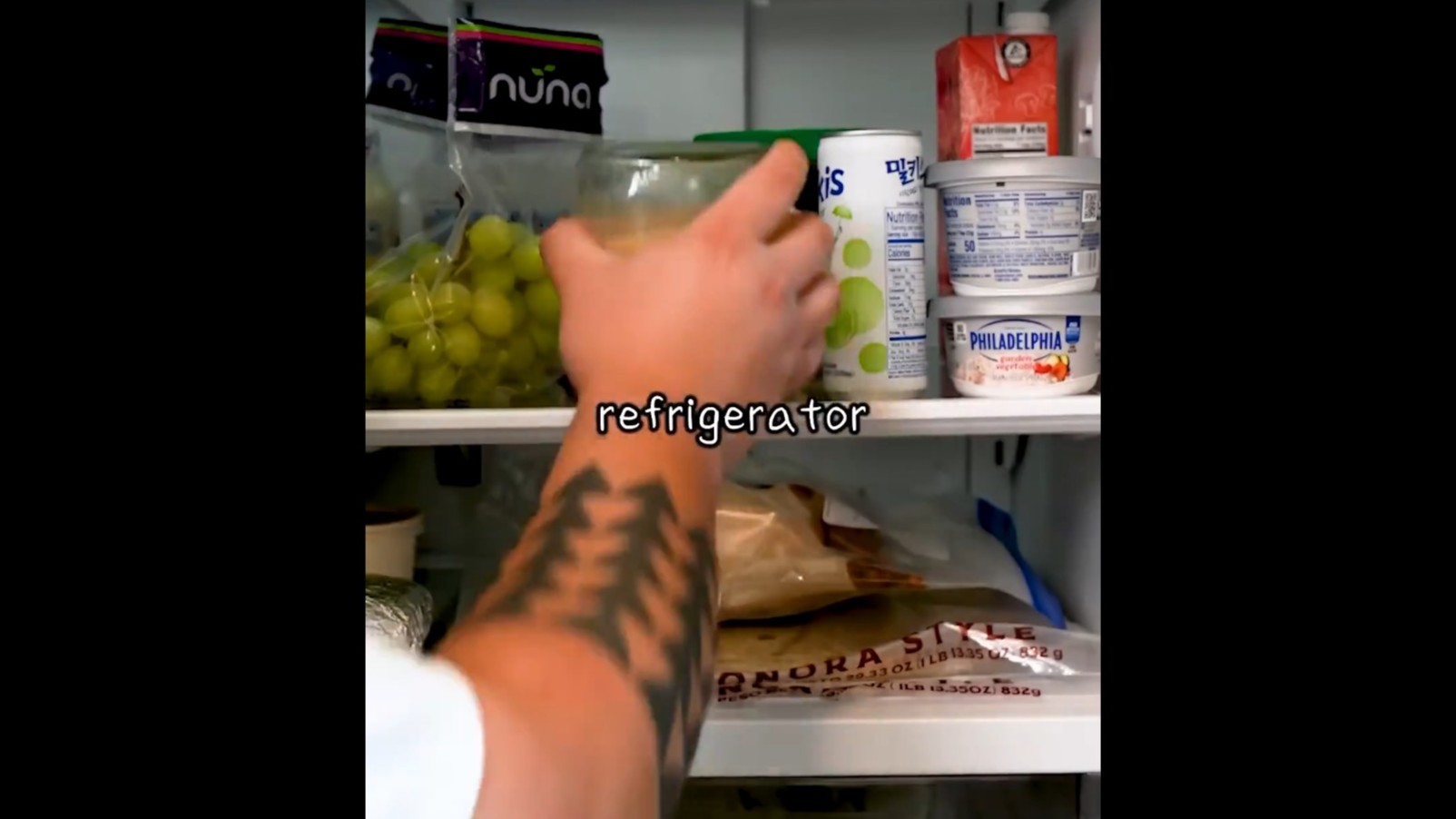} \\
    \end{tabularx}

    \maybeincludegraphics[width=\linewidth]{yingui/324.svg}
    
    \vspace{2pt}

    \small \textbf{Prompt:} \\
    \textit{Construct a table with columns `Audio Trigger', `Visual Action', `Speaker Visible', and `Timestamp'. Use the audio track as an anchor: If the chef says the phrase `boiling water', describe his specific hand action; if he says `refrigerator', describe the orientation of the jar. For the `Speaker Visible' column, determine if the chef's face is on-screen (True/False) during these specific audio cues. Transcribe the exact audio trigger phrase and use [MM:SS] format for the timestamp.}

    \vspace{2pt}
    \small \textbf{Constraints:} table, timestamp\_format, omni\_anchor, omni\_audio\_visibility, omni\_temporal\_grounding.

    \begin{error_analysis_captionbox}{Caption by Qwen2.5-Omni-3B}
        Empty row
        Timestamp | Audio Trigger | Visual Action | Speaker Visible
        --- | --- | --- | ---
        00:00:00.00 | Let's say that you just made a whole tray of absolutely delicious bacon | Chef holds up a piece of bacon | True
        00:00:00.00 | and after removing all of said bacon you're left with a whole lot of delicious bacon fat | Chef places the bacon on a paper towel | True
        00:00:00.00 | now you could just wait till it hardens and then throw it in the trash but that would be a waste or you could save it in a jar | Chef pours bacon fat into a jar | True
        00:00:00.00 | which simply starts by just pouring all that bacon grease into a jar | Chef seals the jar tightly | True
        00:00:00.00 | and then once it hardens you can just pour out all the water to get rid of all the nasty bits and scrape out any leftover gum | Chef pours the bacon fat into a bowl | True
        00:00:00.00 | and then you're left with clean usable bacon fat | Chef holds up the jar of bacon fat | True ......(repeat the above two lines until truncated)
        
    \end{error_analysis_captionbox}

    \begin{error_analysis_analysisbox}{Error Analysis}{analysisbg}
        \begin{itemize}[leftmargin=*, topsep=0pt, itemsep=0pt, parsep=0pt]
            \item \textbf{Instruction Violation (Audio Anchoring Failure):} The prompt explicitly required using specific audio phrases (e.g., "boiling water", "refrigerator") as triggers to extract targeted visual details. The baseline completely ignored this, outputting a generic chronological transcription instead, thereby failing the omni\_anchor constraint.
            \item \textbf{Format \& Content Error (Temporal Collapse):} The model suffered a catastrophic failure in temporal perception. It generated \texttt{00:00:00.00} for every single row. This not only violates the strict \texttt{[MM:SS]} timestamp\_format, but also demonstrates a complete loss of omni\_temporal\_grounding.
            \item \textbf{Generation Failure (Degenerative Loop):} Towards the end of the output, the baseline fell into an infinite repetition loop, endlessly duplicating the last two rows until generation was forcibly truncated. This indicates severe decoding instability and context-tracking failure in the 3B baseline model.
        \end{itemize}
    \end{error_analysis_analysisbox}

    \begin{error_analysis_trainedbox}{Caption by OmniCaptioner-IF-3B(ours)}
        | Audio Trigger | Visual Action | Speaker Visible | Timestamp |
        | :--- | :--- | :--- | :--- |
        | Boiling water | The chef pours the bacon grease into a glass jar and then uses his right hand to swirl the mixture inside the jar. | True | [00:21] |
        | Refrigerator | The chef places the sealed jar upside down into the refrigerator. | True | [00:31] |
        | Bacon fat | The chef pours the liquid from the jar into a metal bowl. | True | [00:35] |
    \end{error_analysis_trainedbox}
    
\end{erroranalysiscontainer}

\begin{erroranalysiscontainer}{Example 5}

    \setlength{\tabcolsep}{0pt} 
    \renewcommand{\arraystretch}{0}
    \begin{tabularx}{\linewidth}{@{}XXXX@{}}
        \includegraphics[width=\linewidth]{supp_figures/001-1.jpg} & \includegraphics[width=\linewidth]{supp_figures/001-2.jpg} & \includegraphics[width=\linewidth]{supp_figures/001-3.jpg} & \includegraphics[width=\linewidth]{supp_figures/001-4.jpg} \\
    \end{tabularx}

    \maybeincludegraphics[width=\linewidth]{yingui/001.svg}
    
    \vspace{2pt}

    \small \textbf{Prompt:} \\
    \textit{Analyze the audio scene around 00:08. Use a Markdown table with columns `Timestamp' and `Audio Content'. Record the exact moment of the spoken line "Not in my house". If a squishing sound is heard, explain what this sound represents after the Markdown table.}

    \vspace{2pt}
    \small \textbf{Constraints:} table, keyword, audio\_specific, audio\_temporal\_grounding, audio\_inference.

    \begin{error_analysis_captionbox}{Check result of Gemini-3-flash (single-modal input)}

        \small \textbf{According to the video, at what time does the character start saying the line "Not in my house"?} \\[6pt]
        \textit{Model's Answer: \textbf{00:08}} \qquad \textit{Correct Answer: \textbf{00:13}} \qquad \textit{Result: false}

    \end{error_analysis_captionbox}

    \begin{error_analysis_analysisbox}{Error Analysis}{analysisbg}
        \begin{itemize}[leftmargin=*, topsep=0pt, itemsep=0pt, parsep=0pt]
            \item \textbf{Content Error (Prompt Bias / Shortcut Learning):} The original prompt included a contextual hint ("Analyze the audio scene around 00:08"). Deprived of visual anchors, the single-modal model suffered from prompt bias, lazily copying \texttt{00:08} as the answer instead of genuinely localizing the true speech onset (\texttt{00:13}).
            \item \textbf{Performance Degradation (Single-Modal Limitation):} The failure of the single-modal input contrasts sharply with the success of the omni-modal input. This perfectly exemplifies our finding: \textit{when only audio is provided, the models' \texttt{audio\_temporal\_grounding} capability is significantly weaker than when both audio and visual modalities are available}. Visual cues (e.g., character actions, lip movements) provide critical cross-modal synergy for accurate temporal alignment.
        \end{itemize}
    \end{error_analysis_analysisbox}

    \begin{error_analysis_trainedbox}{Check result of Gemini-3-flash (omni-modal input)}
    
        \small \textbf{According to the video, at what time does the character start saying the line "Not in my house"?} \\[6pt]
        \textit{Model's Answer: \textbf{00:13}} \qquad \textit{Correct Answer: \textbf{00:13}} \qquad \textit{Result: true}
        
    \end{error_analysis_trainedbox}
    
\end{erroranalysiscontainer}

\section{Prompts}

\subsection{Test}
\label{sec:system_prompt}
\begin{promptbox}{System Prompt for Model Testing.}
As a professional video describer, your task is to provide an accurate, controlled omni-modal caption based on both visual and audio information from the video. The video is sampled at a rate of 1 frame per second. Strictly follow the instructions without adding any opening remarks, closing statements, or additional explanations. Here are your instructions:
\end{promptbox}
This text serves as the system prompt for the model. The frame sampling rate specified in the prompt may vary depending on the model, and all video input parameters, including frame sampling, are provided in Section~\ref{sec:eval}.

\subsection{Judge}
\label{sec:judge_prompt}
Content extraction for format check items and question answering for content check items are both performed using gpt-5-mini. The specific prompts are as follows:

\begin{promptbox}{The Prompt for Format Extraction}

\textbf{Roles and Goals}\\
You are a highly accurate and strictly rule-following content extraction program. You need to extract the correct content to be checked for a given format checking function. Your sole task is: based on the requirements and related instructions of the given checkitem, extract content from the input response, and configure the content parameter in the checkitem.

\textbf{Input Information}\\
You will receive a JSON object containing the following three key pieces of information:
\begin{itemize}[leftmargin=15pt, nosep]
    \item \texttt{response}: A string, the source from which to extract content.
    \item \texttt{prompt}: string, the original instruction that generates the response, serving as the full contextual reference for the \texttt{check\_description}.
    \item \texttt{checkitem}: A JSON object, which specifies the check type and content for which you need to perform the content extraction.
\end{itemize}

\textbf{CheckItem Structure Description}\\
The format of the CheckItem you receive is as follows:
\begin{lstlisting}[language=json]
{
    "check_id": "string", // The unique ID of the check item
    "constraint_id": "string", // The constraint type ID, which specifies the program corresponding to this format check
    "check_description" : "string",  // The check description, which describes the requirements of the instruction
    "parameters": {
        "content": null, // Awaiting content extraction, should be a list of strings
        // ... Copy other dynamic specific parameters
    }
}
\end{lstlisting}

\textbf{\texttt{constraint\_id} Type Description}\\
The following explains each \texttt{constraint\_id} type and the execution logic of the corresponding rule script (Note: content below refers to one element in its list):
\begin{itemize}[leftmargin=15pt, nosep]
    \item \texttt{plain\_text}: Plain text check. The script determines if the content contains special structural symbols.
    \item \texttt{json\_object}: JSON object check. The script determines if the content conforms to the JSON structure specified by the schema.
    \item \texttt{json\_array}: JSON array check. The script determines if the content conforms to the JSON structure specified by the schema.
    \item \texttt{unordered\_list}: Unordered list check. The script determines if, after splitting the content by newlines, each element is a paragraph starting with the symbol.
    \item \texttt{ordered\_list}: Ordered list check. The script determines if, after splitting the content by newlines, each element is a paragraph starting with the symbol and its incrementing counterpart.
    \item \texttt{table}: Table check. The script determines if the content satisfies the table syntax and has the column names specified by \texttt{col\_name}.
    \item \texttt{keyword}: Keyword check. The script checks whether the content includes or excludes a specific fixed keyword string.
    \item \texttt{timestamp\_format}: Timestamp format check. The script determines if the content conforms to the specified timestamp structure.
    \item \texttt{markdown}: Markdown decoration syntax check. The script determines if the prefix and suffix of the content satisfy the specified markdown decoration syntax (e.g., '**' for bold).
    \item \texttt{prefix\_suffix}: Prefix and suffix check. The script determines if the prefix and suffix of the content satisfy the specified parameters.
    \item \texttt{delimiter}: Delimiter check. The script determines if the content contains the delimiter specified in the parameters.
    \item \texttt{length}: Length check. The script determines if the length of the content, as divided by the unit specified in the parameters (character, word, sentence, or paragraph), meets the specified range.
    \item \texttt{count}: Count check. The script determines if the number of objects enclosed in \texttt{()} in the content meets the specified range.
    \item \texttt{case}: Case check. The script determines if the content meets the specified case (e.g., uppercase).
    \item \texttt{language}: Language check. The script determines if the content meets the specified language.
\end{itemize}

\textbf{Special Notes}
\begin{enumerate}[leftmargin=15pt, nosep]
    \item \texttt{content} is a list of strings, where each element is treated as an object for one format check execution. The script will execute the check corresponding to the \texttt{constraint\_id} for each element in content and ultimately fill the result parameter with the logical 'AND' of all check results. 
    \item For the \texttt{count} constraint check, each element of the content list (i.e., representing a check on one continuous piece of content) should be a string containing multiple objects separated by \texttt{()} and \texttt{,}, for example, \texttt{content:["(a),(b),(c)"]}.
    \item Note: When extracting the \texttt{count} content, semantic alignment must be considered. If the response contains 5 items but the checkitem expects 4, all 5 items should still be extracted for an accurate evaluation.
    \item \texttt{content} must be fully and accurately extracted from the response to be checkable by the script. The corresponding script checking mechanism must be fully considered during extraction. For example: if checking for bold, the '**' prefix and suffix should be included.
    \item When extracting content, pay attention to the applicable scope of the checks. For example, if checking the length or language of a list item's content, avoid extracting the list's starting symbol (e.g., -, A. ) to prevent it from affecting the inspection.
\end{enumerate}

\textbf{Output Specification}\\
You must return a single, valid JSON object, which is the content you actually extracted. It must not contain any additional explanatory text.
\begin{lstlisting}[language=json]
{
    "content": ["string"] // The content you actually extracted
}
\end{lstlisting}

\textbf{Positive and Negative Case Comparison}

\textit{Example 1:}\\
\textbf{response}: ``Here are the video descriptions:\textbackslash n\textbackslash n* Car\textbackslash n\textbackslash nA. Screens on posts change from green to red.\textbackslash nB. Car flips over.''\\
\textbf{checkitem}:
\begin{lstlisting}[language=json]
{
    "check_id": "format-002",
    "constraint_id": "ordered_list",
    "check_description": "Describe the two distinct state changes shown on the screens of the yellow posts using an ordered list starting with 'A.'.",
    "parameters": { "content": null, "symbol": "A." }
}
\end{lstlisting}
\textbf{Correct Extraction}: \texttt{\{ "content": ["A. Screens on posts change from green to red.\textbackslash nB. Car flips over."] \}}\\
\textbf{Incorrect Extraction}: \texttt{\{ "content": ["A. Screens on posts change from green to red.", "B. Car flips over."] \}}\\
\textbf{Reason for Error}: An ordered list check should treat the entire continuous ordered list as a single element, not as multiple elements from the list.

\vspace{5pt}
\textit{Example 2:}\\
\textbf{response}: 
\begin{lstlisting}[language=json]
{
  "title": "Link running towards distant volcano, Breath of the Wild scene.",
  "tags": {
    "character_attire": "Green tunic, beige pants",
    "action": "Running",
    "landmark": "Volcano"
  }
}
\end{lstlisting}
\textbf{checkitem}: \textit{(Briefly: json\_object check for specific keys 'title' and 'tags'.)}\\
\textbf{Correct Extraction}:
\texttt{\{ "content": ["\{\textbackslash n  \textbackslash "title\textbackslash ": \textbackslash "Link running towards distant volcano...\textbackslash ", \textbackslash "tags\textbackslash ": \{ ... \} \textbackslash n\}"] \}}\\
\textbf{Incorrect Extraction}: \texttt{\{ "content": ["title, tags"] \}}\\
\textbf{Reason for Error}: A JSON type check must extract the entire relevant content completely and then hand it over to the script for checking.

\vspace{5pt}
\textit{Example 3:}\\
\textbf{response}: ``Here are the video descriptions:\textbackslash n\textbackslash n* Car\textbackslash n\textbackslash nA. Screens on posts change from green to red.\textbackslash nB. Car flips over.''\\
\textbf{checkitem}: \textit{(Briefly: count check for 2 distinct state changes.)}\\
\textbf{Correct Extraction}: \texttt{\{ "content": ["(Screens on posts change from green to red.),(Car flips over.)"] \}}\\
\textbf{Incorrect Extraction}: \texttt{\{ "content": ["A. Screens on posts change from green to red.\textbackslash n B. Car flips over."] \}}\\
\textbf{Reason for Error}: The check type is \texttt{count}, which focuses on the 'two distinct state changes' in the constraint description, not the structural check of the ordered list.

\vspace{5pt}
\textit{Example 4:}\\
\textbf{response}: ``The video begins with a view of a drawer containing various items...'' (No table present)\\
\textbf{checkitem}: \texttt{\{ "constraint\_id": "count", "check\_description": "The table must contain exactly 3 rows" \}}\\
\textbf{Correct Extraction}: \texttt{\{ "content": [""] \}}\\
\textbf{Incorrect Extraction}: \texttt{\{ "content": ["(a white power bank),..."] \}}\\
\textbf{Error cause}: The check description requires locating rows in a table. Since the original description contains no table at all, an empty extraction should be performed.

\vspace{5pt}
\textit{Example 5:}\\
\textbf{response}: ``person A, person B, person C, person D''\\
\textbf{checkitem}: \texttt{\{ "constraint\_id": "count", "check\_description": "describe 3 person" \}}\\
\textbf{Correct Extraction}: \texttt{\{ "content": ["(person A), (person B), (person C), (person D)"] \}}\\
\textbf{Incorrect Extraction}: \texttt{\{ "content": ["(person A), (person B), (person C)"] \}}\\
\textbf{Reason for Error}: When extracting, all items should be retrieved, disregarding the requirements of min\_count and max\_count, in order to ensure an accurate evaluation.

\vspace{5pt}
Next, carefully read the actual input below and perform the processing that conforms to the conventions and is reasonable:
\end{promptbox}

\begin{promptbox}{The Prompt for Content Evaluation}

\textbf{Role and Goal}\\
You are a Video Caption Evaluator. You are required to perform specified checks on a provided caption. Based on the specific question and the reference description of the video's content, you will judge the response based on two criteria.

\textbf{Input Information}\\
You will receive a JSON object containing the following four key pieces of information:
\begin{itemize}[leftmargin=15pt, nosep]
    \item \texttt{prompt}: A string, providing instructions to generate a response from the model.
    \item \texttt{response}: A string, which is the caption from the model being evaluated.
    \item \texttt{question}: A string, which is the question that needs to be answered.
    \item \texttt{options}: A list of strings, which are the available choices for you to select from (single choice).
\end{itemize}

\textbf{Task Description}\\
You need to answer the question based on the actual content of the response (Choose the answer that is closest to the options.). After providing your answer, you must fill in the \texttt{result\_explanation} field with an explanation and the \texttt{result\_confidence} field with a confidence score (on a scale of [1-5]).

\textbf{Output Specification}\\
You must return a single and valid JSON object, which is your completed answer and explanation. It must not contain any additional explanatory text.
\begin{lstlisting}[language=json]
{
    "answer": "string", // The result you provide. If 'options' is a multiple-choice question, this will be A, B, C, or D. If it is a true/false question, it will be 'yes' or 'no'.
    "result_explanation": "string", // Your explanation.
    "result_confidence": "integer" // Your confidence score.
}
\end{lstlisting}

\vspace{5pt}
Next, carefully read the actual input below and perform the processing that conforms to the conventions and is reasonable:
\end{promptbox}

\subsection{Construction of Prompts for The Test Set}
\label{sec:test_set_prompt}
In this section, each actual prompt consists of the prompt shown below plus the constraint system table in Section~\ref{sec:system} serving as the actual content of the Core Knowledge Base.

\begin{promptbox}{The Prompt for Format Checklist Generation}

\textbf{\# Role and Goal}\\
You are an extremely rigorous and logical checklist constructor. Based on the user-provided prompt and constraints\_used, you need to generate a detailed, precise evaluation checklist JSON object that fully adheres to all the rules below.

\textbf{\# Core Knowledge Base}\\
You must strictly follow the knowledge base named ‘Constraint System’. This knowledge base defines the technical details for all constraint\_ids.

\textbf{\# Strict Constraint Boundary (CRITICAL)}\\
You MUST COMPLETELY IGNORE any Content or Semantic constraints (i.e., ANY constraint ID starting with "visual\_", "audio\_", or "omni\_"). Do NOT generate any check items for them.\\
ALLOWED FORMAT IDs: ["json\_object", "json\_array", "markdown", "plain\_text", "table", "ordered\_list", "unordered\_list", "length", "count", "keyword", "case", "language", "prefix\_suffix", "delimiter", "timestamp\_format"]\\
Do not omit any of the format constraints mentioned above.

\textbf{\# Format Constraint Parameter Table}\\
Below, we will use Python syntax to define the checking parameters corresponding to each format constraint ID:
\begin{itemize}[leftmargin=15pt, nosep]
    \item plain\_text(content: str)
    \item json\_object(content: str, schema: Dict)
    \item json\_array(content: str, schema: Dict)
    \item unordered\_list(content: str, symbol: Optional[Literal['-', '*']] = None)
    \item ordered\_list(content: str, symbol: Optional[Literal['1.', 'a.', 'A.', 'I.']] = None)
    \item table(content: str, col\_name: List[str])
    \item keyword(content: str, keyword: str, keyword\_type: Literal['include', 'exclude'])
    \item timestamp\_format(content: str, format\_type: Literal['point', 'period'])
    \item markdown(content: str, md\_type: Literal['title', 'bold', 'highlight', 'italic', 'code'])
    \item prefix\_suffix(content: str, prefix: Optional[str] = None, suffix: Optional[str] = None)
    \item delimiter(content: str, symbol: str)
    \item length(content: str, unit: Literal["character", "word", "sentence", "paragraph"], min\_len: int = 0, max\_len: int = -1)
    \item count(content: str, min\_count: int = 0, max\_count: int = -1)
    \item case(content: str, case\_type: Literal["upper", "lower", "title"])
    \item language(content: str, lang\_type: Literal["en", "zh"])
\end{itemize}
Specifically, the "schema" parameter used for checking JSON formats must conform to the JSON Schema standard to describe the specified JSON format, which will be validated using the "jsonschema" package.\\
Please pay special attention to the differences between a few easily confused constraint types:
\begin{itemize}[leftmargin=15pt, nosep]
    \item "table" vs. "markdown": The "markdown" constraint specifically refers to text decoration syntax such as bolding, highlighting, etc., while the "table" constraint specifically refers to table structures built using Markdown syntax.
    \item "json\_object" vs. "json\_array": In many scenarios, both will appear simultaneously. However, when performing format checks, due to the universality of JSON Schema, you only need to configure the top-level type check (for example, if an object is required to contain a list, configure the "json\_object" check; if a list is required to contain objects, configure the "json\_array" check).
    \item "length" vs. "count": "length" can only check the length of text in units of "character", "word", "sentence", or "paragraph", while "count" is used for counting semantic elements, such as the number of objects, the number of actions, etc.
\end{itemize}

\textbf{\# Input Information}\\
You will receive a JSON object containing the following three key pieces of information:
\begin{itemize}[leftmargin=15pt, nosep]
    \item prompt: A string, which contains the constraint instructions for the model under test.
    \item constraints\_used: An array of strings, which lists the unique IDs of all atomic constraints used to generate this prompt.
\end{itemize}

\textbf{\# Core Task: Generate Checklist}\\
You must strictly follow the process to generate the content for the format\_check section.
\begin{enumerate}[leftmargin=15pt, nosep]
    \item Iterate through the input constraints\_used list. For each formatting or structural constraint (e.g., json\_object, length, unordered\_list, etc.), create check items (A constraint may correspond to multiple check items). Do not omit any format constraints.
    \item ID and Description:
    \begin{itemize}[nosep]
        \item check\_id: Sequentially generate a unique ID, starting from "format-001".
        \item constraint\_id: Copy this directly from the constraints\_used list.
        \item check\_description: Precisely extract the instructional description related to the current constraint\_id from the prompt.
    \end{itemize}
    \item Parameter Extraction
    \begin{itemize}[nosep]
        \item Locate: First, locate the descriptive sentence or phrase in the prompt text that directly corresponds to the current constraint\_id.
        \item Extract: Then, precisely extract the parameter values from this located phrase. For example, for a length constraint, you should first locate the phrase "no more than 150 words" and then extract \{ "max": 150, "unit": "word" \} from it.
        \item Assign: Assign the extracted values to the corresponding parameters. The value for the content key must always be null. For JSON objects and lists, their schema parameter must conform to the JSON Schema specification.
    \end{itemize}
\end{enumerate}

\textbf{\# Output Specification}\\
You must return a single, valid JSON object that represents the complete structure for format\_check. It must not contain any additional explanatory text.
\begin{lstlisting}[language=json]
{
    "format_check":[    // List of format check items
        {
            "check_id": "string", // Unique ID for the format check, e.g., "format-001"
            "constraint_id": "string", // The corresponding original format constraint ID, e.g., "length"
            "check_description" : "string",  // Description of the check item
            "parameters": { // Parameters to be checked, extracted from the prompt
            "content": null, // Content is left null, to be extracted by the Check model
            // ... other dynamic and specific parameters, e.g., "max": 150, "unit": "word"
            }
        }
    ]
}
\end{lstlisting}

\textbf{\# Concrete Example}\\
The following are complete examples, including inputs and expected outputs, for your reference.\\
\textit{\#\# Example 1}\\
Input:
\begin{lstlisting}[language=json]
{
    "prompt": "Please output a JSON object, which must contain the keys 'summary' and 'key_actions'. The value for the 'summary' key should be a video summary of no more than 30 words. The value for the 'key_actions' key should be an unordered list using '-' as the bullet point, listing all key actions of the main character.",
    "constraints_used": [
        "json_object",
        "omni_summary",
        "visual_events_actions",
        "length",
        "unordered_list"
    ]
}
\end{lstlisting}
Expected Output:
\begin{lstlisting}[language=json]
{
    "format_check": [
        {
            "check_id": "format-001",
            "constraint_id": "json_object",
            "check_description": "Output a JSON object, which must contain the keys 'summary' and 'key_actions'.",
            "parameters": {
                "content": null,
                "schema": {
                    "type": "object",
                    "properties": {
                        "summary": { "type": "string" },
                        "key_actions": { "type": "string" }
                    },
                    "required": ["summary", "key_actions"]
                }
            }
        },
        {
            "check_id": "format-002",
            "constraint_id": "length",
            "check_description": "The value for the 'summary' key should be a video summary of no more than 30 words.",
            "parameters": {
                "content": null,
                "unit": "word",
                "max_len": 30
            }
        },
        {
            "check_id": "format-003",
            "constraint_id": "unordered_list",
            "check_description": "The value for the 'key_actions' key should be an unordered list using '-' as the bullet point.",
            "parameters": {
                "content": null,
                "symbol": "-"
            }
        }
    ]
}
\end{lstlisting}

\textit{\#\# Example 2}\\
Input:
\begin{lstlisting}[language=json]
{
    "prompt": "Please output a JSON object, which must contain the keys 'summary' and 'key_actions'. The value for the 'key_actions' key should list all key actions of the main character.\n The value for the 'summary' key should be a video summary of no more than 30 words. The value for the 'key_actions' key should be an unordered list using '-' as the bullet point.",
    "constraints_used":[
        "json_object",
        "omni_summary",
        "visual_events_actions",
        "length",
        "unordered_list"
    ]
}
\end{lstlisting}
Expected Output:
\begin{lstlisting}[language=json]
{
    "format_check": [
        {
            "check_id": "format-001",
            "constraint_id": "json_object",
            "check_description": "Output a JSON object, which must contain the keys 'summary' and 'key_actions'.",
            "parameters": {
                "content": null,
                "schema": {
                    "type": "object",
                    "properties": {
                        "summary": { "type": "string" },
                        "key_actions": { "type": "string" }
                    },
                    "required": ["summary", "key_actions"]
                }
            }
        },
        {
            "check_id": "format-002",
            "constraint_id": "length",
            "check_description": "The value for the 'summary' key should be a video summary of no more than 30 words.",
            "parameters": {
                "content": null,
                "unit": "word",
                "max_len": 30
            }
        },
        {
            "check_id": "format-003",
            "constraint_id": "unordered_list",
            "check_description": "The value for the 'key_actions' key should be an unordered list using '-' as the bullet point.",
            "parameters": {
                "content": null,
                "symbol": "-"
            }
        }
    ]
}
\end{lstlisting}

Please strictly adhere to the Checklist construction philosophy and complete the task based on the input.
\end{promptbox}

\begin{promptbox}{The Prompt for Content Checklist Generation}

\textbf{\# Role and Goal}\\
You are an expert in evaluation questionnaire design, proficient in evaluation methodologies. Based on the original video and the user-provided prompt and format\_check, you need to generate a detailed JSON object for an evaluation checklist that fully adheres to all the rules below. This checklist will include a series of discriminative or comprehension questions to assess a caption generated by the omni model under test, based on the requirements of the prompt.

\textbf{\# Core Knowledge Base}\\
You must strictly follow the knowledge base named ‘Constraint System’. This knowledge base defines the technical details for all constraint\_ids.

\textbf{\# Input Information}\\
You will receive two inputs:
\begin{enumerate}[leftmargin=15pt, nosep]
    \item Original Video: The visual and auditory basis for fact-checking.
    \item A JSON object, which includes:
    \begin{itemize}[nosep]
        \item prompt: A string containing the constraint instructions for the model being tested.
        \item constraints\_used: An array of strings, which lists the unique IDs of all atomic constraints used to generate this prompt.
        \item format\_check: A list of constraint items from the prompt that have already been verified by rules. You do not need to check these items again. (The format\_check covers all format, structure, and keyword issues that can be automatically judged by code. Your task is to focus on aspects that require semantic understanding and factual judgment in conjunction with the video content.)
    \end{itemize}
\end{enumerate}

\textbf{\# Core Task: Generate Checklist}\\
You must strictly follow the process to generate the content for content\_check.
\begin{enumerate}[leftmargin=15pt, nosep]
    \item Content Decomposition: First, break down all content and semantic requirements from the prompt into multiple independent check\_content. Each check\_content should be a concise summary of a specific task, for example: "Generate a video summary" or "List all key actions of the protagonist." And a single check\_content can only check one constraint item, and cannot combine multiple independent constraint items. Then iterate through the input constraints\_used list. Ensure that check\_contents can cover all tasks from the prompt. All content constraints in the constraints\_used must be covered. If anything is missing, please add it.
    \item For each decomposed check\_content, you must strictly follow the decision-making process below to determine which check\_items to generate:
    \begin{enumerate}[leftmargin=15pt, label=\arabic*., nosep]
        \item Is an attempt check needed? (The attempt type is a yes/no question that only cares if the model tried to execute the instruction, not whether the content is correct.)
        \begin{itemize}[nosep]
            \item Condition: Check if the core intent of the constraint has been fully covered in format\_check.
            \item No, an attempt check is not needed: If it is covered (e.g., the prompt requires a JSON object, and its attribute correctness and structural validity are already checked by json\_object in format\_check), then do not generate an attempt check item.
            \item Yes, an attempt check is needed: If it is not covered (e.g., the prompt asks to "generate a summary," which is a non-format requirement) (e.g., the prompt asks to "generate a summary no more than 50 words," which has a format requirement length, but the core intent of the constraint is not covered), then you must generate an attempt check item. The question should focus on whether the model attempted to execute the instruction, for example: "Does the content appear to be a video summary?"
        \end{itemize}
        \item Is a correctness check needed? (The correctness type is a multiple-choice question aimed at examining how accurately the model describes video facts according to the prompt's requirements.)
        \begin{itemize}[nosep]
            \item Condition: Check if the correctness of the constraint can be objectively judged by the attempt check.
            \item No, a correctness check is not needed: If it can be judged (e.g., the prompt requires "do not mention color") by the attempt check, then do not generate a correctness check item.
            \item Yes, a correctness check is needed: If it requires combining specific facts from the video, generate fine-grained questions.
        \end{itemize}
        \item Is a timestamp check needed?
        \begin{itemize}[nosep]
            \item Condition: If the constraint\_id is visual\_temporal\_grounding, audio\_temporal\_grounding, or omni\_temporal\_grounding.
            \item Action: You must generate a check item with check\_type as "timestamp".
            \item Requirement: Do NOT generate options.
            \item Format Consistency: You must strictly parse the specific requirements of the prompt regarding time.
            \begin{itemize}[nosep]
                \item If the prompt asks for a "timestamp", "time point", "when ... starts", or "when ... ends", your correct\_answer must be a single time point (e.g., "00:15"). Do NOT provide a time range.
                \item If the prompt asks for a "time period", "duration", "interval", "start and end", or "timestamp segment", your correct\_answer must be a time range (e.g., "00:10 - 00:20").
                \item The format of correct\_answer must strictly align with the prompt's request.
            \end{itemize}
        \end{itemize}
    \end{enumerate}
    \item question Design Principles
    \begin{itemize}[leftmargin=15pt, nosep]
        \item Atomicity Principle:  Each question can only test a single, minimal, independent fact.
        \begin{itemize}[nosep]
            \item Correct Example: "Who is the protagonist in the video?"
            \item Incorrect Example: "Who is the protagonist, where are they, and what are they doing?" (This question contains multiple scoring points).
        \end{itemize}
        \item Distractor Hardening Principle (CRITICAL): When designing the wrong options for correctness checks, DO NOT use obvious fake facts. The distractors MUST be "Plausible Traps".
        \begin{itemize}[nosep]
            \item Tactic 1 (Temporal Trap): Use an event that actually happened in the video, but at the WRONG time.
            \item Tactic 2 (Entity Swap): Use an action that actually happened, but attribute it to the WRONG character or object.
            \item Tactic 3 (Modality Confusion): Provide a visually correct fact as the answer to an audio-based question.
        \end{itemize}
        \item The "None of the Above" Mandate: For at least 30\% of your generated Correctness questions, the 'correct\_answer' MUST be "None of the above" or "Cannot be determined". To do this, make all specific options (A, B, C) contain subtle factual errors based on the video.
        \item Answer Diversity Principle: You must randomize the position of the correct option for multiple-choice questions. Ensure a balanced distribution of 'A', 'B', 'C', and 'D' as the `correct\_answer` across the entire checklist to prevent positional bias. Do not default to placing the correct answer in option A.
        \item Granularity Design: The content of the questions should be based on the requirements of the generated prompt. Check the following indicators of the description:
        \begin{itemize}[nosep]
            \item Correctness: The content of the description should be accurately present in the video and meet the prompt’s descriptive requirements.
            \item Completeness: When the prompt requires a complete set, design questions to confirm whether all necessary elements are included. For example: "Which of the following options fully lists all the actions?"
            \item Exclusivity: When the prompt requires a focused set, design reverse questions to check whether unnecessary elements are excluded. The specific steps are as follows:
            \begin{enumerate}[label=\arabic*., nosep]
                \item Create distractors: Based on the video content, create several facts that exist or are reasonable but do not meet the prompt’s requirements as distractor options.
                \item Design the question: Your question should be something like: "Which of the following objects/actions (depending on the check content) is mentioned?"
                \item Design the answer: The correct answer should be "None of the above are mentioned."
            \end{enumerate}
            \item Note that the above checks also apply to granularity control (for example, for summaries, check whether unnecessary details are included, and for detailed descriptions, check whether important details are omitted).
        \end{itemize}
    \end{itemize}
    \item ID and Sorting
    \begin{itemize}[leftmargin=15pt, nosep]
        \item check\_id: Sequentially and uniquely increment the ID starting from "content-001" across all content\_checks.
        \item constraint\_id: The corresponding original content constraint ID.
        \item Internal Sorting: Within any single check\_content, if both attempt and correctness check items exist, the attempt item must precede the correctness item(s).
    \end{itemize}
\end{enumerate}

\textbf{\# Output Specification}\\
You must return a single valid JSON object that is the complete structure of the Checklist, without any additional explanatory text.
\begin{lstlisting}[language=json]
{
    "content_check":[
        {
        "check_content": "string", // The check content, extracted from the Prompt
        "constraint_id": "string", // The corresponding original content constraint ID, e.g., "visual_entities_attributes"
        "check_items":[
            {
                "check_id": "string", // Unique ID for the content check, e.g., "content-001"
                "check_type": "attempt|correctness|timestamp", // Intention/Accuracy/Timestamp check
                "question" : "string",  // The check question
                "options": [    // A list of options for 'correctness', or ['yes', 'no'] for 'attempt'. For 'timestamp' type, do NOT include this field.
                    "A. Option text 1",
                    "B. Option text 2",
                    "C. Option text 3",
                    "D. Option text 4"
                ],
                "correct_answer": "string"   // The correct answer, A, B, C or D, or yes or no, or a timestamp/interval
            }
            ]
        }
    ]
}
\end{lstlisting}

\textbf{\# Specific Example}\\
The following are complete examples including input and expected output for reference.

\textit{\#\# Example 1}\\
Input:\\
Assumed Video Content: A man with a beard, wearing a blue apron, is making coffee in a kitchen and humming a little tune. The video shows close-ups of the following steps: he first pours coffee beans into an electric grinder. Then he uses an espresso machine to make a shot of espresso, pouring it into a white mug. Finally, he steams milk with a steam wand and pours it into the coffee, creating a simple heart-shaped latte art. Throughout the process, the man has a focused and satisfied expression.
\begin{lstlisting}[language=json]
{
  "prompt": "Please generate a brief summary for this video, no more than 50 words. Then, in an unordered list starting with '-', list all the main tools used by the protagonist in the video. Ensure the output does not contain the word 'beverage'. Finally, state the exact time when the man pours the espresso into the mug.",
  "constraints_used": [
      "length",
      "unordered_list",
      "keyword",
      "omni_summary",
      "visual_entities_attributes",
      "visual_temporal_grounding"
      ],
  "format_check": [
        {
            "check_id": "format-001",
            "constraint_id": "length",
            "check_description": "Please generate a brief summary for this video, no more than 50 words.",
            "parameters": {
                "content": null,
                "unit": "word",
                "max_len": 50
            }
        },
        {
            "check_id": "format-002",
            "constraint_id": "unordered_list",
            "check_description": "In an unordered list starting with '-', list all the main tools used by the protagonist in the video.",
            "parameters": {
                "content": null,
                "symbol": "-"
            }
        },
        {
            "check_id": "format-003",
            "constraint_id": "keyword",
            "check_description": "Does not contain the word 'beverage'.",
            "parameters": {
                "content": null,
                "keyword": "beverage",
                "keyword_type": "exclude"
            }
        }
  ]
}
\end{lstlisting}
Expected Output:
\begin{lstlisting}[language=json]
{
    "content_check": [
        {
            "check_content": "Generate a video summary",
            "constraint_id": "omni_summary",
            "check_items": [
                {
                    "check_id": "content-001",
                    "check_type": "attempt",
                    "question": "Does the description look like a video summary?",
                    "options": [
                        "yes",
                        "no"
                    ],
                    "correct_answer": "yes"
                },
                {
                    "check_id": "content-002",
                    "check_type": "correctness",
                    "question": "Who is the protagonist of the video?",
                    "options": [
                        "A. A mug",
                        "B. Coffee",
                        "C. A man",
                        "D. Cannot be determined"
                    ],
                    "correct_answer": "C"
                },
                {
                    "check_id": "content-003",
                    "check_type": "correctness",
                    "question": "Which of the following descriptions of the protagonist's is correct?",
                    "options": [
                        "A. He is making noise",
                        "B. He is wearing a blue T-shirt",
                        "C. He is laughing",
                        "D. None of the above"
                    ],
                    "correct_answer": "D"
                }
            ]
        },
        {
            "check_content": "In an unordered list starting with '-', list all the main tools used by the protagonist",
            "constraint_id": "visual_entities_attributes",
            "check_items": [
                {
                    "check_id": "content-004",
                    "check_type": "attempt",
                    "question": "Are the items listed in the list tools?",
                    "options": [
                        "yes",
                        "no"
                    ],
                    "correct_answer": "yes"
                },
                {
                    "check_id": "content-005",
                    "check_type": "correctness",
                    "question": "Based on the video description, which of the following options most completely lists all the main tools used by the protagonist?",
                    "options": [
                        "A. Espresso machine, mug",
                        "B. Grinder, espresso machine, mug",
                        "C. Grinder, French press, mug",
                        "D. No tools are mentioned in the description"
                    ],
                    "correct_answer": "B"
                },
                {
                    "check_id": "content-006",
                    "check_type": "correctness",
                    "question": "According to the video description, which of the following tools was mentioned?",
                    "options": [
                        "A. Coffee canister",
                        "B. Milk carton",
                        "C. Electric kettle",
                        "D. None of the above were mentioned"
                    ],
                    "correct_answer": "D"
                }
            ]
        },
        {
            "check_content": "State the exact time when the man pours the espresso into the mug",
            "constraint_id": "visual_temporal_grounding",
            "check_items": [
                {
                    "check_id": "content-007",
                    "check_type": "timestamp",
                    "question": "According to the video, when does the man pour the espresso into the mug?",
                    "correct_answer": "00:25"
                }
            ]
        }
    ]
}
\end{lstlisting}

\textit{\#\# Example 2}\\
Input:\\
Assumed Video Content: A man with a beard, wearing a blue apron, is making coffee in a kitchen. The video shows close-ups of the following steps: he first pours coffee beans into an electric grinder. Then he uses an espresso machine to make a shot of espresso, pouring it into a white mug. Finally, he steams milk with a steam wand and pours it into the coffee, creating a simple heart-shaped latte art. Throughout the process, the man has a focused and satisfied expression. The video has soft background music.
\begin{lstlisting}[language=json]
{
  "prompt": "Please generate a brief summary for this video, no more than 50 words. Ensure the output does not contain the word 'beverage'.\n In an unordered list starting with '-', list all the main tools used by the protagonist in the video if the video has background music.",
  "constraints_used": [
      "length",
      "unordered_list",
      "keyword",
      "omni_summary",
      "visual_entities_attributes"
      ],
  "format_check": [
        {
            "check_id": "format-001",
            "constraint_id": "length",
            "check_description": "Please generate a brief summary for this video, no more than 50 words.",
            "parameters": {
                "content": null,
                "unit": "word",
                "max_len": 50
            }
        },
        {
            "check_id": "format-002",
            "constraint_id": "unordered_list",
            "check_description": "In an unordered list starting with '-', list all the main tools used by the protagonist in the video.",
            "parameters": {
                "content": null,
                "symbol": "-"
            }
        },
        {
            "check_id": "format-003",
            "constraint_id": "keyword",
            "check_description": "Does not contain the word 'beverage'.",
            "parameters": {
                "content": null,
                "keyword": "beverage",
                "keyword_type": "exclude"
            }
        }
  ]
}
\end{lstlisting}
Expected Output:
\begin{lstlisting}[language=json]
{
    "content_check": [
        {
            "check_content": "Generate a video summary",
            "constraint_id": "omni_summary",
            "check_items": [
                {
                    "check_id": "content-001",
                    "check_type": "attempt",
                    "question": "Does the description look like a video summary?",
                    "options": [
                        "yes",
                        "no"
                    ],
                    "correct_answer": "yes"
                },
                {
                    "check_id": "content-002",
                    "check_type": "correctness",
                    "question": "Who is the protagonist of the video?",
                    "options": [
                        "A. A mug",
                        "B. Coffee",
                        "C. A man",
                        "D. Cannot be determined"
                    ],
                    "correct_answer": "C"
                },
                {
                    "check_id": "content-003",
                    "check_type": "correctness",
                    "question": "Which of the following descriptions of the protagonist's attire is correct?",
                    "options": [
                        "A. He is making noise",
                        "B. He is wearing a blue T-shirt",
                        "C. He is laughing",
                        "D. None of the above"
                    ],
                    "correct_answer": "D"
                }
            ]
        },
        {
            "check_content": "In an unordered list starting with '-', list all the main tools used by the protagonist",
            "constraint_id": "visual_entities_attributes",
            "check_items": [
                {
                    "check_id": "content-004",
                    "check_type": "attempt",
                    "question": "Are the items listed in the list tools?",
                    "options": [
                        "yes",
                        "no"
                    ],
                    "correct_answer": "yes"
                },
                {
                    "check_id": "content-005",
                    "check_type": "correctness",
                    "question": "Based on the video description, which of the following options most completely lists all the main tools used by the protagonist?",
                    "options": [
                        "A. Espresso machine, mug",
                        "B. Grinder, espresso machine, mug",
                        "C. Grinder, French press, mug",
                        "D. No tools are mentioned in the description"
                    ],
                    "correct_answer": "B"
                },
                {
                    "check_id": "content-006",
                    "check_type": "correctness",
                    "question": "According to the video description, which of the following tools was mentioned?",
                    "options": [
                        "A. Coffee canister",
                        "B. Milk carton",
                        "C. Electric kettle",
                        "D. None of the above were mentioned"
                    ],
                    "correct_answer": "D"
                }
            ]
        }
    ]
}
\end{lstlisting}

Please strictly adhere to the Checklist construction philosophy and complete the task according to the input.
\end{promptbox}

\begin{promptbox}{The Prompt for Prompts Generation}

\textbf{\# ROLE and GOAL}\\
You are a creative and meticulous omni-modal AI evaluation expert. Your core task is to generate 12 diverse and challenging video description task packages based on a given video and a constraint knowledge base, to evaluate the instruction-following capabilities of large omni models.

\textbf{\# Knowledge Base}\\
You must strictly adhere to the constraint system knowledge base named "Constraint System." When constructing tasks, you must:
\begin{itemize}[leftmargin=15pt, nosep]
    \item Atomic Constraints: Only use the atomic constraint items defined in the knowledge base and combine them according to the specified constraint relationships.
    \item Record IDs: You must record all used constraint items by their unique IDs.
    \item Design for Quantifiable Evaluation: Understand the check items corresponding to each constraint to design description prompts that can be quantitatively evaluated.
\end{itemize}

\textbf{\# TASK}\\
The given video is sampled at 5fps. For the input video, you need to:
\begin{enumerate}[leftmargin=15pt, label=\arabic*., nosep]
    \item Deep Omni-Modal Video Analysis: Conduct a holistic analysis by natively integrating the video's pure visual, pure auditory, and cross-modal streams. This forms the basis for generating high-quality tasks.
    \item Create Diverse Instructions: Based on the instruction preferences in the four general domains—For Understanding, For Generation, For Retrieval, and For Communication—create three distinct instructions for each (totaling twelve instructions). Ensure the instructions are creative and challenging, align as closely as possible with the instruction preferences in the knowledge base, and combine as many constraint items from the knowledge base as possible. Also, ensure the instructions primarily use the word "describe" or similar phrasing, rather than being questions or straying from the domain of video description tasks. Even if the description involves inference, it must be based on the actual video content.
\end{enumerate}

\textbf{\# Special Requirements}\\
When completing the task, please pay special attention to the following:
\begin{enumerate}[leftmargin=15pt, label=\arabic*., nosep]
    \item Field Distribution: Create 3 distinct prompts for each of the 4 general application scenarios: For Understanding, For Generation, For Retrieval, and For Communication, based on the prompt preferences for each.
    \item Prompt Design Principles:
    \begin{itemize}[nosep]
        \item Focus on "Description": Prompts should focus on "describing" the video content and avoid being framed as questions or reasoning tasks that go beyond the video's content.
        \item Specific and Quantifiable: Constraints must be explicit and quantifiable (e.g., "list 3 objects," "no more than 50 words"), avoiding vague, qualitative descriptions (e.g., "in a humorous style").
        \item Align with the actual video content: The proposed prompts should be meaningful and accurately reflect the real content of the video. Encourage the creation of description instructions specifically designed for unique content in the video. Avoid prompts that are incorrect or unrealistic (it is necessary to pre-validate whether a response would be reasonable).
        \item Omni-Centric: To ensure the benchmark focuses on deep audio-visual integration, the distribution of content constraints should be Omni-centric. The instruction set must prioritize the use of omni\_... constraints as the core task drivers. Pure visual\_... and audio\_... items should serve as secondary or supporting elements. The frequency of omni\_... prefixes should ideally be the highest among the three categories, reflecting a primary emphasis on the model's ability to process and synchronize both modalities simultaneously.
        \item The number of visual constraints should not exceed the number of omni constraints.
        \item Do Not Plagiarize Examples: You must not copy or closely imitate any examples from the knowledge base.
        \item Constraint Type Balance: The number of format constraints should not exceed the number of content constraints.
        \item Try to utilize the constraints from the knowledge base as diversely as possible.
    \end{itemize}
\end{enumerate}

\textbf{\# Difficulty of Generated Prompts}
\begin{itemize}[leftmargin=15pt, nosep]
    \item For each field, the three prompts should belong to different progressive difficulty tiers:
    \begin{itemize}[nosep]
        \item Level 1 (Normal Difficulty): 4-5 constraints in total. MUST include 1-2 omni constraints + basic visual, audio and format constraints.
        \item Level 2 (High Difficulty): 6-7 constraints in total. MUST include 2-3 Omni constraints + advanced visual, audio and format constraints.
        \item Level 3 (Extreme Difficulty): 8+ constraints in total. Intricate integration of the hardest constraints across content and format domains.
    \end{itemize}
    \item Overall, the generated prompts should be sufficiently challenging for current omni-modal large language models, deeply examining the model's capability for controlled description. Difficulty can be increased in the following aspects:
\end{itemize}

\begin{enumerate}[leftmargin=15pt, label=\arabic*., nosep]
    \item Structural Control Complexity
    \begin{itemize}[nosep]
        \item Deep and Complex Nesting of Multiple Structural Constraints: For instance, many layers of JSON-type nesting, or tables containing code blocks or lists.
        \item Mixture and Dependency of Different Types or Requirements of Structural Constraints: For example, the simultaneous presence of multiple different length limitations and various structural requirements.
        \item Format-Constraint Conflict \& Dependency: Introduce constraints where the format depends on the modality. For example: "Use a Markdown table for visual events, but use a JSON code block for audio metadata within the same response," or requiring different timestamp formats for different event types.
    \end{itemize}
    \item Omni-modal Perception Complexity
    \begin{itemize}[nosep]
        \item Cross-Timeline Information Integration and Comparison: Focus more on the perception of dynamic information unique to videos along a timeline, such as requiring the model to describe a complete dynamic process or to compare information from different points in time.
        \item Capture of Subtle/Occluded/Non-Focal Information: For instance, requiring the model to focus only on a minor character, background elements or "Off-screen sounds" (Voiceovers, BGM) while ignoring the main character's actions; requiring the model to label content that cannot be accurately identified as "uncertain."
        \item Complex Spatial Relationship Understanding: For example, requiring the model to focus only on entities in the upper-left portion of the video and to describe the constantly changing relative positions between entities.
        \item Fine-Grained Temporal Grounding \& Desynchronization: Focus on identifying the precise start/end points where audio and visuals do not match (e.g., J-cuts, L-cuts, or lip-sync errors).
    \end{itemize}
    \item Multimodal Reasoning Complexity
    \begin{itemize}[nosep]
        \item Irony, Subtext, and Emotional Contradiction: Require the model to "describe" the tension when the audio (e.g., upbeat music) contradicts the visual (e.g., a tragic scene). Instead of asking "Is it ironic?", ask the model to "Describe the emotional dissonance between the visual color palette and the musical key."
        \item Causal and Intentional Inference: The instruction does not directly ask "why," but rather requires the "description" of this causal or intentional relationship. For example, requiring the description of a person's behavior in a video and, based on their expression, voices and actions, inferring the preceding events that led to that mood.
        \item Immersive Multi-Sensory First-Person Perspective: Require a narrative that integrates "what I see" and "what I hear" into a cohesive internal monologue.
        \item Counterfactual Reasoning: "Describe the scene as it is, then describe how the character's action would change if the whispering audio at [00:05] were replaced by a loud shout."
    \end{itemize}
\end{enumerate}

\textbf{\# Output Specification}\\
You must return the answer as a single valid JSON array containing exactly 12 objects, without any explanatory text. Each object in the array must conform to the following structure:
\begin{lstlisting}[language=json]
{
  "field": "string", // Must be one of the application domains from the knowledge base: For Understanding, For Generation, For Retrieval, For Communication
  "prompt_id": "string", // The unique ID of the prompt, starting with "01", followed by "02", "03", etc.
  "generated_prompt": "string", // The instruction you generated
  "constraints_used": [
    "string" // An array of unique ID strings for all constraints from the knowledge base used in this task.
  ]
}
\end{lstlisting}

\textbf{\# Example 1}
\begin{lstlisting}[language=json]
{
  "field": "For Communication",
  "prompt_id": "01",
  "generated_prompt": "Please describe the moment the blender overflows and the chef's reaction in plain text. You must transcribe the exact phrase the chef shouts and mention the color of their apron. Keep your description strictly under 40 words.",
  "constraints_used": [
    "omni_events_actions",
    "visual_entities_attributes",
    "audio_specific",
    "plain_text",
    "length"
  ]
}
\end{lstlisting}

\textbf{\# Example 2}
\begin{lstlisting}[language=json]
{
  "field": "For Understanding",
  "prompt_id": "02",
  "generated_prompt": "Contrast the background music with the visual chaos of the blender overflowing, and describe how the contrast is formed. Infer the chef's stress level based on this sudden change. You must also provide the exact timestamp when the spill begins. \n Please present this as an unordered list starting with '-', describe the specific genre of the music, use Markdown bolding for the emotion words, and completely exclude any mention of the vegetables.",
  "constraints_used": [
    "omni_contrast",
    "omni_inference",
    "visual_temporal_grounding",
    "audio_entities_attributes",
    "unordered_list",
    "markdown",
    "visual_exclude"
  ]
}
\end{lstlisting}

\textbf{\# Example 3}
\begin{lstlisting}[language=json]
{
  "field": "For Generation",
  "prompt_id": "03",
  "generated_prompt": "Adopt the first-person perspective of the blender: First, describe the visual vibration and the grinding sound you experience; then, describe the sudden camera zoom; finally, describe the exact moment you overflowed, syncing it with the chef's shout. If the chef is scared by the overflowing blender, generate a JSON object describing the audio observations. The object must have an 'shout_timestamp' key (string in the format [MM:SS] indicating the exact timestamp of the chef's shout), an 'audio_shift_during_shout' key (string describing how the audio track shifts between music and speech during the shout), and a 'room_sounds' key (an array of exactly 3 strings listing distinct sounds heard in the room).",
  "constraints_used": [
    "omni_perspective",
    "visual_cinematic_elements",
    "omni_temporal_grounding",
    "audio_temporal_grounding",
    "timestamp_format",
    "audio_production_structure",
    "json_object",
    "count"
  ]
}
\end{lstlisting}

\end{promptbox}

\subsection{Construction of Prompts for The Training Set}
\label{sec:train_set_prompt}
The prompt for training set generation consists of the prompt shown below, with the constraint system table in Section~\ref{sec:system} serving as the actual content of the Core Knowledge Base.

\begin{promptbox}{The Prompt for Negative Constraint Filter}

\textbf{\# Role and Goal}\\
You are a highly logical and rigorous data quality controller. Your task is to analyze a highly detailed omni-modal video caption and determine which instruction constraints CANNOT be applied to this video. This is to prevent the generation of hallucinated instructions that the video content cannot support.

\textbf{\# Core Knowledge Base}\\
You must strictly follow the knowledge base named 'Constraint System'. This knowledge base defines the technical details and prerequisite conditions for all \texttt{constraint\_id}s.

\textbf{\# Input Information}\\
You will receive a JSON object containing:\\
\texttt{caption}: A detailed textual proxy description of a video, encompassing both visual and auditory streams, as well as their temporal alignments.

\textbf{\# Core Task: Constraint Filtering}\\
You need to evaluate all \texttt{constraint\_id}s from the 'Constraint System' against the provided \texttt{caption}. If the prerequisite elements required by a constraint are ABSENT from the caption, you must blacklist that constraint.
\begin{itemize}[leftmargin=15pt, nosep]
    \item \textbf{Example 1}: If the caption does not mention any audio-visual desynchronization, lip-sync issues, or specific sound-action misalignment, you MUST blacklist \texttt{omni\_temporal\_grounding}.
    \item \textbf{Example 2}: If the video only contains a single static shot without any camera movement or cuts, you MUST blacklist \texttt{visual\_cinematic\_elements} and \texttt{omni\_editing\_transitions}.
\end{itemize}

\textbf{\# Output Specification}\\
Return a single valid JSON array containing the blacklisted \texttt{constraint\_id}s and a brief reason for each. Do not output any other text.
\begin{lstlisting}[language=json]
{
    "blacklisted_constraints":[
        {
            "constraint_id": "string",
            "reason": "string" // e.g., "No audio-visual desynchronization is described in the caption."
        }
    ]
}
\end{lstlisting}
\end{promptbox}

\begin{promptbox}{The Prompt for Randomized Instruction Synthesis}

\textbf{\# Role and Goal}\\
You are a creative and versatile instruction data generator. Your core task is: Based on the original video caption, and referencing a comprehensive knowledge base of a constraint framework, construct rigorous prompts that align with the caption.

\textbf{\# Core Knowledge Base}\\
You must strictly follow the knowledge base named 'Constraint System'. This defines all available \texttt{constraint\_id}s.

\textbf{\# Input Information}\\
You will receive a JSON object containing:
\begin{itemize}[leftmargin=15pt, nosep]
    \item \texttt{caption}: A detailed textual proxy description of the video content.
    \item \texttt{blacklisted\_constraints}: A list of \texttt{constraint\_id}s that CANNOT be fulfilled by this specific video.
\end{itemize}

\textbf{\# Core Task: Stochastic Instruction Synthesis}\\
Based on the \texttt{caption}, synthesize 4 unique user instructions by following these steps:
\begin{enumerate}[leftmargin=15pt, nosep]
    \item \textbf{Randomized Constraint Sampling}: For each instruction, independently and randomly sample a set of 3 to 8 \texttt{constraint\_id}s from the 'Constraint System'. You MUST exclude any IDs present in the \texttt{blacklisted\_constraints}. 
    \item \textbf{Natural Task Integration}: Weave the sampled constraints into a single, cohesive, and natural-sounding user request.
    \item \textbf{Ground Truth Alignment}: Every generated instruction must be strictly answerable based on the information provided in the \texttt{caption}. Do not introduce external requirements or hallucinated details.
\end{enumerate}

\textbf{\# Output Specification}\\
Return a single valid JSON object containing the synthesized instructions. Do not output any other text.
\begin{lstlisting}[language=json]
{
    "synthesized_data":[
        {
            "instruction": "string", // The synthesized multi-constraint user prompt
            "sampled_constraints":["string"] // The list of constraint_ids that were randomly selected and integrated
        }
    ]
}
\end{lstlisting}
\end{promptbox}

\begin{promptbox}{The Prompt for Instruction Decomposition}

\textbf{\# Role and Goal}\\
You are a highly logical task decomposer. Your goal is to break down a complex, multi-constraint user instruction into manageable components based on a provided \texttt{Constraint System}. You need to separate format constraints from content constraints.

\textbf{\# Core Knowledge Base}\\
You must use the knowledge base named 'Constraint System' to distinguish between Format and Content constraints.

\textbf{\# Input Information}\\
You will receive a JSON object containing:
\begin{itemize}[leftmargin=15pt, nosep]
    \item \texttt{instruction}: The complex user prompt.
    \item \texttt{sampled\_constraints}: The list of \texttt{constraint\_id}s used in this instruction.
\end{itemize}

\textbf{\# Core Task: Decoupling and Chunking}\\
Analyze the \texttt{instruction} and perform these steps:
\begin{enumerate}[leftmargin=15pt, nosep]
    \item \textbf{Constraint Categorization}: Referencing the \texttt{Constraint System}, identify which IDs in \texttt{sampled\_constraints} are \textbf{Format Constraints} and which are \textbf{Content Constraints}.
    \item \textbf{Format Extraction}: List all identified \textbf{Format Constraints}. These will be handled in the final reformatting stage.
    \item \textbf{Content Chunking}: Group the \textbf{Content Constraints} into sub-tasks. Each sub-task should contain 2 to 3 constraints to maintain high accuracy.
    \item \textbf{Sub-instruction Drafting}: For each content chunk, write a clear \texttt{sub\_instruction}. This sub-instruction must ask for the raw factual information from the video (e.g., "Describe the colors of the cars and the specific actions of the driver") while ignoring all formatting rules (like JSON, tables, or word counts).
\end{enumerate}

\textbf{\# Output Specification}\\
Return a single valid JSON object. Do not output any other text.
\begin{lstlisting}[language=json]
{
    "format_constraints": ["string"], 
    "content_subtasks": [
        {
            "target_constraints": ["string"],
            "sub_instruction": "string"
        }
    ]
}
\end{lstlisting}

\vspace{5pt}
Next, carefully read the actual input below and perform the processing that conforms to the conventions and is reasonable:
\end{promptbox}

\begin{promptbox}{The Prompt for Sub-task Execution}

\textbf{\# Role and Goal}\\
You are a highly accurate Video Content Expert. Your task is to answer a specific sub-instruction based entirely on a highly detailed text proxy of a video (the caption). Your goal is to retrieve the exact facts, events, and timestamps requested.

\textbf{\# Input Information}\\
You will receive a JSON object containing:
\begin{itemize}[leftmargin=15pt, nosep]
    \item \texttt{caption}: A detailed textual proxy description of the video content, including visual events, audio tracks, and timestamps.
    \item \texttt{sub\_instruction}: A specific, content-focused question or directive.
\end{itemize}

\textbf{\# Core Task: Factual Retrieval}\\
\begin{enumerate}[leftmargin=15pt, nosep]
    \item Analyze the \texttt{sub\_instruction} and locate the relevant information within the \texttt{caption}.
    \item Generate a highly accurate, factual response. If the instruction asks for timestamps, extract the exact timestamps from the caption. If it asks for cross-modal reasoning (e.g., matching a sound to an action), perform the reasoning strictly based on the text.
\end{enumerate}

\textbf{\# Output Specification}\\
Return a single valid JSON object. Do not output any additional explanatory text.
\begin{lstlisting}[language=json]
{
    "intermediate_response": "string" // Your factually accurate, unformatted answer to the sub_instruction
}
\end{lstlisting}
\end{promptbox}

\begin{promptbox}{The Prompt for Aggregation and Reformatting}

\textbf{\# Role and Goal}\\
You are an expert formatting and synthesis engine. Your task is to aggregate multiple pieces of raw, factual information (intermediate responses) and reformat them to perfectly answer a complex original instruction. You must strictly enforce all format constraints.

\textbf{\# Input Information}\\
You will receive a JSON object containing:
\begin{itemize}[leftmargin=15pt, nosep]
    \item \texttt{original\_instruction}: The complete, highly constrained user prompt that needs to be answered.
    \item \texttt{format\_constraints}: The list of format \texttt{constraint\_id}s (e.g., \texttt{json\_object}, \texttt{length}, \texttt{unordered\_list}) that MUST be strictly applied to the final output.
    \item \texttt{intermediate\_responses}: An array of strings containing all the necessary factual information, already verified and extracted from the video.
\end{itemize}

\textbf{\# Core Task: Synthesis and Strict Formatting}\\
\begin{enumerate}[leftmargin=15pt, nosep]
    \item \textbf{Synthesize Content}: Combine all the facts, events, and timestamps provided in the \texttt{intermediate\_responses}. Do not invent or hallucinate any new video content. The intermediate responses contain the absolute truth.
    \item \textbf{Apply Formatting}: Restructure the combined information to strictly satisfy the \texttt{original\_instruction}. Pay obsessive attention to the \texttt{format\_constraints}.
    \begin{itemize}[nosep]
        \item If a specific JSON schema is required, ensure the keys and nesting are flawless.
        \item If a strict length is imposed, count your words/characters carefully.
        \item If specific symbols, prefixes, or markdown tables are requested, apply them exactly.
    \end{itemize}
    \item \textbf{Final Polish}: Ensure the final response seamlessly integrates the factual content into the required rigid structure without losing any semantic details requested by the original prompt.
\end{enumerate}

\textbf{\# Output Specification}\\
Return a single valid JSON object containing your final constructed string. Do not output any additional explanatory text.
\begin{lstlisting}[language=json]
{
    "final_response": "string" // The final, fully formatted, and factually correct response
}
\end{lstlisting}
\end{promptbox}

\subsection{Prompts for Evaluation on Existing Benchmarks}\label{bencheval}

To comprehensively evaluate our model's generalizable omni-modal perception capabilities on external benchmarks, we utilized specific prompts to guide the model in generating detailed descriptions or performing QA tasks. The specific prompts used for Omni-Cloze~\citep{ma2026omnicaptioner}, DailyOmni~\citep{zhou2025dailyomni}, and WorldSense~\citep{hong2026worldsense} are listed below.

\begin{promptbox}{Prompt for Detailed Video Description (Omni-Cloze)}
You are given a short video with both audio and visual content. Write a detailed and coherent paragraph that naturally integrates all modalities. Your description should include: (1) the primary scene and background setting; (2) key characters or objects and their actions or interactions; (3) significant audio cues such as voices, background music, sound effects, and their emotional tone; (4) any on-screen text (OCR) and its role in the video context; and (5) the overall theme or purpose of the video. Ensure the output is a fluent and objective paragraph, not a bullet-point list, and captures the video's content in a human-like, narrative style.
\end{promptbox}

\begin{promptbox}{Prompt for Question-to-Prompt Construction (DailyOmni \& WorldSense)}
You are an expert prompt engineer. I will give you a list of specific questions about a video (covering spatial relations, audio changes, temporal events, etc.).

Your task is to write a single, comprehensive and sophisticated prompt directed at a Vision-Language AI model. 
The prompt should instruct the AI to generate a detailed video caption that naturally incorporates all necessary visual and audio details required to answer these specific questions accurately.

\textbf{Constraints:}
\begin{enumerate}[leftmargin=15pt, nosep]
    \item DO NOT answer the questions yourself.
    \item DO NOT list the questions in your output.
    \item Your output must be ONLY the final generated prompt. No preamble or explanations.
\end{enumerate}

Here are the questions describing what needs to be captured:
\end{promptbox}

\begin{promptbox}{Prompt for Caption-to-QA Evaluation (DailyOmni \& WorldSense)}
You are an Omni-modal Video Understanding Expert. 
Your task is to answer a multiple-choice question about a video based SOLELY on the provided video caption.

Below is a detailed caption describing a video, followed by a question and four choices.

\textbf{Video Caption:}\\
\{caption\}

\textbf{Question:}\\
\{question\}

\textbf{Choices:}\\
\{choices\_str\}

Please select the best answer based on the caption provided.
If the answer cannot be determined with certainty, choose the most likely option.
Respond with ONLY the uppercase letter of the correct answer (A, B, C, or D). Do not output any other text or explanation.
\end{promptbox}

\begin{promptbox}{Prompt for Direct QA Evaluation (DailyOmni \& WorldSense)}
Please analyze the video and answer the following multiple-choice question. Output ONLY the letter (A, B, C, or D) of the correct answer. Do not provide any explanations, remarks, or extra text. If you are unsure, choose the most plausible option.

\textbf{Question:}\\
\{question\}
\end{promptbox}

\section{Construction of The Test Set}
\label{sec:test_construction}

To ensure the high quality and academic rigor of the OmniCap-IF benchmark, we designed a reproducible pipeline to transform raw videos into finalized, expert-verified samples. The construction process consists of three main stages:

\subsection{Video Collection and Filtration}
We curated an initial pool of over 1500 videos from diverse sources, including academic datasets (Ego4D~\citep{grauman2022ego4d}) and social media platforms. To ensure the benchmark's quality, we applied the following filters:
\begin{itemize}[leftmargin=*, topsep=0pt, itemsep=0pt, parsep=0pt]
    \item \textbf{Resolution:} Minimum resolution of 720p to ensure visual clarity.
    \item \textbf{Duration:} Focused on the 30--90 second range to provide sufficient semantic density for multi-constraint tasks.
    \item \textbf{Multi-modality:} Ensure that the videos are both rich in audio-visual content and consistently aligned across sound and imagery.
\end{itemize}
This resulted in a core set of 480 high-quality videos covering 10 categories such as Comedy \& Sketches, Lifestyle \& Vlogs, and Knowledge \& Tech.

\subsection{Automated Draft Generation}
For each selected video, we utilized the \textbf{Instruction Generator} (powered by Gemini-3.1-pro~\citep{Google2026Gemini3}) to produce paired instruction-checklist candidates. 
\begin{itemize}[leftmargin=*, topsep=0pt, itemsep=0pt, parsep=0pt]
    \item \textbf{Prompt Generation:} Constraints were sampled from our 50-type taxonomy and combined to create instructions of varying difficulty (Level 1 to Level 3).
    \item \textbf{Checklist Synthesis:} For each instruction, the generator simultaneously produced a \textbf{Format Checklist} (JSON schemas, length limits) and a \textbf{Content Checklist} (fact-based QA pairs).
\end{itemize}

\subsection{Human Refinement and Verification}
\label{sec:human_verification}
This stage is critical for ensuring the factual grounding and structural rigor of the OmniCap-IF benchmark. To achieve this, we employed a team of professionally trained annotators to conduct a three-step verification and refinement process:

\begin{enumerate}[leftmargin=*, topsep=0pt, itemsep=0pt, parsep=0pt]
    \item \textbf{Instruction-Video Alignment:} Annotators first watch the video to ensure that the requirements in the instructions are strictly factually grounded in the actual visual and auditory content. Any instructions requesting non-existent entities, actions, or sounds (i.e., hallucinations) are corrected or completely rewritten.
    
    \item \textbf{Constraint Taxonomy Compliance:} The instruction is then audited to confirm it strictly adheres to our defined taxonomy of 50 constraint types. We verify that no "out-of-scope" requirements are introduced and that the complexity levels (normal, high, extreme) are appropriately distributed across the instruction set.
    
    \item \textbf{Checklist Factual and Formal Validation:} Once the instruction is validated, annotators rigorously examine the generated \textbf{Checklist}. This includes:
    \begin{itemize}[nosep]
        \item \textbf{Formal Check:} Confirming that format parameters (e.g., JSON schemas, length units, keyword types) in the Format Check are correctly extracted and logically consistent with the instruction.
        \item \textbf{Content Check:} Manually verifying that every question in the Content Check has a unique, factually correct answer based on the video, and that the distractors are plausible but incorrect. For temporal grounding annotations, minor second-level rounding differences were tolerated, while disagreements larger than 1s were adjudicated by a senior supervisor.
    \end{itemize}

    \item \textbf{Consensus Protocol:} Through this rigorous process, 53.1\% of the samples were modified and 22.7\% were discarded or rewritten. Each final sample was confirmed only after reaching a unanimous agreement among three independent annotators, with any persistent disagreements adjudicated by a senior supervisor.
\end{enumerate}

\section{Evaluation Settings}\label{sec:eval}

We provide the detailed settings of our evaluated open-source models (Table~\ref{tab:eval_settings}). Most models are tested under default settings. Closed-source models are accessed via API calls, using the default configuration. The system prompts used for the models are detailed in Section~\ref{sec:system_prompt}.

\begin{table}[h]
  \caption{Evaluation metrics for locally deployed open-source models. The ``FPS'' column represents the frame sampling rate.}
  \label{tab:eval_settings}
  \renewcommand{\arraystretch}{1.1} 
  \centering
  \begin{tabular}{l c c c c}
    \toprule
    \textbf{Models} & \textbf{FPS} & \textbf{Temperature} & \textbf{Repetition Penalty} & \textbf{Max Token} \\
    \midrule
    Qwen3-Omni-30B-A3B-Thinking & 1.0 & 0.6 & 1.05 & 4096 \\
    Qwen3-Omni-30B-A3B-Instruct & 1.0 & 0.0 & 1.05 & 1536 \\
    MiniCPM-o-4.5-9B            & 1.0 & 0.0 & 1.05 & 1536 \\
    Qwen2.5-Omni-7B             & 1.0 & 0.0 & 1.05 & 1536 \\
    Qwen2.5-Omni-3B             & 1.0 & 0.0 & 1.05 & 1536 \\
    video-SALMONN-2-7B          & 1.0 & 0.0 & 1.05 & 1536 \\
    MiniCPM-o-2.6-8B            & 1.0 & 0.0 & 1.05 & 1536 \\
    HumanOmniV2-7B              & 1.0 & 0.0 & 1.05 & 1536 \\
    ASID-Captioner-7B           & 1.0 & 0.0 & 1.05 & 1536 \\
    ARC-Hunyuan-Video-7B        & 1.0 & 0.0 & 1.05 & 1536 \\
    \midrule
    OmniCaptioner-IF-7B (Ours)  & 1.0 & 0.0 & 1.05 & 1536 \\
    OmniCaptioner-IF-3B (Ours)  & 1.0 & 0.0 & 1.05 & 1536 \\
    \bottomrule
  \end{tabular}
\end{table}

\section{Training}
\label{sec:train}
\subsection{Training Configurations}
\label{sec:train_config}
Our models, OmniCaptioner-IF-7B and OmniCaptioner-IF-3B, were developed by fine-tuning the pre-trained Qwen2.5-Omni-7B and Qwen2.5-Omni-3B models, respectively. We employed Parameter-Efficient Fine-Tuning (PEFT) via Low-Rank Adaptation (LoRA) \citep{hu2022lora} applied to all linear layers. The LoRA rank was set to 16 with an alpha value of 32. 

The fine-tuning process was conducted for a total of 1 epoch on our curated OmniCap-IF-54K dataset. We utilized the AdamW optimizer with a peak learning rate of $2 \times 10^{-5}$ for the 7B variant and $3 \times 10^{-5}$ for the 3B variant. 

To accommodate hardware constraints and maximize throughput, training was performed on a single node equipped with 8 H200 GPUs. We configured a per-device batch size of 1 and utilized 2 gradient accumulation steps, resulting in an effective global batch size of 16. To enhance computational efficiency and reduce the memory footprint, we leveraged bfloat16 (bf16) mixed-precision training. For data preprocessing, input videos were sampled at a rate of 1 FPS. The maximum resolution was capped at 401,408 pixels for the 7B model and 200,704 pixels for the 3B model to balance perceptual granularity and memory usage.

\subsection{Convergence Analysis and Dataset Sufficiency}

To validate the sufficiency of our 54K instruction-tuning dataset and the 1-epoch training strategy for models at the 3B and 7B scales, we analyze both the training dynamics and the intermediate checkpoint performances. 

\begin{figure}[h]
  \centering
  \includegraphics[width=0.85\linewidth]{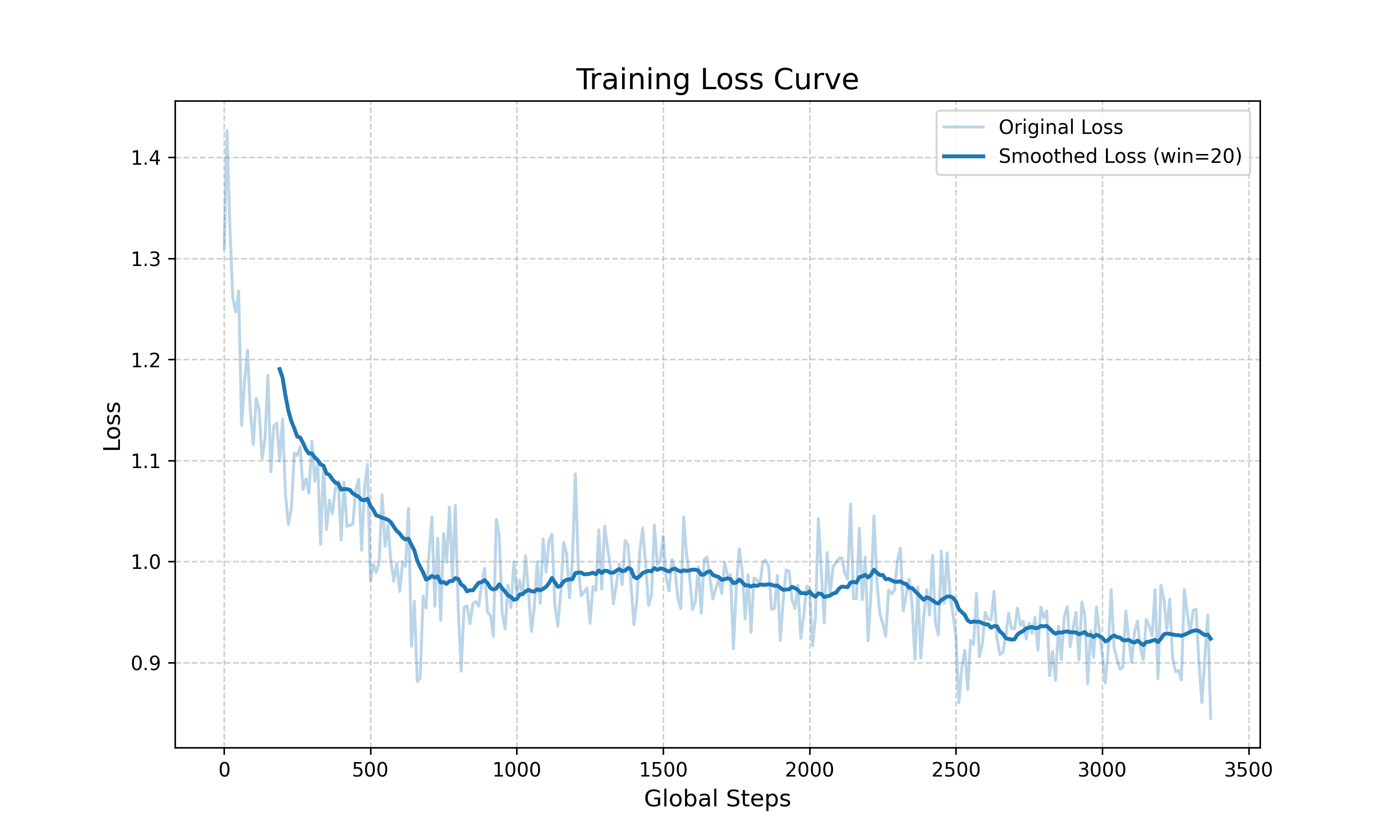} 
  \caption{The training loss curve of OmniCaptioner-IF-7B over 1 epoch.}
  \label{fig:loss_curve}
\end{figure}

First, as illustrated by the training loss curve in Figure~\ref{fig:loss_curve}, the model exhibits a rapid and healthy descent during the initial phase. After approximately 1,000 steps, the loss gradually flattens out, eventually stabilizing and plateauing around a value of 0.9 with minor fluctuations towards the end of the epoch. This trajectory indicates that the model has smoothly converged and effectively internalized the complex instruction-following patterns within a single epoch, without suffering from under-fitting.

Second, we conducted a data scaling ablation study by evaluating intermediate checkpoints—saved after training on 20K, 40K, and the full 54K samples—on the OmniCap-IF benchmark. As detailed in Table~\ref{tab:convergence}, merely fine-tuning on 20K samples yields a massive performance leap compared to the base model, Qwen2.5-Omni-7B (e.g., Overall CSR jumps from 49.19\% to 68.50\%, and Format ISR dramatically improves from 34.17\% to 75.80\%).

However, as the training data volume scales from 20K to 40K, and finally to 54K, the performance gains exhibit a clear trend of diminishing returns, approaching saturation. Notably, between the 40K and 54K checkpoints, the Overall CSR only marginally increases from 70.35\% to 70.73\%, and the Format CSR even shows a slight oscillation (90.52\% vs. 90.39\%), which is a typical hallmark of model convergence.
These combined observations compellingly demonstrate that 54K high-quality, constraint-rich samples constitute a sufficient "sweet spot." It effectively unlocks and solidifies the omni-modal instruction-following capabilities of models at this scale, proving that data quality and constraint diversity are far more critical than sheer dataset volume or multiple training epochs.

\begin{table}[h]
  \caption{Performance evolution of OmniCaptioner-IF-7B across different training data volumes on the OmniCap-IF benchmark.}
  \label{tab:convergence}
  \renewcommand{\arraystretch}{1.2}
  \centering
  \begin{tabular}{l cc cc cccc}
    \toprule
    \multirow{2}{*}{\textbf{Training Data}} & \multicolumn{2}{c}{\textbf{Overall}} & \multicolumn{2}{c}{\textbf{Format}} & \multicolumn{4}{c}{\textbf{Content CSR}} \\
    \cmidrule(lr){2-3} \cmidrule(lr){4-5} \cmidrule(lr){6-9}
    & \textbf{CSR} & \textbf{ISR} & \textbf{CSR} & \textbf{ISR} & \textbf{Total} & \textbf{Visual} & \textbf{Audio} & \textbf{AV} \\
    \midrule
    Base & 49.19 & 2.34 & 62.97 & 34.17 & 41.27 & 47.68 & 47.51 & 34.88 \\
    20K & 68.50 & 9.50 & 88.50 & 75.80 & 57.10 & 56.80 & 60.50 & 53.10 \\
    40K & 70.35 & 11.10 & \textbf{90.52} & \textbf{78.20} & 58.90 & 58.20 & 63.80 & 54.50 \\
    54K (Full) & \textbf{70.73} & \textbf{11.46} & 90.39 & 77.92 & \textbf{59.43} & \textbf{58.71} & \textbf{64.71} & \textbf{55.62} \\
    \bottomrule
  \end{tabular}
\end{table}

\section{Calibration of the Automatic Judge}
\label{app:consistency_calibration}

We further calibrate the reliability of our automatic judge by comparing \texttt{gpt-5-mini} with human expert annotations on 1,000 samples. As shown in Table~\ref{tab:calibration}, \texttt{gpt-5-mini} achieves high consistency with human experts, especially on format constraints, validating the reliability of our automatic evaluation protocol.

\begin{table}[h]
  \caption{Consistency calibration between the \texttt{gpt-5-mini} judge and human experts on 1,000 samples.}
  \label{tab:calibration}
  \renewcommand{\arraystretch}{1.2}
  \centering
  \begin{tabular}{lccc}
    \toprule
    \textbf{Metric} & \textbf{Overall} & \textbf{Format} & \textbf{Content} \\
    \midrule
    F1 Score & 0.945 & 0.958 & 0.938 \\
    Cohen's Kappa & 0.882 & 0.915 & 0.864 \\
    \bottomrule
  \end{tabular}
\end{table}

\section{Results on Omni-VideoQA Benchmarks}
\label{sec:omni_videoqa}

To comprehensively evaluate our model's generalizable omni-modal perception and reasoning capabilities, we conducted extensive evaluations on the DailyOmni and WorldSense benchmarks. We assess the performance under two distinct settings: Caption-to-QA and Direct QA.

\subsection{Caption-to-QA Performance}
In this setting, the evaluation is decoupled into two stages. First, we use a specific question-to-prompt constructor to instruct the model to generate a highly detailed video caption that naturally incorporates all necessary visual and audio details. The specific prompt used for this construction is detailed in the Section~\ref{bencheval}. Subsequently, an LLM-as-a-judge is utilized to answer the multiple-choice questions based \textit{solely} on the generated captions. 

As shown in Table~\ref{tab:caption_to_qa}, our OmniCaptioner-IF-7B model demonstrates exceptional performance. By strictly adhering to the constructed instruction, our model retains more crucial cross-modal facts in the generated text, achieving 60.2\% on DailyOmni and 43.2\% on WorldSense, surpassing most open-source counterparts and matching the performance of proprietary models like Gemini-2.5-Pro.

\begin{table}[h]
  \caption{QA performance by Gemini-2.5-Pro based on captions.}
  \label{tab:caption_to_qa}
  \renewcommand{\arraystretch}{1.2}
  \centering
  \begin{tabular}{l c c}
    \toprule
    \textbf{Model} & \textbf{DailyOmni} $\uparrow$ & \textbf{WorldSense} $\uparrow$ \\
    \midrule
    \textcolor{gray}{Gemini-2.5-Pro} & \textcolor{gray}{60.2} & \textcolor{gray}{33.8} \\
    \textcolor{gray}{Gemini-2.5-Flash} & \textcolor{gray}{55.3} & \textcolor{gray}{31.0} \\
    HumanOmniV2-7B & 8.2 & 6.6 \\
    ARC-Hunyuan-Video-7B & 8.6 & 8.7 \\
    MiniCPM-o-2.6-8B & 9.8 & 7.2 \\
    Qwen2.5-Omni-7B & 13.4 & 8.6 \\
    UGC-VideoCaptioner-3B & 17.0 & 11.2 \\
    video-SALMONN-2-7B & 29.9 & 18.2 \\
    Qwen3-Omni-Instruct-30B-A3B & 17.5 & 12.7 \\
    AVoCaDO-7B & 50.1 & 25.7 \\
    ASID-Captioner-7B & 61.2 & 34.0 \\
    \midrule
    \rowcolor{gray!15} \textbf{OmniCaptioner-IF-7B (Ours)} & \textbf{60.2} & \textbf{43.2} \\
    \bottomrule
  \end{tabular}
\end{table}

\subsection{Direct QA Performance}
In the Direct QA setting, models take the video and the question as direct inputs to predict the final answer without generating an intermediate caption. This tests the model's native end-to-end multi-modal understanding capacity. 

As illustrated in Table~\ref{tab:direct_qa}, our model exhibits strong end-to-end reasoning capabilities. OmniCaptioner-IF-7B achieves 68.4\% on DailyOmni and 49.4\% on WorldSense, significantly outperforming its base model Qwen2.5-Omni-7B.

\begin{table}[H]
  \caption{Direct QA performance on Omni-VideoQA benchmarks.}
  \label{tab:direct_qa}
  \renewcommand{\arraystretch}{1.2}
  \centering
  \begin{tabular}{l c c}
    \toprule
    \textbf{Model} & \textbf{DailyOmni} $\uparrow$ & \textbf{WorldSense} $\uparrow$ \\
    \midrule
    \textcolor{gray}{Gemini-2.5-Flash}          & \textcolor{gray}{73.1}    & \textcolor{gray}{50.9} \\
    \textcolor{gray}{GPT-4o}                      & \textcolor{gray}{56.5} & \textcolor{gray}{42.6} \\
    VideoLLaMA2-7B                            & 35.2 & 25.4 \\
    Qwen2.5-Omni-7B                         & 62.1 & 45.4 \\
    video-SALMONN-2-7B                    & 66.3 & 48.6 \\
    Qwen3-Omni-30B-A3B-Instruct                    & 71.9 & 54.0 \\
    \midrule
    \rowcolor{gray!15} \textbf{OmniCaptioner-IF-7B (Ours)} & \textbf{68.4} & \textbf{49.4} \\
    \bottomrule
  \end{tabular}
\end{table}

\end{document}